*[Machine Learning: Engineering]* **Roadmap**

# 2026 Roadmap on Artificial Intelligence and Machine Learning for Smart Manufacturing

**Jay Lee[1,32,33], Hanqi Su[1,32,33], Marco Macchi[2], Adalberto Polenghi[2], Wei Wu[3], Zhiheng Zhao[3], George Q. Huang[3], Kiva Allgood[4], Devendra Jain[4], Benedikt Gieger[4], Vibhor Pandhare[31], Soumyabrata Bhattacharjee[5], Ram Mohril[5], Lingbao Kong[6], Qiyuan Wang[6], Xinlan Tang[6], Sungjong Kim[7], Chan Hee Park[8], Byeng D. Youn[7,9], Guo Dong Goh[10], Xi Huang[10], Wai Yee Yeong [10,11], Yung C Shin[12], He Zhang[13], Zitong Wang[13], Fei Tao[13,14], Jagjit Singh Srai[15], Satyandra K. Gupta[16], Byung Gun Joung[17], Albin John[17], John W. Sutherland[17], Sang Won Lee[18], Olga Fink[19], Vinay Sharma[19], Faez Ahmed[20], Wei "Wayne" Chen[21], Mark Fuge[22], Arild Waaler[23], Martin G. Skjæveland[23], Dimitris Kyritsis[23], Wei Chen[24], Vispi Nevile Karkaria[24], Yi-Ping Chen[24], Ying-Kuan Tsai[24], Joseph Cohen[25], Xun Huan[26], Jing (Janet) Lin[27], Liangwei Zhang[28], Gregory W. Vogl[29], Aaron W. Cornelius[29], Xiaodong Jia[30], Dai-Yan Ji[1], Takanobu Minami[1], Ruoxin Wang[1]**

[1] Center for Industrial Artificial Intelligence, Department of Mechanical Engineering, University of Maryland, College Park, 20742, United States of America

[2] Department of Management, Economics and Industrial Engineering, Politecnico di Milano, Milan, Italy

[3] Department of Industrial and Systems Engineering, The Hong Kong Polytechnic University, Hong Kong 999077, People's Republic of China

[4] Centre for Advanced Manufacturing & Supply Chains, World Economic Forum, Cologny/Geneva, Switzerland

[5] Department of Mechanical Engineering, Indian Institute of Technology Indore, Indore, India

[6] Future Information Innovative College, Fudan University, Shanghai, China

[7] Department of Mechanical Engineering, Seoul National University, Seoul 08826, Republic of Korea

[8] Department of Mechanical and Information Engineering, University of Seoul, Seoul 02556, Republic of Korea

[9] Onepredict Corp., Seoul 06105, Republic of Korea

[10] School of Mechanical and Aerospace Engineering, Nanyang Technological University, 50 Nanyang Avenue, Singapore 639798, Singapore

[11] Singapore Centre for 3D Printing, Nanyang Technological University, 50 Nanyang Avenue, Singapore 639798, Singapore

[12] Mechanical Engineering, Purdue University, West Lafayette, Indiana, U.S.A.

[13] Digital Twin International Research Center, International Institute for Interdisciplinary and Frontiers, Beihang University, Beijing, China

[14] School of Automation Science and Electrical Engineering, Beihang University, Beijing, China

[15] Department of Engineering, University of Cambridge, UK

[16] Center for Advanced Manufacturing, University of Southern California, Los Angeles, CA, USA

[17] School of Sustainability Engineering and Environmental Engineering, Purdue University, West Lafayette, USA

[18] School of Mechanical Engineering, Sungkyunkwan University, Suwon-si, Republic of Korea

[19] Intelligent Maintenance and Operations Systems, EPFL, Lausanne, Switzerland

[20] Department of Mechanical Engineering, Massachusetts Institute of Technology, Cambridge, USA

[21] J. Mike Walker '66 Department of Mechanical Engineering, Texas A&M University, College Station, USA

[22] Department of Mechanical and Process Engineering, ETH Zürich, Switzerland

[23] Department of Informatics, University of Oslo, Norway
[24] Department of Mechanical Engineering, Northwestern University, Evanston, IL, USA
[25] Department of Mechanical and Aerospace Engineering, Rutgers University, Piscataway, NJ, USA
[26] Department of Mechanical Engineering, University of Michigan, Ann Arbor, MI, USA
[27] Department of Civil, Environmental and Natural Resources Engineering, Luleå University of Technology, Luleå, Sweden
[28] Department of Industrial Engineering, Dongguan University of Technology, Dongguan, China
[29] Engineering Laboratory, National Institute of Standards and Technology, Gaithersburg, USA
[30] Department of Mechanical and Materials Engineering, University of Cincinnati, Cincinnati, USA
[31] Department of Mechanical Engineering, Indian Institute of Technology Bombay, Mumbai, India

[32] Guest Editors of the Roadmap.
[33] Author to whom any correspondence should be addressed.

E-mails: leejay@umd.edu, hanqisu@umd.edu

## Abstract

The evolution of artificial intelligence (AI) and machine learning (ML) is reshaping smart manufacturing by providing new capabilities for efficiency, adaptability, and autonomy across industrial value chains. However, the deployment of AI and ML in industrial settings still faces critical challenges, including the complexity of industrial big data, effective data management, integration with heterogeneous sensing and control systems, and the demand for trustworthy, explainable, and reliable operation in high-stakes industrial environments. In this roadmap, we present a comprehensive perspective on the foundations, applications, and emerging directions of AI and ML in smart manufacturing. It is structured in three parts. The first highlights the foundations and trends that frame the evolution of AI in smart manufacturing. The second focuses on key topics where AI is already enabling advances, including industrial big data analytics, advanced sensing and perception, autonomous systems, additive and laser-based manufacturing, digital twins, robotics, supply chain and logistics optimization, and sustainable manufacturing. The third section explores non-traditional machine learning approaches that are opening new frontiers, such as physics-informed AI, generative AI, semantic AI, advanced digital twins, explainable AI, RAMS, data-centric metrology, large language models, and foundation models for highly connected and complex manufacturing systems. By identifying both opportunities and remaining barriers across these areas, this roadmap outlines the advances needed in methods, integration strategies, and industrial adoption. We hope this roadmap will serve as a guide for researchers, engineers, and practitioners to accelerate innovation, align academic and industrial priorities, and ensure that AI-driven smart manufacturing delivers reliable, sustainable, and scalable impact for the future of manufacturing ecosystems.

## Contents

# Introduction

**Hanqi Su[1] and Jay Lee[1]**

[1] Center for Industrial Artificial Intelligence, Department of Mechanical Engineering, University of Maryland, College Park, 20742, United States of America

E-mail: hanqisu@umd.edu

The evolution toward smart manufacturing can be traced through several stages over the past six decades. Summarized in **Figure 1**, it highlights five phases: the early foundational phase (1960s–2000s), the new foundational phase (2000s–2010s), the rise of AI/ML-enabled smart manufacturing (2014–2025), and the progression toward next-generation AI for future manufacturing (2025–2035).

In the mid-1960s, the concept of flexible manufacturing systems (FMS) was introduced to enable automated machining that could adapt to different products, with the first implementations appearing in the late 1960s. By the 1970s and 1980s, advances in digital technologies gave rise to computer-integrated manufacturing systems (CIMS), emphasizing the integration of CAD, CAM, robotics, and enterprise systems for end-to-end production management. In the early 1990s, the notion of agile manufacturing systems (AMS) emerged, focusing on responsiveness and adaptability in the face of globalization and rapidly changing customer needs. Around the same time, intelligent manufacturing systems (IMS) became a formal international research program. The IMS Program, launched in 1990–1991, was a collaborative initiative involving Japan, European Union, United States, and later other countries. It was coordinated by organizations such as the International IMS Steering Committee and supported by governments and industries. The goal was to develop intelligent, distributed, and adaptive manufacturing systems through global cooperation. These milestones collectively formed the **early foundational phase**, establishing automation and integration as the baseline for modern manufacturing.

With the rise of digital infrastructure in the 2000s and 2010s, the **new foundational phase** was shaped by advances in data-driven connectivity, sensing, and operation. The introduction of the Internet of Things (IoT), cyber-physical systems (CPS), cloud computing, industrial big data analytics, prognostics and health management (PHM), and digital twins accelerated the digital and automated integration of design, production, inspection, and supply chain systems [1-4]. These advances laid the foundation for **Industry 4.0**, transforming traditional production into a mode characterized by digitalization, automation, and intelligence [5,6].

From 2014 onward, **AI/ML-enabled smart manufacturing** emerged as a central theme. **Smart manufacturing**, the cornerstone of modern manufacturing, is defined by the National Institute of Standards and Technology (NIST) as "fully integrated, collaborative manufacturing systems that respond in real time to meet changing demands and conditions in the factory, in the supply network, and in customer needs" [7]. This phase began with the application of deep learning (deep neural networks), transfer learning, explainable ML, and early multimodal fusion to solve specific manufacturing tasks and pilot deployments [8-10]. Building on the foundations of Industry 4.0, Industry 5.0 was proposed in 2021, introducing a vision that emphasizes human-centricity, sustainability, and resilience [11]. While Industry 4.0 focused on digitalization and automation for efficiency [6], Industry 5.0 stresses collaboration between humans and smart machines, circular economy principles to support sustainable production, and robust systems to ensure adaptability under disruptions [12]. At this stage, AI and ML methods are increasingly applied to reinforce these principles, placing human needs at the center of manufacturing processes. More recently, the focus has expanded to accelerating and scaling advanced methods such as federated learning, multi-modal learning, physics-informed learning, and large language models (LLMs) into complex workflows [8-10]. Looking forward, the next-generation AI for manufacturing phase (2025–2035) is expected to be enabled by generative AI, agentic AI, industrial LLMs, and large-scale foundation models, driving new productive ecosystems for scalable and sustainable manufacturing [13, 14].

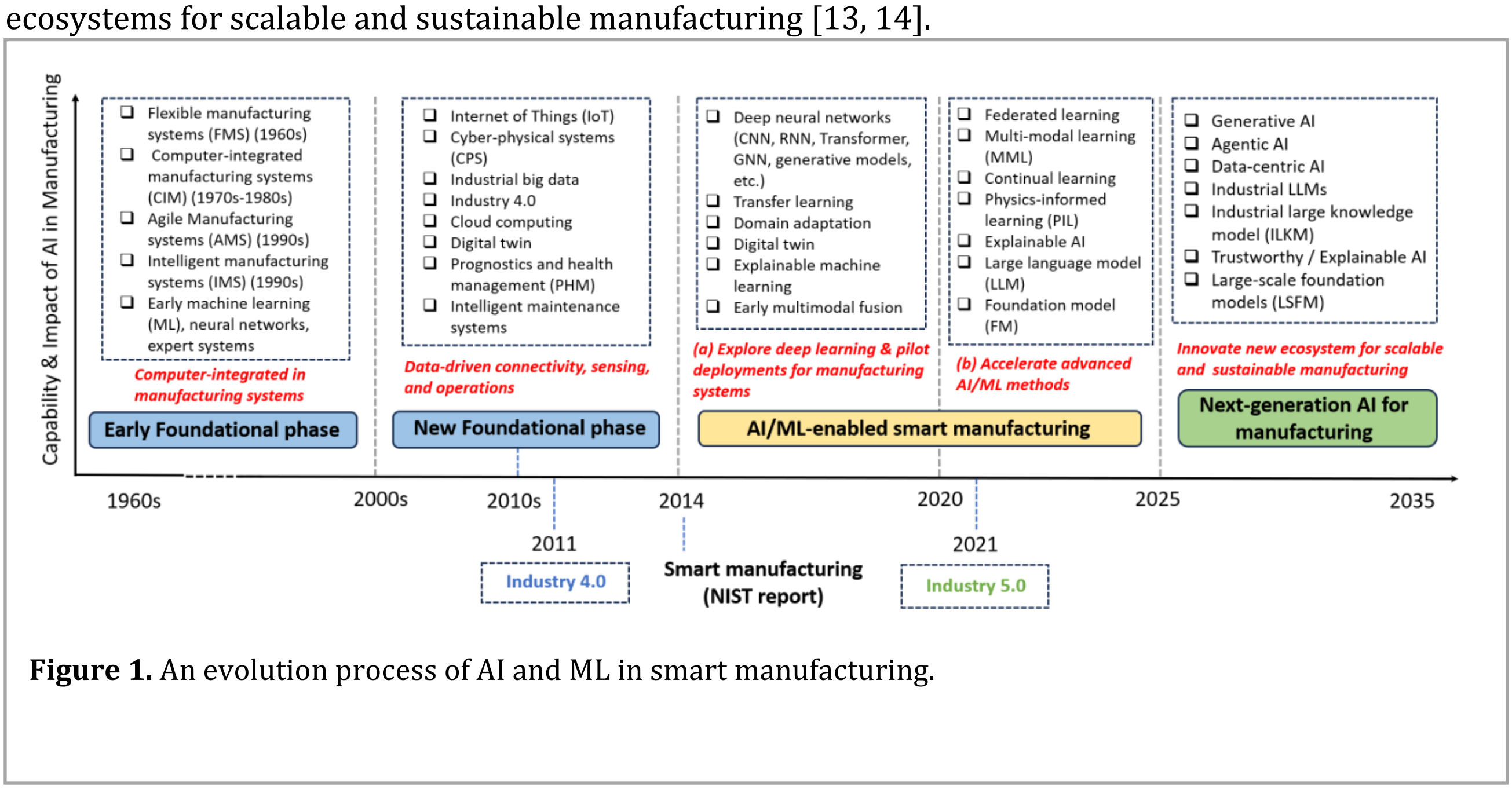

**Figure 1.** An evolution process of AI and ML in smart manufacturing.

In recent years, AI and ML have emerged as foundational technologies applied to many aspects of smart manufacturing [10, 15, 16]. On the one hand, with the growing availability of data and the increasing demand for analytics in the era of industrial big data, AI and ML techniques enable the efficient processing of large-scale data streams from equipment, sensor networks, and supply chains. These methods support the extraction of actionable insights from complex datasets and help manufacturers to perform intelligent decision-making in a timely manner. On the other hand, the integration of AI and ML has significantly advanced the automation of manufacturing processes. Previously, tasks such as quality control, product inspection, equipment maintenance, and production scheduling relied heavily on manual intervention. Right now, these tasks can be handled by intelligent algorithms. This shift reduces human workload and improves accuracy, consistency, and operational efficiency. Furthermore, with growing demand for customized and personalized products, AI and ML models can facilitate the analysis of consumer preferences and enable dynamic adjustments to production lines. As a result, manufacturers can reduce time-to-market (TTM) and improve overall market responsiveness. In addition, another major advancement in smart manufacturing is the shift from reactive to predictive maintenance. Traditional manufacturing depends on periodic maintenance and manual inspections to prevent equipment failures. In contrast, AI- and ML-based techniques for predictive maintenance and fault diagnosis can detect early signs of degradation or abnormal behaviour, and provide timely alerts before failures occur. By leveraging data-driven prediction and classification methods, manufacturing enterprises can reduce sudden shutdowns and production disruptions, achieving lower maintenance costs.

Although AI and ML have brought significant potential to smart manufacturing, notable gaps remain between current AI and ML capabilities and the practical requirements of modern manufacturing systems. One of the most critical challenges is data quality. Industrial data are often noisy, incomplete, or presented in inconsistent formats, which can make them difficult to use in practice. These issues prevent effective training of AI and ML models and compromise the reliability of their predictions [17]. Another limitation is the lack of interpretability in many AI models. While black-box models such as deep neural networks can achieve strong performance, their internal decision-making processes are often unclear. This lack of transparency makes it difficult for engineers and practitioners to fully understand or trust the outputs of AI models, particularly in high-stakes industrial applications [18]. Moreover, many manufacturing processes are governed by complex physical principles that are difficult to represent using purely data-driven models. Traditional approaches often struggle to effectively integrate these complex physical laws into AI and ML models [19]. In addition, AI and ML models trained on data from one machine, production line, or factory often perform poorly when applied to different domains, due to distribution shifts and limited availability of labeled data in the target domain. This challenge is commonly referred to as domain adaptation [20]. Finally, although AI models often demonstrate strong performance in laboratory settings, their deployment in real-world production faces challenges. These include system integration complexity, real-time performance requirements, and scalability across multiple factories and distributed supply chains.

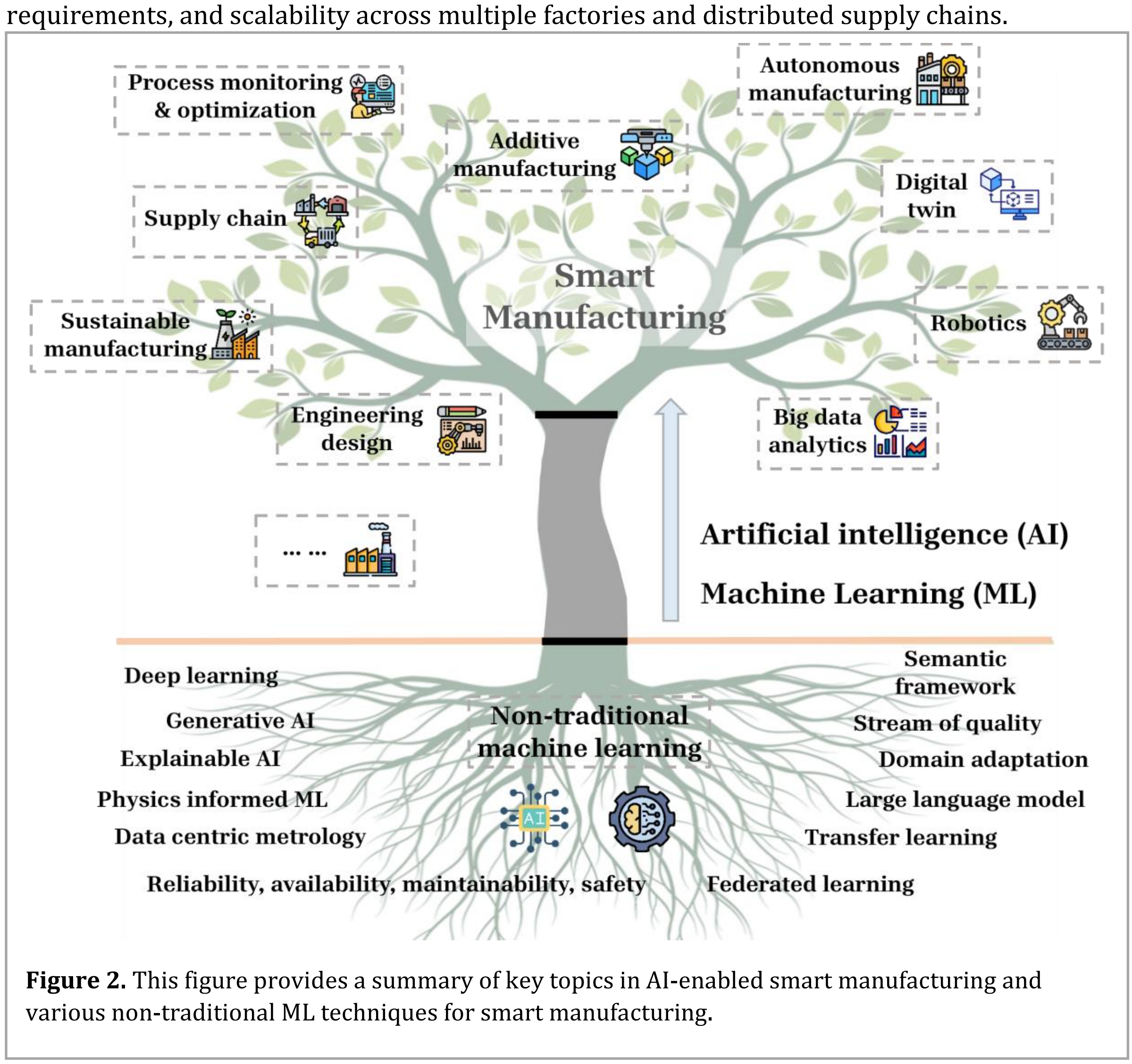

**Figure 2.** This figure provides a summary of key topics in AI-enabled smart manufacturing and various non-traditional ML techniques for smart manufacturing.

In response to the growing importance of AI and ML in transforming modern manufacturing, the aim of this roadmap on artificial intelligence and machine learning for smart manufacturing is to provide an overview of different research areas and technological developments driving progress in smart manufacturing, as shown in **Figure 2**. It outlines opportunities, challenges, and technological advancements for the next-generation manufacturing industry. The roadmap is organized into three main sections:

- Foundations and trends in AI for smart manufacturing: This section reviews the evolution of AI and ML in manufacturing, emphasizing their growing role in enhancing efficiency, adaptability, and automation for future manufacturing. It also discusses the outlook of AI technologies and their potential to transform manufacturing industry and value chains.
- Key topics in AI-enabled smart manufacturing: This section focuses on major application areas of AI and ML in smart manufacturing. It covers topics such as industrial big data analytics, autonomous manufacturing, additive manufacturing, process monitoring and optimization, digital twins, engineering design, smart supply chain and logistics, sustainable and green manufacturing, and AI-enhanced robotics.
- Non-traditional machine learning techniques for smart manufacturing: This section discusses emerging non-traditional machine learning paradigms and their relevance to smart manufacturing. Topics covered include deep learning, generative AI, physics-informed ML, semantic frameworks, and trustworthy and explainable AI. It also highlights developments in data-centric metrology, reliability, availability, maintainability, and safety (RAMS) in AI-enabled systems, as well as advances in large language models, industrial large knowledge models, domain adaptation, transfer learning, stream-of-quality analysis, and federated learning.

We hope that this roadmap offers a comprehensive perspective and a long-term strategic foundation for advancing AI and ML in smart manufacturing. Each contribution, authored by leading researchers in their domains, presents the current state of the field, identifies key challenges, outlines the advances in science and technology needed to address them, and proposes future perspectives. In the end, we encourage broader collaboration among academic researchers, industry practitioners, funding agencies, and policymakers to work together to shape the future of manufacturing.

## Acknowledgements

This work was supported by the U.S. Department of Education through the Fund for the Improvement of Postsecondary Education (FIPSE) under Grant No. P116S230014.

# Section 1: Foundations and Trends in AI for Smart Manufacturing

# The evolution of Artificial Intelligence and Machine Learning for Smart Manufacturing

**Marco Macchi[1] and Adalberto Polenghi[1]**

[1] Department of Management, Economics and Industrial Engineering, Politecnico di Milano, Milan, Italy

E-mail: marco.macchi@polimi.it, adalberto.polenghi@polimi.it

### Status

The advances in various cutting-edge technologies and the opportunities for transformation emerging in the industrial environment are today placing the power of Artificial Intelligence (AI) applied to industrial processes at the top of the research agenda. A new term, Industrial AI, has been recently coined [1,2]; this was introduced to emphasize that, based on AI as technological foundation, on data and algorithms, and on software and hardware components, it is now imperative to bring AI to work in industrial systems through scalable applications with sustainable performance [3].

As we know, AI is not the only driver. The push forward is leading to an evolution towards a new paradigm envisaged by smart manufacturing [4,5]. The transition is supported by Industrial Internet of Things (IIoT), virtual manufacturing, Industrial AI and other enablers. The cloud-to-edge continuum ultimately supports the proximity of physical and virtual spaces in the deployment of computational intelligence [6].

In a broader perspective, the physical-digital convergence is a long-lasting development, fostered by the adoption of Cyber-Physical Systems (CPS), but not only. It is an effect resulting from the development of multiple technologies due to computer science, information and communication technologies (ICT), manufacturing science and technology, finally leading to the convergence between the physical and virtual worlds [7,8]. As an aggregate impact, the cyber-physical integration problem should be addressed, and Industrial AI is an integral part of this problem.

Looking at the virtual space, one can even envision a synergistic development for the coming years where AI will be a key component of Digital Twins of physical entities and systems in manufacturing, thanks to its capability to provide insights aimed at identifying hidden patterns and establishing correlations, and making predictions and optimizations of the future behaviour [9,10]. Then, a foreseeable scenario is to proceed towards a convergence where the Industrial Metaverse (IM), employing advanced technologies such as IIoT, Blockchain and Augmented/Virtual Reality (A/VR), will enable the construction of an immersive virtual space able to seamlessly interact with the physical space, facilitating human interaction in an advanced collaborative manufacturing [11,12] where Human-To-X collaboration is central.

In this evolutionary trend, AI will play an essential role, both in industrial processes and in the relationship with humans within the decision-making loop [13]. It will support knowledge representation, machine/deep learning, reasoning and optimal problem-solving, thus integrating the advanced modelling and simulation technologies at the core of Digital Twins and Industrial Metaverse.

**Current and future challenges**

Current challenges in AI/ML are long-lasting to be addressed so that R&D activities lead to the provision of industrially ready solutions in the context of large-scale implementation of AI and machine learning (ML) in smart manufacturing. The integration of AI in smart manufacturing is not straightforward: technological and conceptual developments are required to make those solutions effective and scalable so that companies can leverage them to improve and achieve sustainable performance.

Identified challenges that are timely and relevant to face come from proper mix of research and industrial experiences and are hereafter synthetised:

1. Data-driven approaches showed limited impacts in terms of adoption in manufacturing companies. Knowledge of industrial processes is essential and must complement data-driven approaches. Indeed, domain-specific knowledge is necessary in order to rapidly develop capabilities in new tasks for new technologies and products as well as manufacturing processes and equipment [14].
2. AI and ML are mainly limited to reaction to local and confined drifts and anomalies [15], but AI and ML need to scale up to handle complex systems, predicting and optimizing their behaviours globally, and should be challenged by the increased responsiveness to adapt to changes arising from new products, equipment and technologies, processes.
3. Most AI research and development focus on technical performance of the model/solution without tackling the way in which the solution is embedded in a complex socio-technical environment as, for example, manufacturing shopfloors are, where only a multi-disciplinary approach can work out. However, so far, AI state of art shows that there are very few examples of AI-powered solutions that embrace such new research paradigm [16].
4. Cognitive adaptation of manufacturing systems, implying autonomous execution of actions based on certain inputs and triggers from the system, is currently a look-ahead in research. It requires the collection and elaboration of data related to the system and to the context, properly elaborated so that Digital Twins and, generally speaking, AI-powered solutions, may be agent rather than pure informative systems for human-based decisions and actions [17].

Solving these challenges will then open future ones connected to the way machines, humans and AI will interact. The physical-digital convergence will lead to shadowed shopfloors in which what is “real” is merged between physical and virtual inputs and the Industrial Metaverse will be new way in which manufacturing companies should work from the design to the management of manufacturing systems.

**Advances in science and technology to meet challenges**

AI and machine learning are expected to enhance the Digital Twins of manufacturing systems. This will lead to synergies that will enable decision intelligence to grow towards higher levels of adaptability, intelligence and cognitive traits [17]. To support this growth, the AI-powered Digital Twins of manufacturing systems will be enriched by capabilities built on the adoption of behavioural models of human operators, continuous and reciprocal learning between humans and AI/machine learning models, human experience between virtual and real worlds, and augmentation of decisions through an increasingly cognitive collaboration between physical systems and human decision-makers [17,18].

In this path, technological advances are drastically increasing the capabilities to improve manufacturing operations. Regarding AI and ML, the hype is today focused on no-code AI and vibe-coding, which are making easier the development of advanced solutions; it is anyway relevant to remind that industry and business-grade AI-powered applications require heavy computer engineering and ICT expertise. AI itself is seeing an empowerment in terms of explainability capabilities so as to better engage with humans, providing not only results but the reason why such results has been obtained. Finally, LLMs (Large Language Models) and co-pilots, and underneath foundation models, are completely reshaping the way in which humans and machines interact, leading to new forms of HMI (Human Machine Interface) and dashboarding, that is more natural for human decision-makers; this interaction mode will be even more fruitful if complemented by immersive XR (Extended Reality) technologies towards full Industrial Metaverse realisation.

Besides technological growth, new advancements also concern frameworks and theories to first assess and then introduce and scale-up AI-powered Digital Twins within manufacturing organisations considering human decision-makers as core actor [19]. Approaches such as systems engineering and MBSE (Model Based System Engineering) are thus important to conceptualise the relation between AI-powered Digital Twins with other technologies and human as well non-human agents. Furthermore, theories like SUT (System Usability Theory) or UTAUT (Unified Theory of Acceptance and Use of Technology) must be considered and managed within the scope of AI engineering and deployment. To this end, it is advisable for researchers and practitioners to work synergistically so that the solution is first defined within a use case and then checked for fit against business/economics, operational performance and human behaviour [20].

The vision for the future is outlined in the following Figure 1. To achieve it, a manufacturing company follows a path toward smart manufacturing that starts with existing manufacturing plants and systems. AI is a key pillar for advances in decision intelligence and is integrated into an evolving platform resulting from the combination of different technology stacks, both due to legacy IT systems and manufacturing equipment, and new equipment and tools, also those designed to support humans in IT/OT systems. Therefore, physical-digital convergence is envisioned in a future Industrial Metaverse as a natural trend originating in the IIoT and evolving through the development of digital twins of machines, humans and manufacturing systems and their XR extension. In this framework, AI plays a key role for the intelligence in terms of perception, learning, prediction, interaction, adaptation, reasoning and creativity.

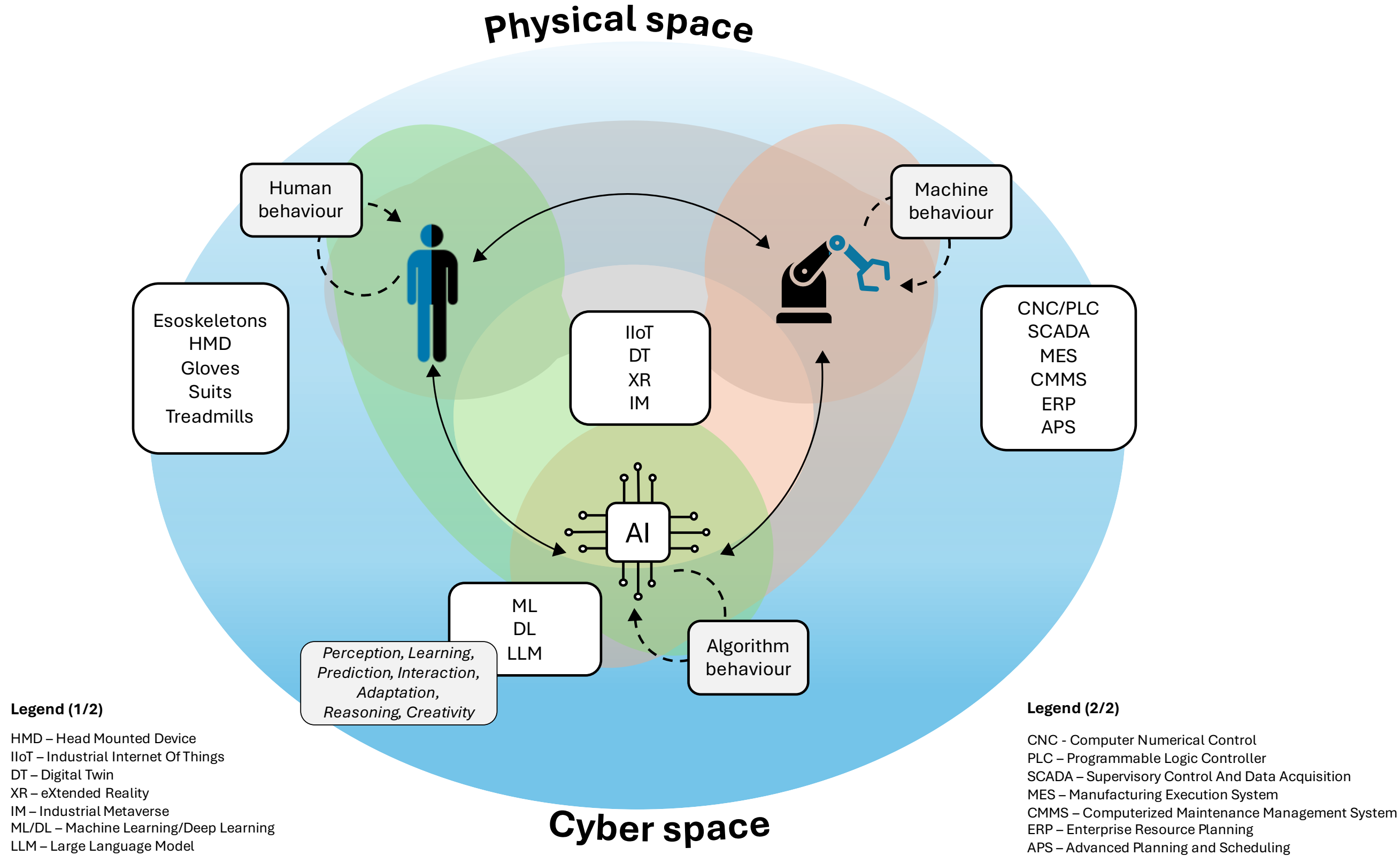

**Figure 1** - Look-ahead for smart manufacturing systems

## Concluding remarks

The way in which AI is permeating manufacturing companies is continuously evolving especially pushed by technological advancements. AI-powered Digital Twins are now a foreseeable reality in the manufacturing realm and will support humans, from operators to engineers and managers, in design, production and maintenance of products and manufacturing systems. Most of the development as of now is one-way, from AI to human, but for the former to become a critical part in manufacturing, adaptability is necessary as key capability to look for autonomous systems able to manage and react to non-trivial, context-dependent events. Therefore, AI-powered solutions become crucial resources with which humans can interact, being mutually informed to empower and strengthen decision-making. What is envisioned is that current AI state of art must move forward by disrupting the current human-machine-AI communication means towards a seamless convergence of physical and virtual worlds into the Industrial Metaverse concept. A blend of technologies is necessary for this, from IIoT to XR. New HMI based on LLMs will be the new norm and AI agents with diversified capabilities will be actors in CPS-based smart factories where humans can equally interact between themselves with machines and intangible solutions with their own learning capabilities., leading to different forms of Human-to-X collaboration. This will bring collaboration to a new frontier in industrial engineering only if this will be optimally developed and orchestrated with human-focused methodologies, considering the human-in-the-loop, and combining traditional engineering performance evaluation with aspects such as learnedness, usability and ergonomics.

## Acknowledgements

This research is part of the HumanTech Project, which is financed by the Italian Ministry of University and Research (MUR) for the 2023-2027 period as part of the ministerial initiative “Departments of Excellence”

(L. 232/2016). The initiative rewards departments that stand out for the quality of the research produced and funds specific development projects.

# The importance of AI-driven efficiency, adaptability, and automation for future manufacturing

**Wei Wu[1], Zhiheng Zhao[1] and George Q. Huang[1]**

[1] Department of Industrial and Systems Engineering, The Hong Kong Polytechnic University, Hong Kong 999077, People's Republic of China

E-mail: gq.huang@polyu.edu.hk

**Status**

The future of manufacturing, potentially shaped by Industry 5.0, emphasizes the creation of more human-centric, resilient, and sustainable manufacturing ecosystem capable of mass personalization [1]. Within the transformation, AI is indispensable, fundamentally enhancing efficiency, adaptability and automation across a hierarchy of facilities. Efficiency, in this context, describes the capability of optimizing production processes to maximize output while minimizing resource consumption and operational timelines. AI helps to ensure men, machines, and materials operate cohesively in the right place, at the right time, with minimal inefficiencies [2]. Adaptability denotes the capacity of seamlessly adjusting to dynamic environments, including fluctuating market demands and unforeseen disruptions. The enhanced cyber-physical visibility and traceability empowered by AI facilitate manufacturers to identify disruptions, make data-driven decisions, and quickly adapt processes to meet shifting requirements [3]. Automation concerns the autonomous management and execution of repetitive or complex tasks with minimal human intervention. The convergence of robotics, Internet of Things (IoT), and AI techniques enable individuals or systems to perform accurate, effective, and consistent decision-making, alongside self-learning and self-optimization [4]. In a competitive global market, AI adoption is crucial for maintaining a competitive edge and achieving sustainability objectives.

AI technologies are now pervasively deployed in the manufacturing sector [5] (figure 1). Predictive maintenance [6], for example, utilizes AI to analyze sensor data and forecast equipment failures, thereby mitigating downtime and maintenance cost. AI-powered real-time scheduling and execution [7], underpinned by seamless cyber-physical synchronization, enhance production robustness against operational uncertainties and dynamic changes. The automation of manufacturing tasks [8] via AI-driven robotic systems continues to elevate productivity and streamline workflows. Furthermore, AI plays a pivotal role in optimizing supply chains [9] through improved demand forecasting and inventory management. Generative design tools [10] leverage AI to explore extensive design possibilities based on historical prototypes. These diverse applications are converging towards the realization of "smart factories" that ensure highly automated, efficient, and adaptive production environments.

Further advances in AI promise even more profound impacts on manufacturing. We can anticipate greater levels of autonomy in manufacturing processes. Enhanced human-machine collaboration will see AI augmenting human capabilities, allowing workers to focus on more complex, creative, and strategic tasks. The ability to offer mass personalization and highly flexible production systems will become increasingly prevalent, allowing manufacturers to respond rapidly to changing market demands and individual customer preferences. The ongoing evolution of generative AI, in particular, is expected to drive further innovation and transformative changes across the manufacturing domain.

**Current and future challenges**

Despite the unprecedented potential of AI in manufacturing, its broad and effective implementation is impeded by significant challenges. A primary hurdle lies in data-related issues [11]. The efficacy of AI systems is heavily dependent on access to large quantities of high-quality, consistent, and accurately labelled data. However, many manufacturing enterprises grapple with outdated legacy systems, widespread data silos, and a lack of integrated data governance. These limitations often result in datasets that are noisy, incomplete, or poorly contextualized, necessitating laborious and costly pre-processing. Additionally, safeguarding data security and privacy is a paramount concern [12], particularly with the proliferation of distributed AI models. Protecting sensitive manufacturing data and intellectual property from cyber threats remains a critical challenge.

Integration complexity represents another significant obstacle [13]. Modern manufacturing environments are characterized by a heterogeneous technological landscape, wherein advanced information systems coexist with aging legacy equipment that often lacks standardized communication protocols or digital interfaces. The integration of AI solutions into such disparate infrastructures is technically intricate and operationally disruptive. Moreover, the lack of interoperability between AI platforms and off-the-shelf systems, such as Manufacturing Execution Systems (MES) or Enterprise Resource Planning (ERP), further complicates seamless deployment. It requires substantial infrastructure upgrades, which would increase both cost and time.

Equally critical are concerns surrounding the safety, reliability, and trustworthiness of AI systems [14]. Ensuring fairness, transparency, and accountability in AI-driven decision-making is important in safety-critical applications. The use of biased or unrepresentative training datasets for AI models can reinforce existing inequities and produce distorted operational outcomes that potentially compromise process efficiency. Furthermore, the opaque nature of many advanced AI algorithms, commonly referred to as the 'black box' problem, introduces notable difficulties in validation, debugging, and fostering trust among human operators. Strengthening the robustness of these systems against adversarial attacks and unpredictable variations is essential to maintain long-term reliable operations.

Looking ahead, future challenges will prominently feature the need to warrant the scalability and flexibility [15] of AI solutions across diverse and evolving manufacturing environments. Transitioning to industry-wide deployment requires effective AI operations frameworks capable of accommodating increasing data volumes and rising model complexity. Moreover, AI systems must demonstrate enhanced adaptability to shifting production demands, reconfiguring manufacturing cells, or launching new product lines. Addressing these challenges is of paramount significance in unlocking the full potential of AI to revolutionize manufacturing systems on a global scale.

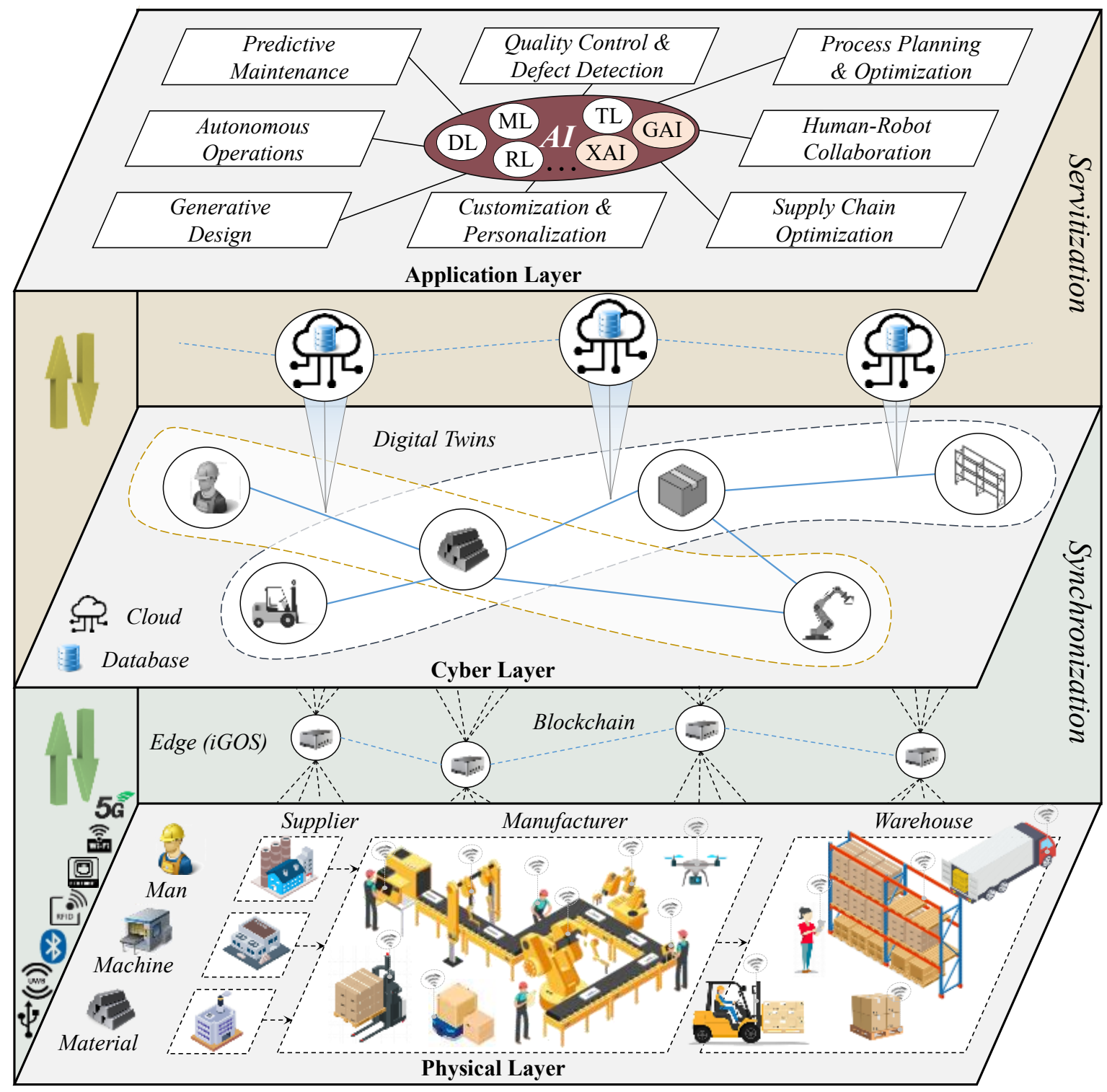

**Figure 1.** A roadmap of AI-driven manufacturing with the assistance of multiple advanced technologies. The physical layer, at the bottom, illustrates the fundamental elements in manufacturing, man, machine, and material, equipped with IoT devises for data collection, transmission, and computation. The cyber layer is designed to seamlessly mirror physical world through digital twins, while blockchain technology ensures information security and reliability. The top-level application layer leverages various AI techniques based on streams of data to enable a suite of smart services for manufacturing operations.

### Advances in science and technology to meet challenges

Several cutting-edge technologies are emerging and evolving to overcome the challenges hindering AI expansion in manufacturing, including digital twin (DT), hierarchical computing, and blockchain (figure 1). These technologies are expected to promote effective integration, robust reliability and scalable deployment.

DT technology [16] offers a solid solution for resolving integration complexity and ensuring reliable AI performance. By creating identical replicas of physical assets, processes, and systems in the cyber space, DTs enable real-time monitoring, simulation, and optimization of manufacturing operations. These virtual models can not only facilitate seamless synchronization between legacy systems and AI-driven platforms but augment numerous high-quality data in a generative way, thus considerably elevating model accuracy and consistency. Additionally, DTs enhance transparency and trust by providing a visible sandbox for validating AI algorithms and reducing the latency of decision-making.

Cloud-Fog-Edge-End computing architectures [17] establish the scalability and flexibility of AI deployment. Distributing computational resources across cloud, fog, edge, and end devices can enable efficient data processing closer to the source of generation, thereby improving resource utilization and response speed. Such architectures give rise to real-time decision-making in dynamic production environments and also ensure that AI systems can scale across geographically distributed facilities while maintaining operational efficiency. Edge computing, in particular, enhances data privacy and security by processing sensitive information locally, mitigating risks associated with centralized data storage.

Blockchain technology [18] provides a decentralized and immutable framework for enhancing data security, traceability, and trust in manufacturing environments. By enabling secure and tamper-proof data sharing among stakeholders, blockchain alleviates risks of data breaches and ensures compliance with relevant regulations. Smart contracts can automatically enforce access policies and trigger job sequences, tightening the coupling between data provenance and operational control. Furthermore, blockchain can improve the connectivity and traceability of cloud-fog-edge-end computing systems to strengthen AI safety.

Finally, advances in generative AI (GAI) [19] and explainable AI (XAI) [20] would reshape the manufacturing by fostering creativity, building trust, and optimizing processes. GAI focuses on creating new and original content based on latent representations learned from historical data. In the industrial context, foundation models trained on multimodal corpora can accelerate domain adaptation and provide human-readable instructions. Meanwhile, XAI aims to make AI system decisions and outputs transparent and understandable to humans. By addressing the Black Box problem inherent in many AI models, XAI facilitates human-AI collaboration while enhancing safety and reliability. Together, these technologies chart a credible path toward powerful, scalable, and trustworthy AI for the manufacturing of the future.

## Concluding remarks

In conclusion, the future of manufacturing is deeply intertwined with the advancement of AI. AI serves as a cornerstone for achieving higher levels of efficiency, adaptability, and automation essential to fostering a competitive, sustainable, and human-centric industrial ecosystem. While significant progress has been made in deploying AI within the manufacturing domain, several critical challenges persist, including issues related to data quality, the complexity of system integration, and the imperative for trustworthy and reliable AI systems. Emerging technologies such as digital twins, hierarchical computing, and blockchain present promising avenues to address these challenges by enhancing cyber-physical traceability and visibility, model reliability, process security, and system flexibility. Harnessing these innovations provides the means to overcome existing limitations and fully realize the transformative potential of AI in manufacturing. Moreover, GAI and XAI are anticipated to further accelerate innovation and redefine the manufacturing paradigm. This trajectory will pave the way for smart factories that are not only highly efficient and resilient but also capable of delivering mass personalization, thus securing a competitive edge in the global market and driving the next wave of industrial evolution.

## Acknowledgements

This work was supported in part by the Hong Kong RGC TRS Project (T32-707/22-N), in part by Collaborative Research Fund (C7076-22GF), in part by Research Impact Fund (R7036-22), and in part by Innovation and Technology Fund (PRP/007/25LI).

# The Outlook of Artificial Intelligence in Manufacturing and Value Chains

**Kiva Allgood[1], Devendra Jain[1] and Benedikt Gieger[1]**

[1] Centre for Advanced Manufacturing & Supply Chains, World Economic Forum, Cologny/Geneva, Switzerland

E-mail: mailto:kiva.allgood@weforum.org

### Status

AI has evolved from a promising technology to a transformative force, fundamentally reshaping global manufacturing and value chains. Over the past decade, AI and its applications have matured, driven by advances in data availability, algorithms, and compute power. As a result, manufacturers are increasingly recognizing AI's potential to drive step-change improvements in efficiency, sustainability, and resilience when deployed at scale. The World Economic Forum's Global Lighthouse Network[1], a community of advanced manufacturing sites, serves as a compelling showcase of such AI-enabled improvements.

The early use cases of AI focused on predictive maintenance and quality control. Today, AI's integration spans the full value chain: demand sensing, supply planning, autonomous intralogistics, energy optimization, and dynamic scheduling. Notably, much of the current impact still stems from conventional AI models, which continue to drive significant gains - often exceeding 50% in conversion cost, cycle times and defect rates[2]. Importantly, AI is no longer a siloed technology; it is becoming embedded across fit-for-purpose intelligent systems that are digital, adaptive, or autonomous.

In parallel, the entire manufacturing industry faces an unprecedented confluence of pressures: labor shortages, climate challenges, geopolitical tensions, and sustainability imperatives[3]. AI has the potential to close productivity gaps and demographic challenges, localize production, and decarbonize industrial operations. Advances in AI-driven simulation, self-learning agents, and hybrid human-AI collaboration models promise to redefine how products are designed, made, and moved[4].

Despite progress, the journey is far from complete. While some firms are moving toward full-scale deployment, many remain stuck in isolated pilots, hindered by fragmented data ecosystems, legacy infrastructure, talent shortages, or strategic misalignment. Bridging this gap will require scalable digital solutions, sustainability and resilience frameworks, strong data-management and a shift in workforce capabilities[5].

Looking ahead, the focus must shift from experimentation to scaled impact. For manufacturers, that includes positioning themselves along a transformation continuum that reflects the evolving integration of AI into industrial systems. This journey typically unfolds in three progressive phases: digital, adaptive, and autonomous. In the digital phase, firms focus on building foundational capabilities such as connected data infrastructures, real-time visibility, and process automation. In the adaptive phase, AI is leveraged for scenario simulation, predictive insights, and dynamic decision-making, enabling responsiveness to changing market conditions. The autonomous phase marks the emergence of self-optimizing, self-healing operations, where for example AI agents manage complex networks with minimal human intervention[6]. As manufacturers navigate this continuum, those who successfully harness AI as a strategic enabler, will define the next era of intelligent and sustainable value creation.

**Current and future challenges**

Scaled AI adoption is impeded by a set of interrelated technological, organizational, and ethical barriers, also shown in figure 1. Working with global industry leaders and lighthouse factories, the World Economic Forum recognized a consistent set of hurdles that must be addressed to unlock AI's next wave of transformative impact:

- **Data & Digital Core**: Despite the abundance of operational data, much of it remains siloed across departments, limiting end-to-end visibility. Many organizations operate heterogeneous IT systems (including legacy platforms, on-premise databases, and disparate cloud services) that were never designed for AI integration. The lack of interoperability and standardized data models inhibits development of scalable AI applications
- **Governance, Ethics & Transparency**: Accountability, fairness, and transparency is key when using AI, but the 'black-box' nature of many AI algorithms complicates efforts to understand, audit, or explain decisions. Biases embedded in training data or model design can result in discriminatory outcomes, potentially affecting suppliers, workers, or product quality. These risks are amplified by the rapid pace of AI innovation, with new models and capabilities emerging almost daily. Therefore, it is increasingly difficult for manufacturers to assess, validate, and govern these systems effectively[7].
- **Scaling beyond Pilots:** Many firms struggle to translate proof-of-concept initiatives into enterprise-wide platforms due to a lack of clear return on investment, or integration issues with legacy systems. This creates a paradox where firms acknowledge AI's strategic importance but underinvest in its full deployment. Repeated experimentation without systemic impact can also lead to a pilot fatigue, where stakeholders become disillusioned with AI's promised benefits.

- **Talent & Organizational Readiness**: Scaling AI in manufacturing also demands a significant shift in workforce capabilities and organizational culture. The skills required to develop and operate AI systems - ranging from data science and AI literacy to ethical reasoning and systems thinking needed for effective human-machine collaboration - are not yet widely distributed across the industrial workforce. Addressing this gap will require substantial change management efforts[8].
- **Transformation Complexity:** Compounding these challenges is the growing complexity of strategic transformation itself. Manufacturers are increasingly expected to align their AI efforts with both sustainability and resilience objectives. This shift requires the simultaneous optimization of efficiency, environmental impact, and adaptability[9]. Trade-offs between goals like rapid delivery versus carbon reduction can be managed through advanced AI-driven optimization and decision support. However, most organizations lack the cross-functional structures needed to orchestrate such a triple transformation.

Addressing these interdependent challenges is essential to unlocking the full potential of AI in manufacturing.

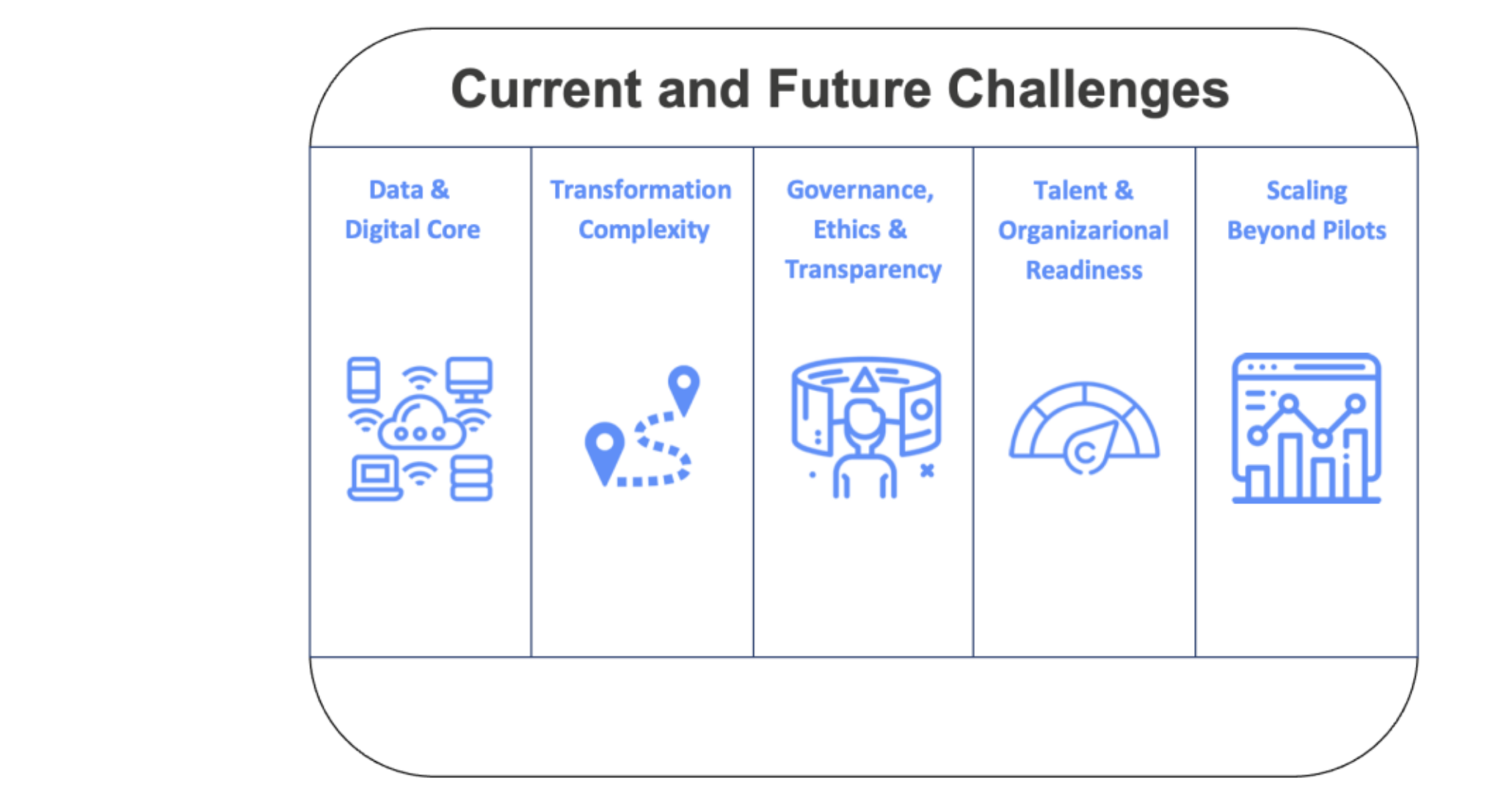

**Figure 1.** Five principal challenges that organizations face in the context of digital transformation and future readiness.

## Advances in science and technology to meet challenges

In response to the multifaceted challenges facing AI adoption in manufacturing, advances in science and technology must be directed to solving persistent barriers. A new generation of scientific and technological advancements is emerging, pushing the boundaries of what's possible in industrial settings:

- **Domain-specific Industrial Foundation Models:** Unlike large, general-purpose foundation models, smaller, domain-specific models trained on manufacturing-specific data such as machine logs and process parameters are on the rise. Their niche focus allows for more accurate, context-aware predictions while significantly reducing the computational resources and energy typically required by large-scale models[10]. Their compact size enhances deployability at the edge like on shop floors or within connected machinery, where latency, bandwidth, and data privacy are critical concerns.

- **Explainability Tools**: The growing availability and integration of **explainability tools** allows interpretations of AI decisions. In industrial environments, where safety, compliance, and trust are paramount, the ability to understand why an AI system has made a specific recommendation is essential. Explainable AI (XAI) techniques enable users to trace outcomes back to input factors and assumptions. This transparency builds trust, facilitates regulatory compliance, and allows human experts to validate AI outputs when necessary, ensuring that automation enhances, rather than undermines, operational integrity.
- **Agentic Systems:** One of the most significant advancements is the development of intelligent operations through AI agents, both virtual and embodied. Virtual agents operate within software environments, while embodied agents perform increasingly sophisticated physical tasks on the factory floor[11]. An illustrative example is an AI agent that autonomously manages shop floor disruptions and in cases like machine downtime, the AI agent proactively reschedules production and orchestrates material flow in real time. This system delivers contextual insights to supervisors in natural language and facilitates swift, informed responses, reinforcing trust between human and machine agents.
- **Human-AI Collaboration Systems**: As AI systems take over routine, deterministic tasks, human roles are shifting toward oversight, exception management, and creative problem-solving. The relationship is increasingly symbiotic: intelligent systems handle complexity and scale, while humans provide contextual judgment, ethical evaluation, and adaptability in unforeseen situations. Also, the interaction between human and AI is becoming a more effective collaboration. XAI, natural language interfaces, and augmented reality tools allow frontline workers to interact with AI systems intuitively. This fosters trust and bridge the digital skill gap by embedding AI into existing workflows[12].

Together, these advances are not only addressing the current limitations of AI adoption but are also laying the groundwork for a new era of intelligent, adaptive, and resilient manufacturing.

### Concluding remarks

AI is poised to redefine manufacturing and value chains, emerging as the fundamental operating system of the next industrial era. It is essential for manufacturers to scale beyond pilots and build a strong digital foundation purpose-built for AI, reducing integration efforts.

Success will also hinge on aligning AI deployment with broader transformation goals and keeping humans at the core of this transformation. The convergence of digitalization, sustainability, and resilience has led to the emergence of a new model for industrial transformation – one that is enabled and orchestrated by AI. Rather than treating them as three separate domains, manufacturers are unifying them into one transformation, a convergence that can be seen as a “triple transformation”. This system effectively creates “self-healing” operations that can anticipate and mitigate shocks before they cascade through the value network[13]. In an era defined by complexity, volatility, and systemic constraints, manufacturers who embrace AI as the enabler of their transformation will be the ones to lead.

### Acknowledgements

For avoidance of doubt, the views expressed are solely those of the authors and do not represent the official position of the authors’ employers.

# Section 2: Key Topics in AI-Enabled Smart Manufacturing

# Streamlining Industrial Big Data Analytics for Smart Manufacturing

**Vibhor Pandhare[1], Soumyabrata Bhattacharjee[2] and Ram Mohril[2]**

[1] Department of Mechanical Engineering, Indian Institute of Technology Bombay, Mumbai, India
[2] Department of Mechanical Engineering, Indian Institute of Technology Indore, Indore, India

E-mail: vibhorpandhare@iitb.ac.in

### Status

Since ancient times, people have recorded their observations. Observations become data when they are stored, processed, and shared through various means. Early tools, such as the Ishango Bone, sufficed for simple data management [1]. As civilisation progressed, so did the data management needs. By 1940, the first data centre appeared at the University of Pennsylvania [2]. The late 2000s saw a surge in internet usage, leading to the rise of 'Big Data' [3]. In 2011, Industry 4.0 introduced Big Data to manufacturing, utilising sensors to create smart, interconnected factories [4]. In 2012, 'Industrial Big Data Analytics' (IBDA) [5] emerged to draw real-time insights and improve manufacturing decisions. Early research explored whether IBDA could be applied to tasks such as alerting operators about anomalies, predicting maintenance needs, automating fault diagnosis, supporting shop-floor decision-making, optimising performance, and recommending process improvements [6].

Today, the need for Industrial Big Data Analytics is evident as ever. 98% of manufacturing organisations struggle to extract actionable insights from vast, varied industrial data [7]. In 2024 itself, unplanned downtime cost the world's largest 500 companies trillions of dollars [8]. For example, around 20% of the unplanned downtime in production lines is due to tool wear-out [9]. Traditionally, industries utilise only 50-80% of a tool's total available life [10], wasting a valuable resource. Additionally, the manufacturing sector accounts for 30% of global energy consumption [11], a figure that may increase with faulty equipment. In machining, it is also challenging to detect deviations in the toolpath during ongoing processes, which increases scrap and hampers product quality. Additional concerns include reconfigurability of the production line and shop floor to address the growing demand for customised products, keeping resources unchanged.

Thus, manufacturing industries are investing heavily in using data-driven decisions to reduce operating costs and carbon footprints while maximising resource utilisation. On these lines, advancements in IBDA are required to maximise the return on investment (ROI) in smart manufacturing (SM). For example, new frameworks are needed to quickly process a large stream of heterogeneous data for real-time anomaly detection and automated quality control. IBDA may also help optimise toolpaths and machine parameters, reducing scrap and improving quality. Monitoring tool and equipment conditions may minimise unplanned downtimes through predictive as well as prescriptive maintenance. This can further reduce the carbon footprint and energy usage in manufacturing value chains. IBDA can enable dynamic reconfiguration of the production processes, with limited resources, to cater to the growing demand for customised products.

**Current and future challenges**

Given the crucial role IBDA can play in Smart Manufacturing, challenges for these advancements are multidimensional. Specifically, each element of IBDA brings their own set of challenges, as presented below:

- **Industrial**: When it comes to industry-related challenges, privacy is one of the prominent concerns in safeguarding intellectual property (IP) amid rising cyber threats [12], as storing and sharing sensitive data risks breaches and unauthorised access, necessitating robust security measures. Heterogeneous data from diverse sources creates integration issues due to varying formats and units [13]. Equipment of the same type and state often generates inconsistent data patterns [14]. Limited failure events in industries lead to imbalanced datasets, lacking sufficient failure data for effective modelling [15]. Legacy machines, with outdated interfaces, are complex to integrate [16]. Licensing restrictions limit sensor integration, hindering comprehensive data collection and analysis for optimising industrial processes. In such cases, even if remote sensors and cameras are deployed, ambient conditions like humidity, temperature, and light hinder their effectiveness. Mobile industrial robots struggle to establish stable reference points, restricting mapping ability in dynamic environments, making human-robot collaboration (HRC) risky [17].
- **Big**: The ubiquitous and indispensable large volume of heterogeneous industrial data demands robust storage and computational infrastructure. Also, protocols are needed to optimise network redundancy and latency for real-time IBDA on data coming at high velocity.
- **Data**: Variation of not only data formats, but also units of the data of the same parameter, complicates integration and analysis, demanding sophisticated standardisation techniques. Data ownership disputes arise when multiple stakeholders, such as manufacturers and third-party vendors, claim rights, leading to legal and ethical dilemmas. Also, ensuring data veracity becomes a challenge when its volume, variety and velocity are high.

- **Analytics**: One of the critical challenges in IBDA is feature learning from noisy datasets, as irrelevant or corrupted data can obscure meaningful patterns, requiring advanced filtering and preprocessing techniques. Using open-source tools raises intellectual property and security concerns, complicating industry adoption [18]. Explainability remains a challenge, as complex models, such as deep learning, often lack transparency, which hinders trust and compliance with regulations. Verification, validation, and uncertainty quantification (VVUQ) are essential for fostering trust in the displayed recommendation, yet difficult, as ensuring model accuracy and quantifying uncertainties in dynamic industrial environments demands rigorous methodologies. These challenges hinder the development of reliable, scalable, and trustworthy analytics, necessitating innovative solutions to advance IBDA.

### Advances in science and technology to meet challenges

To address these challenges, systematic and streamlined advancements are needed in science and technology, such as seamless interoperability and context-aware adaptability of computational models. Privacy-preserving techniques for federated learning also need to evolve, with scalable algorithms tackling heterogeneous data and robust defences against adversarial attacks. Enabling cross-company collaboration and edge-optimised frameworks may shrink communication delays for real-time model synchronisation. While transfer learning may adapt models built on open-source datasets to industry-specific manufacturing data, it is essential to safeguard the organisation's intellectual property.

Large language models may be retrained on industry-specific data to respond to queries in manufacturing jargon. This may lead to the development of Industrial-GPT, which provides uncertainty-quantified recommendations from heterogeneous data, making them easily communicable to human operators and preventing decision paralysis [19]. However, when developing Industrial-GPT, utmost care must be taken to protect the organisation's intellectual property. The simultaneous tackling of multiple challenges through advanced research directions for streamlining Industrial Big Data Analytics for Smart Manufacturing is highlighted in Figure 1.

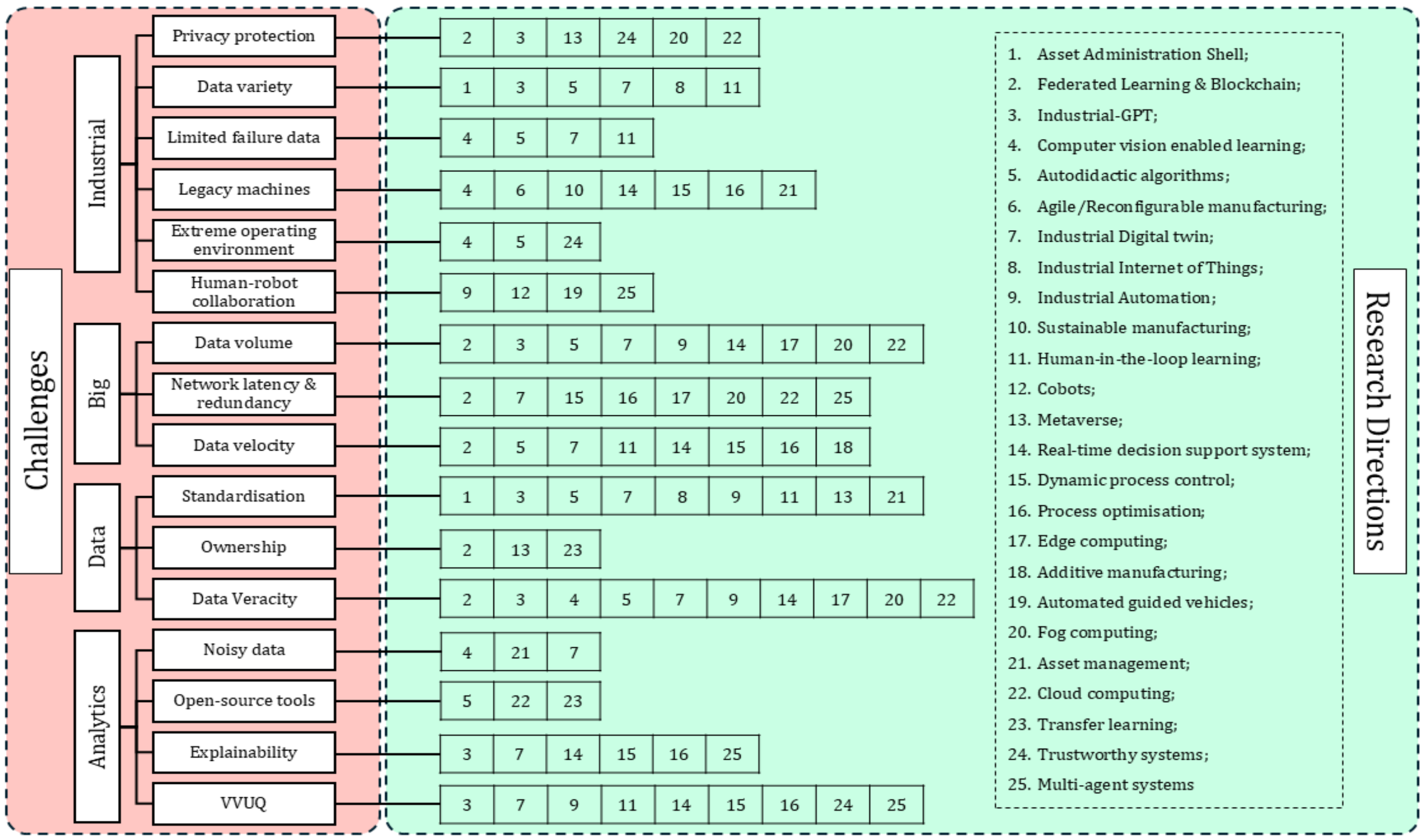

**Figure 1.** Research direction to address the challenges for streamlining IDBA for Smart Manufacturing.

Computer vision systems may also be developed to the point where they can operate effectively in extreme environments, including varying light conditions and occlusions, without hindrance. This enables them to learn the process behaviour independently, thereby reducing dependence on data labelling. Moreover, in industry, often the data generated by systems is unlabelled. In such cases, industries may adopt autodidactic digital twins (DTs), capable of real-time, unsupervised, uncertainty-quantified, and explainable decision-making. These DTs may dynamically monitor and optimise process parameters using IBDA to enhance productivity, reduce energy usage, and lower the carbon footprint for sustainable manufacturing. These DTs could also be lightweight enough to run on edge devices, not just in the cloud. Determining the fidelity level and refresh rate of each such system-level DT may help integrate them to get insights into the entire manufacturing plant, at any given point in time, which may lead to further optimised process parameters for each system. DTs could also be coupled with lightweight physics-based models to improve effectiveness amidst the dynamic nature of manufacturing.

## Concluding remarks

Smart manufacturing, with IBDA at its core, brings significant advantages over traditional practices, for example, lower operational and maintenance costs, a reduced carbon footprint, and improved product quality and resource utilisation. This realisation has led to greater investment in the field. However, maximising ROI requires overcoming specific challenges as highlighted in Figure 1, solving which needs a systematic, simultaneous and multidisciplinary research approach. While addressing all these challenges will take time, innovative solutions may help bridge the gap between legacy and modern manufacturing systems, enabling organisations to begin realising the benefits of smart manufacturing with minimal intervention. It also needs to be noted that preserving the organisation's IP is paramount, irrespective of what technology is developed. Further, no actionable insight can be derived from industrial data by IBDA, unless the veracity of the data is ensured despite its high volume, variety and velocity. Addressing these challenges could help streamline the integration of the vast stream of heterogeneous industrial data, while protecting its intellectual property (IP), to derive actionable insights. This approach would enable the organisation to minimise costs and maximise productivity.

## Acknowledgements

This work is supported by IITI Research Grants: IITI/YFRSG/2023-24/Phase-III/07 and IITI/ YFRSG-Dream Lab/2023-24/Phase-I/03.

# Advanced Sensing, Perception, and Analytics for Manufacturing

**Lingbao Kong*, Qiyuan Wang and Xinlan Tang**

Future Information Innovative College, Fudan University, Shanghai, China

*E-mail: LKong@fudan.edu.cn

### Status

Against the backdrop of the ongoing wave of Industry 5.0, intelligent manufacturing has emerged as a cutting-edge focal point within the realm of industrial manufacturing [1]. In this context, sensing technology, serving as the pivotal bridge linking the physical world to digital signal systems, is undergoing a profound transformation from traditional to intelligent paradigms. In the current era dominated by multi-domain manufacturing, traditional unimodal sensing technologies face significant limitations due to their single information dimension, weak anti-interference capability, and high calibration and maintenance costs. Correspondingly, as shown in Fig. 1, multimodal sensing technologies, which offer rich information, robust redundancy for anti-interference, and low calibration and maintenance expenses, are gradually displacing unimodal sensing technologies across a variety of complex or dynamic scenarios. This transition effectively circumvents the challenges encountered by unimodal sensing in new industrial environments [2], while being better aligned with the urgent demands of modern advanced manufacturing.

In comparison to unimodal sensing, multimodal sensing technology has achieved significant breakthroughs primarily in three key aspects. The first lies in the comprehensive enhancement of perceptual capabilities. By leveraging diverse sensing detectors, multimodal sensing facilitates cross-modal information complementarity. For instance, as Fig. 1, in the inspection of surface scratches or coating defects on automotive components, integrating multiple sensing modalities such as industrial cameras, 3D laser scanners, and infrared thermal imagers can elevate the defect detection rate from 90% to 99.5%, while concurrently reducing the false alarm rate by 60% [3]. The second breakthrough is the marked improvement in robustness and anti-interference capabilities. In robotic object grasping scenarios, even when visual occlusion or blind spots occur, robots can dynamically adjust their actions through tactile and force feedback [4]. This adaptability ensures task continuity and accuracy despite environmental perturbations.

The third aspect centers on innovations in intelligence and system integration. On one hand, cross-modal semantic alignment is realized through multimodal deep learning, with the incorporation of self-supervised mechanisms to reduce reliance on labeled data. For example, in video data processing, visual, auditory, and motion information are automatically correlated, thereby augmenting the model's generalization capacity [5]. On the other hand, edge computing is employed for real-time processing in system integration, mitigating dependence on cloud-based infrastructure. In intelligent logistics, leveraging AGV (Automated Guided Vehicle) navigation and obstacle avoidance technologies, obstacle response times can be minimized to as low as 100 milliseconds [6].

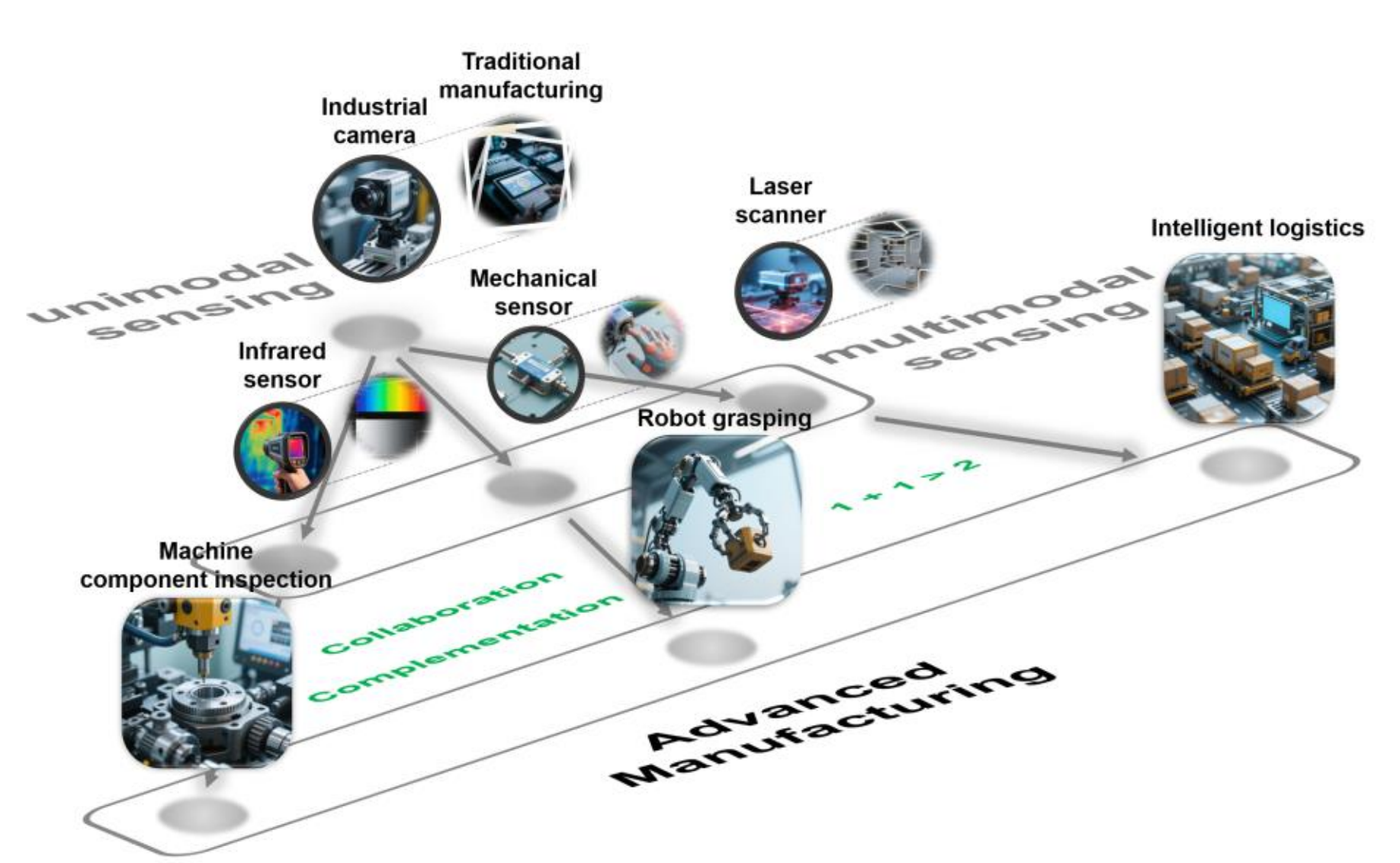

**Figure 1.** Manufacturing paradigms centred exclusively on traditional unimodal technology have become increasingly incompatible with the development of modern advanced manufacturing. Correspondingly, multimodal sensing technologies, by leveraging multidimensional information coordination and complementary data fusion, frequently generate synergistic outcomes that exceed the cumulative performance of isolated modalities (i.e., achieving "1+1>2" effects). These technologies are progressively displacing unimodal approaches in complex industrial scenarios and have emerged as a defining characteristic of contemporary advanced manufacturing systems.

## Current and future challenges

Despite the immense potential that multimodal sensing technology has demonstrated in the realm of intelligent manufacturing, its development continues to grapple with a myriad of challenges.

At the hardware level, the most prominent hurdle lies in the significant disparities in the physical characteristics of diverse sensors, which greatly complicate the seamless integration and alignment of hardware components. For instance, in smart logistics applications, the high-power consumption of LiDAR (Light Detection and Ranging) sensors stands in stark contrast to the low power requirements of cameras, necessitating the design of intricate circuitry and sophisticated cooling solutions. Additionally, the exorbitant production costs pose another formidable obstacle. In predictive maintenance scenarios, the deployment cost of a single sensor can exceed 2 million units of currency, and substantial resources must be further allocated for AI model training and maintenance to analyze the subsequent data.

Moving on to the data and algorithm domain, cross-modal data fusion presents a labyrinth of difficulties. These challenges encompass both semantic alignment issues across different sensors—such as in automated welding processes, where data from arc sensors, high-speed cameras, infrared thermometers, and acoustic emission sensors vary widely in terms of data types and physical significance [7],and the integration of multi-protocol heterogeneous networks. For example, the efficient fusion of 5G networks with existing industrial buses in the context of the Industrial Internet of Things remains an elusive goal [8]. Moreover, the computational complexity and real-time requirements are exceedingly demanding. Take the automatic loading task as an illustration, where the fusion of LiDAR point cloud data (points per frame) with 4K-resolution camera images must be accomplished within a stringent 100-millisecond timeframe [9], placing an enormous strain on computational resources.

Lastly, the integration of human-machine interaction also presents considerable difficulties. The new paradigms of Industry 5.0 advocate for a people-centric approach in industrial manufacturing. However, in current industrial manufacturing workflows, workers are often required to engage in complex programming tasks to adjust robotic operations. Consequently, there is a pressing need to enhance the naturalness and adaptability of human-machine interaction, as well as to improve real-time perception and decision-making capabilities in dynamic environments, all while ensuring a heightened level of safety.

**Advances in science and technology to meet challenges**

To address the aforementioned challenges, multimodal sensing technology must achieve breakthroughs across multiple fronts. The integration of Digital Twin technology offers a viable remedy for the exorbitant hardware costs. By leveraging multimodal sensor data to drive virtual factory simulations, this technology enables the optimization of production strategies, ultimately achieving cost reduction and efficiency enhancement [10].

The continued advancement of edge intelligence and self-learning systems provides effective solutions to mitigate the complexities of cross-modal data fusion and computational demands. Edge intelligence, through the integration of AI chips, facilitates real-time fusion of multimodal data at the terminal device level. This not only meets stringent real-time requirements but also ensures the effective integration of data [11]. On the other hand, self-learning systems employ reinforcement learning to dynamically optimize the weight allocation among multiple sensors, thereby reducing computational complexity, enhancing data reliability, and minimizing redundant data—ultimately lowering the computational burden [12].

Furthermore, human-robot collaborative monitoring technology epitomizes the people-centric ethos of the new industrial paradigm. By utilizing the Kalman filter for spatial alignment, this technology constructs dynamic safety zones that track workers' hand positions in real time. As a result, it curtails the incidence of human-robot collaborative accidents and boosts production efficiency. The development of these technologies not only compensates for the current shortcomings of multimodal sensing in practical applications but also propels mechanical manufacturing technology towards greater efficiency, intelligence, and harmony [13].

Looking ahead, intelligent manufacturing technology is poised for deeper integration with Digital Twin, edge intelligence, self-learning systems, and human-robot collaborative monitoring. This fusion will drive manufacturing systems towards the aspirational goals of "zero defects," "self-awareness," and a "people-first" approach, marking a significant leap forward in the evolution of manufacturing paradigms.

**Concluding remarks**

Multimodal sensing technology has propelled the evolution of traditional unimodal sensing approaches towards greater efficiency and intelligence. By integrating a diverse array of information acquisition modalities, enhancing anti-interference capabilities, and pioneering intelligent, integrated systems, this technology has effectively shattered the robustness barriers in manufacturing environments. It has not only bolstered the standardization of manufacturing processes and the precision of defect detection but also furnished manufacturing systems with high-fidelity, highly reliable data foundations. Consequently, multimodal sensing technology has emerged as the cornerstone of intelligent perception within the Intelligent Manufacturing ecosystem.

Concurrently, the infusion of Digital Twin technology, edge integration architectures, and self-learning systems has further catalyzed the intelligent and miniaturized trajectory of multimodal sensing technology, laying indispensable infrastructure groundwork for contemporary industry. Beyond these technological advancements, the profound integration of human-centric principles stands as a pivotal milestone in the development of multimodal sensing technology. By anchoring human-centricity at the heart of intelligent manufacturing, this paradigm shift fortifies the foundation for IM to align seamlessly with the prevailing ethos of the times, ensuring that technological progress remains intrinsically linked to human needs and aspirations.

### Acknowledgements

The authors would like to express their sincere thanks for the support from National Natural Science Foundation of China (52375414), and Shanghai Science & Technology Committee Innovation Grant (23ZR1404200).

# AI-Enabled Autonomous Manufacturing

**Sungjong Kim[1], Chan Hee Park[2] and Byeng D. Youn[1,3,*]**

[1] Department of Mechanical Engineering, Seoul National University, Seoul 08826, Republic of Korea
[2] Department of Mechanical and Information Engineering, University of Seoul, Seoul 02556, Republic of Korea
[3] Onepredict Corp., Seoul 06105, Republic of Korea
[*] Corresponding author

E-mail: bdyoun@snu.ac.kr

### Status

Automated manufacturing refers to the use of control systems, machinery, and information technologies to execute predefined production tasks with minimal human intervention. While such systems have contributed significantly to productivity gains, they still rely heavily on rule-based logic or expert knowledge, which limits their adaptability to dynamic environments and complex manufacturing tasks. Recently, the combination of a declining skilled experts, rising wages and energy costs, and growing demand for high-mix, low-volume production has highlighted the need for transformative innovation in manufacturing systems.

Autonomous manufacturing represents an evolutionary step forward. It refers to cyber-physical production systems wherein machines, software agents, and embedded systems independently perform sensing, reasoning, and action using distributed intelligence eliminating the need for human oversight in both routine and unstructured scenarios [1]. By digitizing domain expertise and leveraging large-scale process data, autonomous systems provide scalable and adaptive alternatives. Recent advances in artificial intelligence (AI) have enabled these systems to autonomously incorporate real-time feedback, allowing for predictive quality assurance, anomaly detection, and self-optimization of process parameters.

The implementation of AI-driven autonomy has been shown to significantly enhance operational efficiency, reduce overhead costs, and improve system resilience—particularly in globally distributed manufacturing environments where access to expert knowledge is limited. Empirical evidence highlighted the effectiveness of such technologies; for example, the implementation of an autonomous quality management system in the automotive manufacturing sector resulted in a 52% reduction in production costs and a 78% decrease in inspection expenses [2]. Moreover, autonomous manufacturing technologies are expected to exhibit broad applicability across diverse operational domains, including quality control, logistics, energy management, equipment maintenance, and comprehensive process optimization.

### Current and future challenges

Achieving truly AI-enabled autonomous manufacturing requires seamless integration of three foundational components—sensing, reasoning, and action—while also establishing a robust platform for managing the integrated autonomous manufacturing system.

In the sensing stage, manufacturing systems must establish robust and scalable data pipelines capable of reliably extracting, pre-processing, storing, and managing diverse multimodal sensor data. Despite the abundance of available data, current pipeline architectures are often underdeveloped compared to the overall maturity of production systems. These pipelines are frequently designed without sufficient consideration for downstream reasoning and control tasks. Consequently, the acquired data suffers from data availability issues—such as noise, low resolution, inconsistent sampling, an excessive amount of data and poor synchronization with system context—which hinders the systems' ability to transmit only relevant, high-quality data necessary for real-time decision-making and autonomous operation [3].

The reasoning stage involves deriving actionable insights support process-level decisions. At this stage, two central challenges arise: ensuring the interpretability and generalization of AI models. For AI systems to contribute effectively to manufacturing operations, they must provide structured information across key categories, including current and predicted system states (system assessment), identified operational tasks (problem definition), causal factors (root cause diagnosis), and prescriptive recommendations (decision-making). However, many AI models operate as "black boxes," hindering engineers' ability to verify or trust the inferred outputs. Generalization also remains problematic, as models often struggle to maintain robust performance under domain shifts, such as variations in operating conditions, product configurations, or factory environments, leading to physically inconsistent or non-representative results [4].

The action stage requires translating reasoning outputs into executable operations, such as control commands, optimal setpoint selection, or human-readable decision reports. Despite recent progress in AI, current AI models often produce outputs in abstract or model-centric forms that lack the semantic clarity necessary for effective interpretation and implementation within manufacturing systems. Without additional contextualization, these outputs are not readily actionable, requiring engineers to manually interpret the reasoning results and determine appropriate interventions, thereby increasing cognitive burden and delaying operational response [5].

Lastly, current platforms such as manufacturing execution system (MES), and programmable logic controller (PLC) are hierarchical and lack the flexibility to support autonomous manufacturing operations. Key challenges include poor interoperability across distributed manufacturing components, limited support for real-time self-organization and manufacturing lifecycle integration.

**Advances in science and technology to meet challenges**

In the sensing stage, data pipelines integrated with extract-transform-load (ETL) mechanisms are employed to convert raw signals into structured, analysis-ready formats [6]. Virtual sensing techniques are utilized to estimate difficult-to-measure variables by leveraging data acquired from the manufacturing process [7]. To enhance data quality and contextual fidelity, pre-processing methods such as noise removal, sampling rate alignment, and synchronization of heterogeneous data sources are applied [8]. Additionally, ontology-based technologies have been developed to define the identity of collected data and establish contextual relationships among correlated information [9]. By enabling context-aware data linkage and semantic interpretation, it facilitates data filtering and selection in subsequent stages, despite the abundance and heterogeneity of manufacturing data.

In the reasoning stage, interpretability has been advanced through explainable AI techniques, including pre-modelling strategies such as domain-informed feature extraction, as well as post-modelling tools such as attention mechanism analysis, Shapley additive explanations (SHAP) [10]. To improve generalization under domain shifts, lifecycle-aware learning strategies are employed to support data drift detection and continual learning [11]. Furthermore, efforts to integrate physical constraints into AI architectures—through physics-inspired components (e.g., wavelet kernels) and regularization techniques based on governing equations (e.g., differential constraints)—help ensure physical consistency and reliability across diverse operational settings [12,13].

In the action stage, the primary objective is to translate AI outputs into actionable manufacturing decisions. Natural language interfaces powered by large language models (LLMs) enable the summarization and structuring of outputs into human-readable formats, thereby enhancing interpretability and operational readiness[14]. This requires aligning linguistic representations with manufacturing data to ensure contextual relevance and facilitating the integration of domain expertise through instruction tuning and agent-based LLMs [15,16]. Reinforcement learning (RL)-based optimization methods, including proximal policy optimization (PPO) and deep-Q-networks (DQN), are employed to derive adaptive control strategies from reasoning outputs, allowing systems to respond effectively to dynamic operational conditions[17,18]. Additionally, machine learning operations (MLOps) frameworks support the continuity and reliability of AI-driven actions through version control, performance monitoring, and feedback-based retraining, ensuring sustained robustness across the system lifecycle [19].

Building on these advances, a decentralized autonomous manufacturing (DAM) platform architecture was introduced to enable autonomous decision-making, decentralized control, and self-organizing production capabilities [20]. By utilizing multi-agent systems and secure communication protocols, the platform allows distributed manufacturing nodes to collaborate effectively, respond to disruptions, and execute manufacturing tasks without centralized coordination.

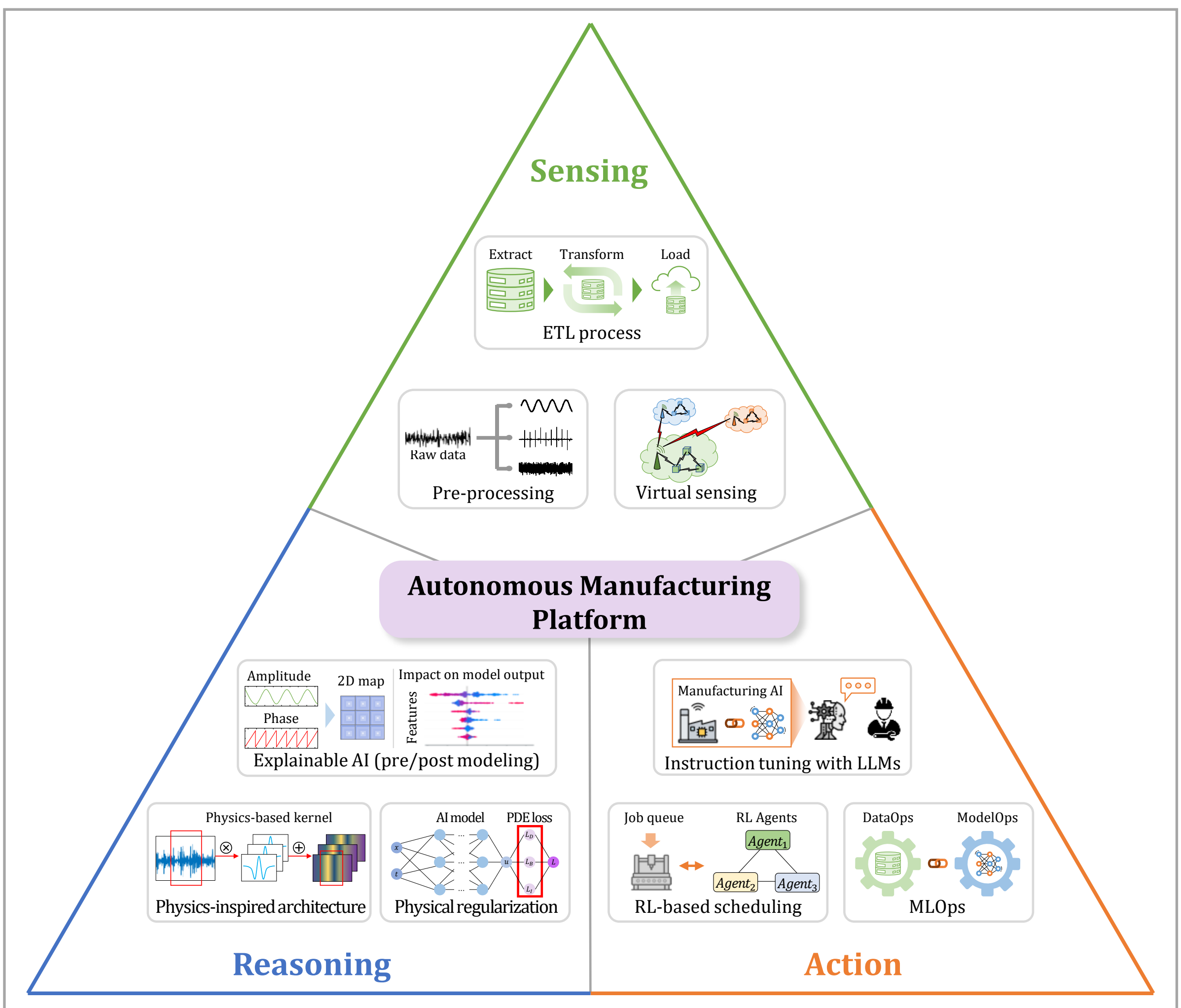

**Figure 1.** State-of-the-art research landscape in the three foundational components of AI-enabled autonomous manufacturing—sensing, reasoning, and action—supported by an integrated autonomous manufacturing platform.

## Concluding remarks

AI-enabled autonomous manufacturing is poised to redefine industrial operations by embedding distributed intelligence across the sensing, reasoning, and action layers of production systems. Moving beyond traditional rule-based automation, autonomous systems leverage advanced AI models to make context-aware decisions, adapt to dynamic environments, and self-optimize processes with minimal human intervention. This paradigm shift is increasingly critical in light of global challenges such as declining skilled experts, escalating operational costs, and rising demand for agile, high-mix production.

Recent technological advances collectively enable machines and software agents to autonomously perceive, interpret, and act within complex manufacturing settings. Moreover, the emergence of decentralized autonomous manufacturing platforms offers a resilient and scalable infrastructure for self-organizing production systems. By integrating multi-agent systems with secure, scalable communication, these platforms allow distributed manufacturing nodes to collaborate effectively, respond to disruptions in real time, and execute tasks autonomously—without relying on centralized control.

The anticipated benefits are far-reaching, encompassing predictive quality control, intelligent maintenance, energy optimization, and logistics coordination. As AI models become increasingly interpretable, robust, and contextually aware, autonomous manufacturing systems are expected to form the backbone of next-generation smart factories—capable of operating efficiently, responding adaptively, and continuously improving under industrial conditions.

## Acknowledgements

This research was partially supported by the International Research & Development Program of the National Research Foundation of Korea (NRF) funded by the Ministry of Science and ICT (No. 2022K1A4A7A04096329), and Technology Innovation Program (or Industrial Strategic Technology Development Program- Automotive Industry Technology Development-Green Car) (RS-2024-00444961, Development and demonstration of Purpose-Built Electric vehicles using design platforms) funded by Ministry of Trade, Industry & Energy (MOTIE, Korea).

# Additive Manufacturing (AM)

**Guo Dong Goh[1], Xi Huang [1] and Wai Yee Yeong [1,2]**

[1] School of Mechanical and Aerospace Engineering, Nanyang Technological University, 50 Nanyang Avenue, Singapore 639798, Singapore
[2] Singapore Centre for 3D Printing, Nanyang Technological University, 50 Nanyang Avenue, Singapore 639798, Singapore
E-mail: wyyeong@ntu.edu.sg

### Status

Additive manufacturing, commonly known as 3D printing, has matured from a prototyping tool into a viable production technology across industries ranging from aerospace and biomedical to electronics and construction. By building objects layer-by-layer, AM enables the fabrication of complex geometries and customized, functionally graded parts with minimal material waste. Ensuring reliability and consistency of printed parts, however, remains a critical concern as defects or process variations can compromise mechanical properties and impede AM's adoption for end-use, safety-critical components (1). In recent years, the convergence of AM with machine learning (ML) has been increasingly viewed as a key enabler for smart manufacturing, addressing these challenges by extracting insights from data and automating decision-making.

Machine learning algorithms excel at recognizing complex patterns in large datasets, and in AM they are being leveraged to unravel the intricate relationships between process parameters, material behavior, and part quality (2). Early successes of ML in AM have been demonstrated across the workflow: in design (e.g., ML-driven topology optimization and generative design for lightweight structures), in materials development (predicting formulations or microstructures to achieve desired properties), in process optimization (tuning print parameters for quality and efficiency), and in in situ monitoring for defect detection (Figure 1).

### Current and future challenges

Despite the enthusiasm, several key challenges must be addressed to fully realize ML's potential in AM. Data acquisition and quality is a foundational hurdle: ML models require large, high-quality datasets, yet AM experiments are time-consuming and sensors can be costly, making data scarce or siloed. Printing conditions vary widely between machines and materials, and there is a lack of standardized data formats and sharing mechanisms across the industry. As a result, models trained on one dataset may struggle to generalize. For instance, an ML model for defect detection might need thousands of labeled images covering different defect types, build geometries, and lighting conditions – data that is often unavailable or expensive to obtain. In metal AM, researchers noted the difficulty of obtaining ground-truth defect data for training computer vision models; high-speed optical cameras capture only surface phenomena, missing subsurface defects, and ex-situ X-ray CT scans are hard to align with the images (3). This highlights a broader sensor and labeling challenge: how to efficiently acquire rich, synchronized data (vision, thermal, acoustic, etc.) and accurate labels (defect locations, material properties) during the 3D printing process.

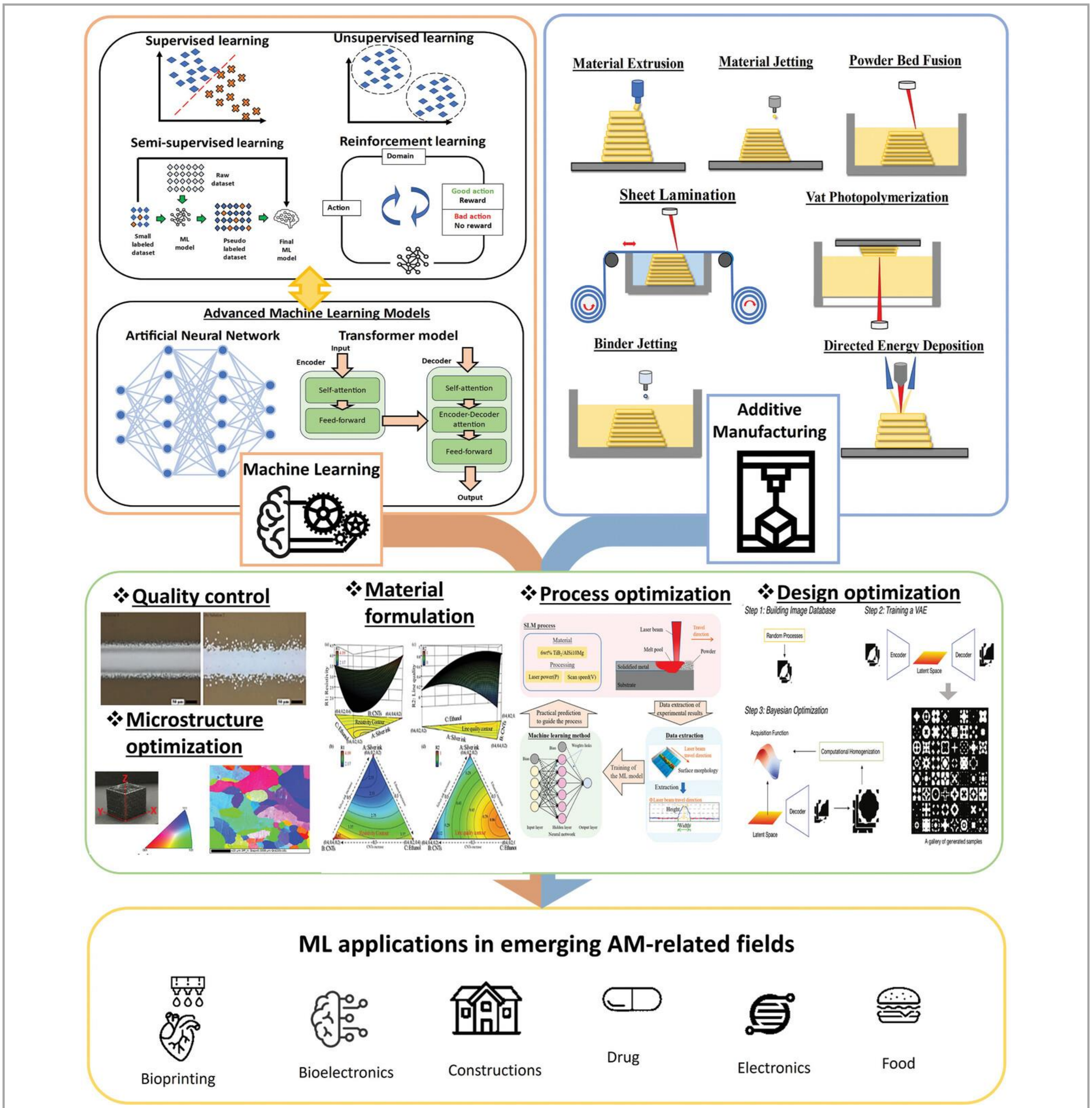

**Figure 1.** The figure provides a summary of how ML is incorporated into AM workflows. On the left, ML approaches are categorized into supervised, unsupervised, semi-supervised, and reinforcement learning, with a note on the rising use of transformer architectures. The right side outlines different AM technologies. At the center, it highlights the advantages ML brings to AM, and along the bottom, it lists practical implementations spanning industries—from aerospace and defense to electronics and food—demonstrating the broad impact of combining ML with advanced manufacturing (2). Reproduced under Creative Commons CC BY license.

Another major challenge is the generalizability of ML models in AM. A model trained for one printer or material often underperforms when applied to a different setup due to variations in machine hardware, calibration, or process dynamics. Adapting an ML-based process optimizer or quality predictor to a new AM machine typically requires laborious data collection and retraining for that specific context. This hampers scalability in production environments where fleets of printers or new machine models are introduced. Techniques like transfer learning and domain adaptation are being explored (4), but ensuring robust cross-machine performance remains non-trivial. Likewise, scalability and real-time implementation pose challenges: embedding ML into the real-time control loop of a printer demands fast inference speeds and reliable hardware/software integration. Many deep learning models are computationally intensive, which could slow down fabrication if not optimized. For example, a complex neural network might detect defects accurately but could become a bottleneck if it cannot run at the printer's frame rate for live monitoring, especially in cases whether high frame rate is required such as melt pool monitoring in powder bed fusion technique (5). Achieving millisecond-level response times may require model compression, edge computing devices, or dedicated accelerators, all of which increase system complexity.

There are also practical deployment challenges. The stochastic nature of some AM processes (e.g., powder bed fusion spatter, filament feed variability) means ML models must handle noisy, high-dimensional data and rare events. Ensuring that models not only detect anomalies but also make reliable corrective decisions without human intervention is a frontier that involves risk: a mistimed or incorrect correction could itself cause a failure. Moreover, the *black-box* nature of many ML algorithms can reduce user trust in critical manufacturing settings. Engineers and certifying agencies may be wary of decisions made by opaque models, highlighting the need for explainable AI and rigorous validation standards (6). Qualification and certification of ML-augmented AM processes is largely uncharted territory – there is a lack of standards on how to approve parts made with ML-driven parameter adjustments or defect correction. Finally, organizational and skill barriers exist; implementing these advanced systems requires interdisciplinary expertise (materials, ML, software) that manufacturing teams are still building.

**Advances in science and technology to meet challenges**

Research efforts are actively advancing the state of the art to address the above challenges, yielding promising results on several fronts. One significant area of progress is in real-time defect detection and correction during printing. For extrusion-based 3D printing, computer vision models have been developed to automatically detect print anomalies such as filament under-extrusion or over-extrusion and intervene mid-build (7). Brion & Pattinson address the need for a truly generalizable error-correction system (1). They built a multi-head neural network trained on 1.2 million automatically labeled images spanning 192 parts, multiple geometries, materials, printers, and toolpaths. By labeling deviations from optimal printing parameters during acquisition, they created a diverse dataset that lets the network detect and correct errors in real time across different extrusion methods. Their control loop not only corrects defects but also provides visualizations of its decision process, enhancing transparency and applicability across varied AM setups.

Another domain of notable progress is multi-objective process optimization using ML, which tackles the challenge of balancing competing quality metrics without exhaustive trial-and-error. Traditional process tuning in AM often involves iterative experiments to achieve a trade-off (e.g., maximizing strength while minimizing porosity). ML-driven surrogate models and optimization algorithms can accelerate this search. Researchers used ML-driven surrogate models to optimize intense pulsed light sintering for aerosol-jet printed nanoink films, balancing film electrical resistance and surface roughness—factors that traditionally trade off (8). Training on a small experimental dataset, their multi-objective algorithm identified process settings yielding both low sheet resistance and low roughness, revealing an optimal window that manual tuning would likely miss. This approach demonstrates how ML can navigate complex AM trade-offs and improve material performance without new hardware. Similarly, Bayesian optimization and reinforcement learning schemes are being explored to tune dozens of AM process parameters simultaneously, accelerating process qualification. For instance, transfer-learning-based frameworks have been able to predict optimal laser processing parameters for new machines using knowledge from prior machines, reducing the effort needed when adopting a new printer model (9). These advances point to a future where "self-optimizing" printers automatically adjust to achieve target outcomes.

In the realm of materials and properties, ML techniques are enabling breakthroughs in achieving application-specific material performance via AM. For instance, researchers trained a neural network on 216 PolyJet-printed samples mixing hard and soft photopolymers to predict Shore hardness and elastic modulus with <1% error—outperforming response surface models (10). By inverting this model, they could specify a desired tissue stiffness and directly obtain the needed material ratios and layer structure. This enables patient-specific anatomical models or prosthetics with tunable tactile properties that trial and error cannot achieve. More broadly, ML is accelerating materials development for AM by identifying complex process–structure–property linkages: for instance, in bioelectronics and bioprinting, where living cells or soft polymers are printed, data-driven models have helped in discovering printable bio-ink formulations and in calibrating process parameters to ensure viability and performance of printed tissues. In electronics printing, ML has been used to predict how printing parameters affect conductivity and to adjust them to produce functional circuits with minimal defects (11). These case studies underscore that by learning from experimental data, ML algorithms can navigate the enormous design space of multi-material and functional printing to meet specific targets.

Researchers are tackling generalizability by combining physics-informed neural networks and digital twin simulations with empirical ML to ground models in physical reality. In metal powder bed fusion, for instance, pore-detection accuracy rose to 87% by augmenting limited experimental data with high-fidelity melt-pool simulations (12). Emerging architectures like transformers are also under exploration for their ability to model sequential, high-dimensional AM data and catch subtle defects (13). Meanwhile, initiatives such as the NIST Additive Manufacturing Material Database are building open benchmarks—compiling build logs, in situ sensor readings, and quality metrics—to spur development of more generalizable AM ML models (14). Together, advances in sensing, data augmentation, algorithm efficiency, and hybrid modeling are transforming AM from a manual, experience-driven practice to a data-driven, adaptive process, building confidence that ML integration will overcome current limitations and unlock higher automation and performance.

In summary, advances in sensing, data augmentation, algorithm efficiency, and hybrid modeling are jointly pushing the boundaries: what was once a manual, experience-driven practice is evolving into a data-driven, adaptive process. With each demonstrated success – from real-time correction systems to predictive material tuning – confidence grows that the integration of ML will resolve many of AM's current limitations and unlock higher levels of automation and performance.

**Concluding remarks**

Machine learning will transform additive manufacturing into a smart, data-driven paradigm. By enabling smarter design, self-optimizing parameters, and autonomous quality control, ML makes production more reliable and efficient. High-quality process data is as essential as hardware for scaling AM to industry. Although challenges remain in data sharing, model transferability, and real-time deployment, ongoing advances—bridging simulation and experiment, standardizing data formats, and developing validation protocols—are paving the way. Future AM systems will continuously learn from each build, reducing errors, improving yield, and expanding design possibilities. Integrating ML with AM thus provides the precision and flexibility needed for agile factories capable of producing complex, customized products with minimal human intervention.

**Acknowledgements**

The authors acknowledge the support of National Research Foundation for NRF Investigatorship Award No.: NRF-NRFI07-2021-0007.

The research is supported by the National Research Foundation, Prime Minister's Office, Singapore under its Medium-Sized Centre funding scheme.

# Machine Learning in Laser-based Manufacturing

**Yung C Shin**

Mechanical Engineering, Purdue University, West Lafayette, Indiana, U.S.A.

E-mail: shin@purdue.edu

**Status**

Machine learning (ML) has been finding increasing adoption in various areas of laser-based manufacturing, such as in predictive modelling, process monitoring, process control, defect detection, prediction of microstructure and mechanical properties, and process parameter optimization. Laser-based manufacturing processes such as laser welding, additive manufacturing and laser cutting involve complex physical mechanisms: including, but not limited to, laser energy absorption, heat transfer, melting, fluid flow, evaporation, solidification, etc. Achieving optimal operating conditions to get the desired mechanical properties and microstructure often involves an extensive amount of experiments with the variation of operating parameters or multi-physics numerical simulations that incur high computational costs and time. As industry is striving to reduce the lead time and the cost of implementing laser processing, machine learning has emerged as a promising approach to establishing data-driven or surrogate models that can significantly reduce the high cost of iteratively finding cause-effect relationships or that can replace the prohibitively computationally expensive physics-based high fidelity modelling in some cases [1,2,3]. In recent years, one can find many examples of using machine learning for process monitoring, particularly with the use of a vision sensor to detect molten pool boundaries [4], surface defects [5], incomplete welds and cuts [6], keyhole depth [7], etc. It has served as a useful tool for automatic process control due to its ability to predict the process condition in real time [8], once developed. Machine learning can also be useful for tuning process parameters or process optimization based on the generated data [9]. It has also been used for predicting the resultant microstructure and hardness after laser processing [3,10]. In addition, some successful efforts have been made to synthesize new materials via machine learning by using additive manufacturing processes. For example, attempts have been made to predict thermodynamically stable phases in high entropy alloys [11,12]. As evidenced by these examples, it is undeniable that the role and use of machine learning will only be increasing as the scientific field of machine learning further advances. In some sense, machining learning might be the only way of realizing predictive science for the optimization, process control and robust implementation of many laser processes in material processing, because the Moore's law indicates that it will take at least another two decades until the computational capabilities, even with massive parallel processing, catch up with the computational speed needed for high fidelity modelling that can be used for real time design, optimization and control.

**Current and future challenges**

Despite the rapidly increasing adoption of machine learning in various applications of laser processing, much of the current machine learning requires an extensive amount of data, which can be very expensive to generate from experiments with physical systems. Furthermore, data-driven models are often applicable only to the specific setup or operation used for the development of the data-driven model, thus lacking the generalization capability to a wide range of process conditions, unlike physics-based predictive models. For example, a data-driven model developed for a particular type of laser and workpiece material may not be readily extendable to another set of laser and material combinations. This will require establishing separate data-driven models for each combination of laser and material. In order to expand its general applicability and reduce the cost of generating a lot of data, more efficient methods of establishing machine learning models would be desirable. For example, physics-informed machine learning would be a promising approach to achieving this goal by integrating well-known physical laws or governing equations that have been developed over the last several decades through extensive scientific research. This will result in a drastic reduction in the amount of data needed to establish a data-driven model and is likely to expand the generalization capability of machine learning models. Another issue lies in how to utilize the existing data, often scattered, albeit abundant. For many of the laser processes for commonly used laser-material combinations, there have been a lot of data generated over the years, but they cannot be easily utilized for constructing a data-driven model since they exist in various formats, sizes, images and resolutions.

Therefore, the community may need to work on establishing the standard for data format or data repositories so that they can be used for developing data-driven models by machine learning. Another challenge is how to combine different types of heterogeneous machine learning models for system-level monitoring, control or optimization. For each laser process, an integrated frame for process monitoring, quantification, and control might be needed. Figure 1 illustrates a possible approach to an integrated quality inspection, process monitoring and feedback control for laser additive manufacturing processes.

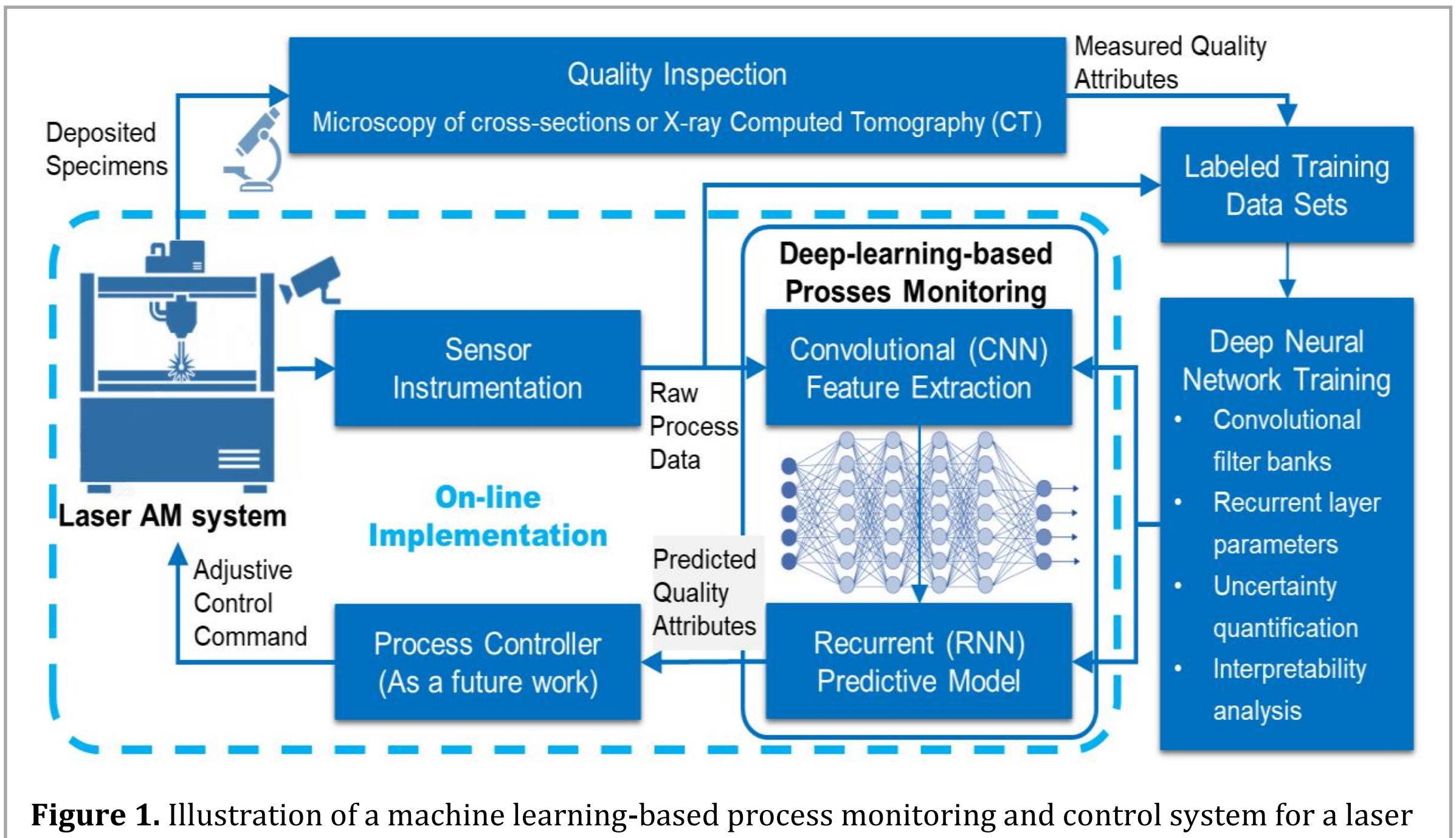

**Figure 1.** Illustration of a machine learning-based process monitoring and control system for a laser additive manufacturing system

**Advances in science and technology to meet challenges**

Many learning methods have been developed over the years, which can be applied to various aspects of laser processing of materials. In finding optimal process parameters, machine learning techniques such as Bayesian optimization, random forests, and various paradigms of artificial neural networks have been utilized. Convolution neural networks, Long Short-Term Memories (LSTMs) and Kalman filters with ML enhancement were often the choices for melt pool monitoring and control. Various convolution neural networks have been popular for the application to defect detections during laser processing with vision systems, x-ray scans, ultrasound scans or scanning electron microscope (SEM) images. People have tried to develop surrogate models of complex physical problems via various neural fuzzy models and physics-informed neural networks. Continuing this success, people need to evaluate a wider range of machine learning models for each application so that the best approaches can be established. The community also needs to work on integrated machine learning models for system level optimization and control. The laser processing community can also piggyback on the rapid advances in artificial intelligence (AI) and machine learning, as more advanced theories and methods are introduced. They also need to pay attention to new types of sensors and sensing techniques that can expand the ML-based process monitoring and diagnostics. Commonly used sensors are cameras, infrared sensors, acoustic emission sensors, photodiodes, spectrometers, etc., while in-situ x-ray devices have also been successfully used for monitoring of molten pool, spattering, etc. These sensors must be easily integrated into commercial laser processing equipment, and provide the requisite speed and resolutions as some of the laser processes, such as laser powderbed fusion and laser welding, are performed at very high speeds.

### Concluding remarks

As described above, machine learning has a very promising future in various laser-based manufacturing processes for process monitoring, control, part quality monitoring and optimization. However, various challenges mentioned in this article must be overcome for a wide use of machine learning in industry, and further advancements in the requisite sensing techniques and sensors must follow. The community needs to work together to establish standards in data formats and repositories so that efforts are not fragmented.

### Acknowledgements

The author wishes to thank many of his former and current students who have contributed to the generation of concepts, advancement in theories and applications of machine learning to many manufacturing processes, which have been used for generating this article.

# Digital Twin in Smart Manufacturing

**He Zhang[1], Zitong Wang[1], and Fei Tao[1,2]****

[1] Digital Twin International Research Center, International Institute for Interdisciplinary and Frontiers, Beihang University, Beijing, China
[2] School of Automation Science and Electrical Engineering, Beihang University, Beijing, China

E-mail: ftao@buaa.edu.cn

### Status

The idea of the digital twin could be traced to the Apollo 13 mission in the 1960s in which multiple simulators were employed to evaluate the failure, train astronauts and mission controllers in response to the oxygen tank explosion. After a period of dormancy, digital twin re-emerged in the 21st century and attracted widespread attention. Prof. Grieves proposed a three-dimensional model of the digital twin and expounded on its value and significance in the full life cycle management of products [1]. NASA listed it as one of the key paths in its future development blueprint [2]. Prof. Tao proposed a five-dimension digital twin model which contains physical objects, virtual models, data, connections, and services to further promote the practice of digital twins [3]. To date, the digital twin has been applied into multiple fields, and the smart manufacturing is one of the most popular fields because it aligns with the core of Industry 4.0, that is, to achieve seamless integration of vertical and horizontal information flows in the supply chain and value chain through digital technology, and to build a highly intelligent production system [4,5]. And the digital twin has been applied into various aspects in smart manufacturing and revolutionized the traditional manufacturing mode.

Although some companies or researchers have carried out the practice of digital twin in smart manufacturing, the current maturity of digital twin application is still not high enough to fully utilize the advantages and value of digital twin due to the limitations of cognitive understanding as well as technology. In addition, in recent years, Industry 5.0, which emphasizes on human-centeredness, sustainability and resilience, has been proposed, putting new requirements on the development of digital twins [6]. And some advanced technologies, such as Large Language Model (LLM), have advanced by leaps and bounds in recent years, bringing new opportunities for digital twin development [7]. In this context, digital twins still need to be further developed to improve its intelligence level, maturity and application scale.

### Current and future challenges

Current and future challenges of digital twin in smart manufacturing contains many aspects, such as application scenarios, key technologies, and security. Current industrial application scenarios of digital twins predominantly focus on real-time condition monitoring, quality prediction, and intelligent control in automobiles, airplanes, ships, and other fields. However, research on digital twins in extreme manufacturing, e.g., microfabrication, ultra-precision manufacturing, and giant-systems manufacturing, is still at a relatively blank stage. And the current level of digital twins is also difficult to handle for extremely complex systems or projects.

Data is a key driver for digital twins in manufacturing [8]. With the development and advancement of sensor and communication technologies, more and more manufacturing process data can be captured [9]. However, on one hand, transient anomalies that may arise during the manufacturing process are still difficult to capture. On the other hand, manufacturing data remains difficult to collect in extreme environments. In addition, the massive data collected from sensors and controllers, combined with that generated by digital twin models, poses a significant challenge for rapid analysis and processing due to limited computing capability

Models are one of the important foundations for realizing digital twins in smart manufacturing. However, the current digital twin models are still constructed as one-off solutions tailored to specific use cases, which limits their generalizability [10]. The requirement for related domain knowledge further limits their application and development. While some scholars have explored the use of purely data-driven digital twins, this approach based on black-box algorithms poses interpretability challenges. Once a problem arises, it is difficult to effectively allocate responsibility.

The implementation of digital twins in smart manufacturing is not possible without the support of related software or platforms. Currently, some companies have developed related tools such as Ansys Twin Builder, Azure Digital Twins and 3DEXPERIENCE. However, there is insufficient compatibility between the different software. The functionality of each piece of software is also insufficient to support the entire digital twin chain in multiple scenarios across different fields [11].

**Advances in science and technology to meet challenges**

To address these challenges, significant scientific and technological innovations are emerging across multiple domains. These advances aim to enhance model accuracy, data interoperability, computational efficiency, and security, which could further enable scalable industrial applications.

The combination of systems engineering thinking and complexity theory with digital twins is a promising approach for applying digital twins to more complex objects and scenarios. Systems engineering frameworks such as Model-Based Systems Engineering (MBSE) contribute to unifying multi-domain models and digital threads, thereby enabling collaborative intelligent manufacturing. Moreover, the complexity science would be helpful for addressing nonlinear dynamics inherent in large-scale industrial systems through multi-scale analysis and complex networks, offering tools to enhance the resilience [12].

To address latency and computational bottlenecks, hybrid edge-cloud architectures are being deployed. The increasing processing power of smart chips at the edge helps to achieve low latency and high real-time data transmission, enhanced data privacy and security, reduced bandwidth consumption, and lower costs [13]. Furthermore, as quantum computing technology develops and matures, it can help form clusters to work together to process larger data sets [14].

Advances in artificial intelligence (AI) are transforming the way of digital twin modelling. Currently, advanced algorithms, such as Physics Informed Neural Networks (PINNs), are integrating domain knowledge more deeply to improve interpretability [15,16]. Besides, generative AI would further enable synthetic data generation to fill gaps in training datasets, enhancing predictive maintenance accuracy. In the future, enabled by generative AI, automatic generation of complex digital twin models based on user requirements is also possible [17]. And blockchain-based traceability solutions are being implemented to mitigate security risks [18].

Standardized frameworks such as ISO 23247 named Automation systems and integration — Digital twin framework for manufacturing are addressing data silos [19]. Tools such as Amazon IoT TwinMaker and Eclipse Ditto enable cross-platform integration through modular APIs and universal asset models. In addition, makeTwin, a unified reference architecture for digital twin software platform, has been proposed [11]. However, the related international standards should be further developed to improve compatibility. Collaboration among all relevant stakeholders is also important for the formation of a digital twin industrial software ecosystem.

Technological advancements, such as AI, edge-cloud collaboration, blockchain, and standardization, are collectively addressing the core challenges of digital twins in smart manufacturing. Continued innovation in quantum computing, explainable AI, and cross-industry collaboration will further accelerate the adoption of digital twin applications.

**Concluding remarks**

Digital twins have become a cornerstone of Industry 4.0 and 5.0, facilitating the integration of physical and virtual systems in smart manufacturing. Despite their superiority in real-time monitoring, predictive analytics, and intelligent control, challenges persist in applications to complex scenarios, such as extreme manufacturing environments, ensuring data integrity, and overcoming computational and interoperability limitations. Collaborative innovation across industries, coupled with robust policy frameworks, will unlock the full potential of digital twins, transforming them from reactive tools into proactive enablers of next-generation industrial intelligence.

### Acknowledgements

The work was supported by Beijing Natural Science Foundation under Grant L243009, National Natural Science Foundation of China under Grant 52120105008, and China Postdoctoral Science Foundation under Grant 2024M754054.

# AI for Smart Supply Chain and Logistics

**Jagjit Singh Srai[1]**

[1] Department of Engineering, University of Cambridge, UK

E-mail: jss46@cam.ac.uk

## Status

Artificial Intelligence (AI), a body of knowledge rather than a single technology, has been decades in development. Whilst we are currently at the foothills of AI technology adoption in supply chain and logistics (SC&L), it promises to be the next major 'Industry 6.0' transformation with the move to cognitive automation [1]. Currently, rapid advancements are being observed in demand forecasting, supply planning, with new physical and digital infrastructures [2] supporting near real-time logistics optimisation but also new business models [3] that enable autonomous operations and hyper-personalisation. These early applications and pilot developments are within a broader digital supply chain transformation that extends and integrates discrete operations across the 'end-to-end' supply chain. Table 1 set out current AI applications within Smart SC&L. These may be classified as '*point-solutions'* in specific areas of the SC such as factory unit operations, and last-mile logistics [4] but also in enabling '*infrastructure*' that supports scaling AI across business enterprises, and most exciting perhaps, multiple connected AI and digital applications that support autonomous '*operating/business models'* [5] involving distributed decision-making through, e.g. Agentic-AI.

**Table 1.** Current status – Examples of AI Applications in Supply Chain and Logistics

| AI Application area | AI Technology Deployed | Enhanced SC outcome |
|---|---|---|
| Demand Forecasting | Machine learning, time-series | Forecast accuracy |
| Last-Mile Delivery | Route optimisation | Enhanced service/less stockouts |
| Warehouse Automation | Robotics-vision systems-robots | Speed, productivity, pick-accuracy |
| Inventory Management | Predictive Analytics | Reduced Inventory |
| Factory unit operations | Machine learning/ Digital Twins | Process and yield optimisation |
| Supplier Management | AI enabled digital platforms | Sourcing flexibility and reliability |

## Current and future challenges

Similar to other technologies deployed in digital SC&L transformation, AI offers huge potential gains, with technology interventions enhancing both productivity and supply chain responsiveness to changing demand. However, there are substantial challenges in the adoption of AI technologies within supply chains, in terms of workforce skills and reluctance to adopt technologies that may impact job security, data quality and data integration challenges, explainability of AI models, potential system biases and governance arrangements for highly distributed systems. Figure 1 summarises AI applications in SCs, emergent challenges and future technology to address the same. In the case of skills, the WEF 2025 Jobs report [6], suggest that one-third of roles by 2030 will involve augmented systems involving human-machine interactions, with a further third fully automated, involving a 50% reduction in manual-only activities from their current levels. This will transform the nature of roles in supply/demand planning with massive reductions in labour/ entry jobs.

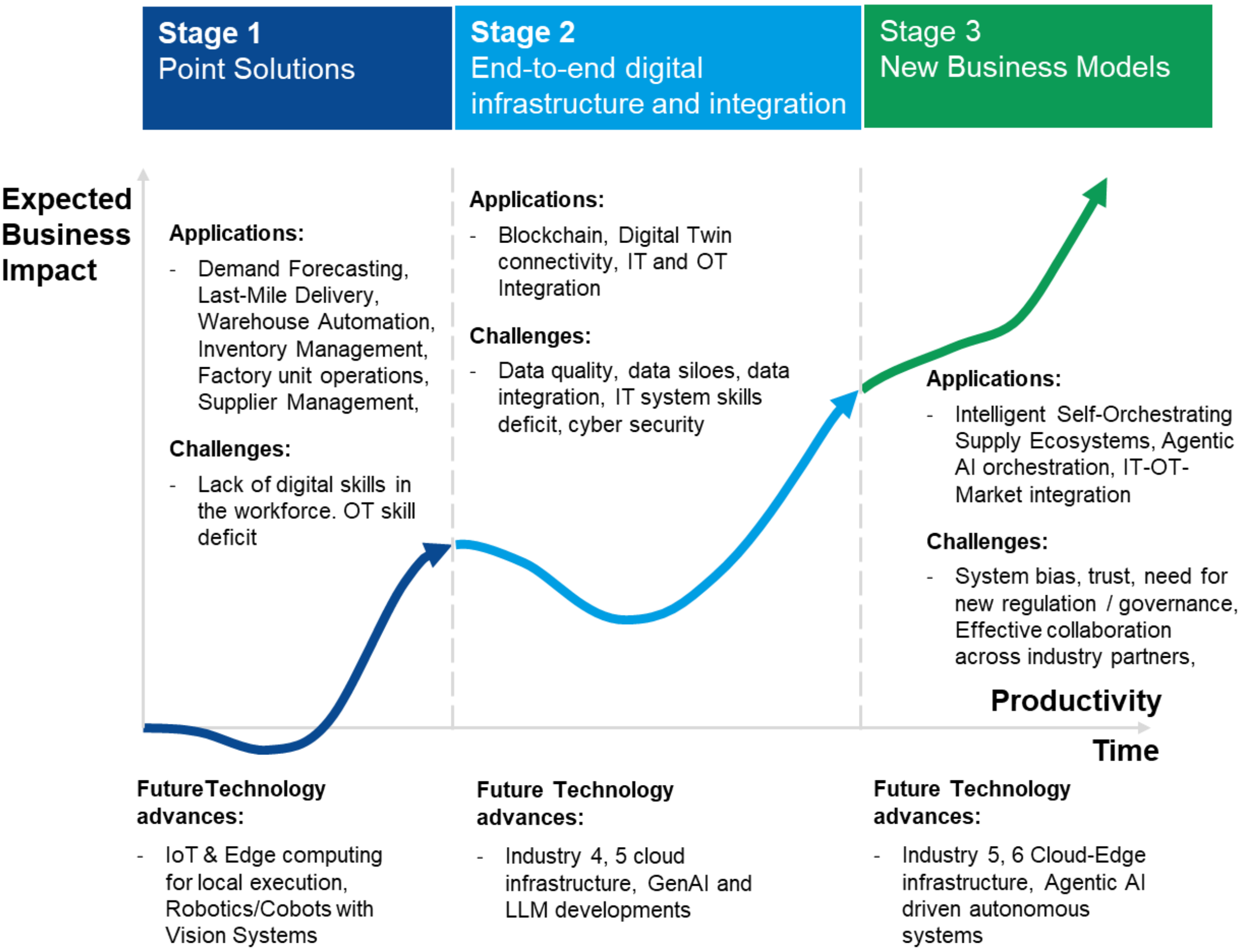

**Figure 1.** AI in SC&L - Applications, Challenges and Future Technology Advances (adapted from Srai et al [5])

Another critical challenge from the adoption of AI in SC and logistics is the issue of attribution of responsibility and accountability as agency is distributed across multiple AI Agents and human actors. This requires multi-actor SC&L collaboration on digital technology adoption [7] and governance mechanisms that limit amplification of bias taking account of interdependence, privacy and system level risks and not just those related to individual/agent decision-making.

## Advances in science and technology to meet challenges

For OM [8] and OR [9] SC&L scholars, AI presents many research challenges but also opportunities to shape its future development. As firms progress beyond single function-specific AI investments, the scaling challenge will require major infrastructure development, with OT and IT professionals collaborating in the data integration activity. The primary challenges are how organisations tackle scaling AI applications across enterprises, and the accountability, data management and privacy issues related to distributed and automated decision-making. To tackle the latter, new regulatory frameworks and governance models will be required, a task most likely to be complicated by SC&L spanning multiple jurisdictions. The development of technologies badged as Industry 5 will see additional human-machine interactions for further productivity gains, that also address material and energy-use efficiency, to address scope 3 Net Zero sustainability challenges. Industry 6 technologies [1] that will underpin intelligent self-orchestrating supply ecosystems, include Agentic AI orchestration supported by data integration across IT-OT-Market platforms., with local Edge computing reducing cloud data transfers and consequent cyber-risks.

## Concluding remarks

AI technologies despite major data, infrastructural and governance challenges are already driving enhanced SC&L performance from improved demand forecasting, near-real time supply re-routing, factory unit-operations productivity gains, and supplier sourcing flexibility. Future developments will start to connect these discrete systems to operate SC&L autonomously. Such developments will see major shifts in the balance between manual, augmented and automated tasks. Unintended consequences, such as changing power within supply networks, systemic risk exposure, hallucinations from model errors, and potential de-skilling through reliance on 'black-box' analysis may lead to trust-deficits and hold back implementation.

**Acknowledgements**

Funding support is acknowledged from digital supply chain transformation research projects; UKRI Resilience in Agrifood Systems: Supply Chain Configuration Analytics Lab (RASCAL) Ref BB/Z516703/1; Made Smarter Innovation Digital Medicines Manufacturing Research Centre, DM2/Ref EP/V062077/1.

# AI-Enhanced Robotics and Autonomous Systems

**Satyandra K. Gupta**[1]

[1] Center for Advanced Manufacturing, University of Southern California, Los Angeles, CA, USA

E-mail: guptask@usc.edu

**Status**

The last decade has seen significant advances in AI techniques such as reinforcement learning, deep learning, large language models, and generative AI [2-6, 10]. These advances are endowing robots and autonomous systems with new capabilities. Most of the AI that we experience in our daily lives is digital AI. Examples include generating a cover letter for a job application, recommendations for watching a movie, creating a painting, and detecting a tumor in a medical image. A different kind of AI is needed to manage the behaviour of robots. For example, a robot performing sanding on an aircraft wing needs AI to operate autonomously. This AI is called physical AI. It is tasked with one or more goals, and it uses sensor data to produce a sequence of actions that the robot executes to achieve the goal. The physical AI monitors task execution using sensors and plans robot actions to perform the task. Physical AI is being used in the following areas related to robots and autonomous systems: (1) perception, (2) planning, (3) control, (4) human robot interaction, (5) learning from human demonstrations, (6) test case generation, and (7) multi robot collaborations.

Figure 1 challenges in realizing AI-powered robotic cells. It also lists advances that are necessary to address these challenges. The risk profile of physical AI applications is often fundamentally different from that of digital AI applications. Risk consists of two aspects: (1) probability of making an error and (2) the consequence of making errors. When the consequence of making an error is not significant, then a higher probability of error can be tolerated. That is why an error probability of 1% is acceptable in many digital AI applications. Conversely, many industrial applications demand errors probabilities better than one in a million. Reducing error probability using a data-driven approach requires using enormous amounts of data. Unfortunately, acquiring data is expensive in industrial applications. Integrating model-based and data-driven approaches is needed to address the data size issue.

Deployment of robotic systems takes a significant amount of human effort due to the time needed to write software and test the system. Increasing complexity of robotic systems is aggravating this problem. Unfortunately, the availability of human expertise can become a bottleneck in robot deployment. Generative AI is emerging as a tool to address this challenge. Digital twins have become a very useful tool for complex physical systems. Increasingly, AI-powered digital twins are being used to support operations of robots and autonomous systems. Finally, AI is creating new modalities for human-robot interactions.

**Current and future challenges**

Digital AI uses a vast amount of data during the train. Collecting high-quality data in many industrial applications takes significant time and incurs prohibitively large costs. Therefore, unfortunately, a purely data-driven AI approach is not a viable model in many industrial applications. We need physical AI to power robotics. Here are two representative use cases to show how physical AI can be used in industrial applications.

- Defect detection is an essential ingredient of robotic manufacturing. Machine learning has emerged as a powerful technique for analyzing and classifying images [11]. However, collecting a large number of images of physical defects needed to train a machine learning system is not possible. An alternative is to develop a pipeline for generating photo-realistic synthetic images. Recent work has demonstrated that a training process that utilizes a combination of photo-realistic synthetic images and real images of defects works well in practice.
- A robotic cell should be capable of building process models for new materials by autonomously conducting experiments [5,14]. While the exact quantitative relationship between the input process parameters and process performance may not be known, often qualitative relationships between many variables are known. We can utilize loss functions during the training phase that penalize deviations from known process constraints. This approach can enforce known models and accelerate the model-building process [9].

A digital twin is a digital counterpart of a real-world system [7,13]. The digital representation used in digital twins is created using data from sensors and Internet of Things devices, and it mimics the physical object or system in real-time. Digital twins are being used to provide information to task planners and schedulers to make decisions about the next tasks to perform based on the current state of the system. Digital twins also monitor the condition and performance of machines and equipment in real-time and use this data to predict when maintenance is needed, reducing unexpected downtime and preventing machine breakdowns. To be useful in the field of robotics, digital twins need to run significantly faster than real-time. AI can be used to power the next generation of digital twins.

Historically, human-robot interfaces in the industrial setting have not been very user friendly. Humans often interact with industrial robots by pressing buttons, turning knobs, and typing on keyboards. These traditional interfaces are hard to master and can be quite frustrating for a new user. Improved human-robot interfaces have potential to change the user experience and improve efficiency of the industrial operations [17,19]. Recent advances in AI are providing new ways for humans to interact with robots.

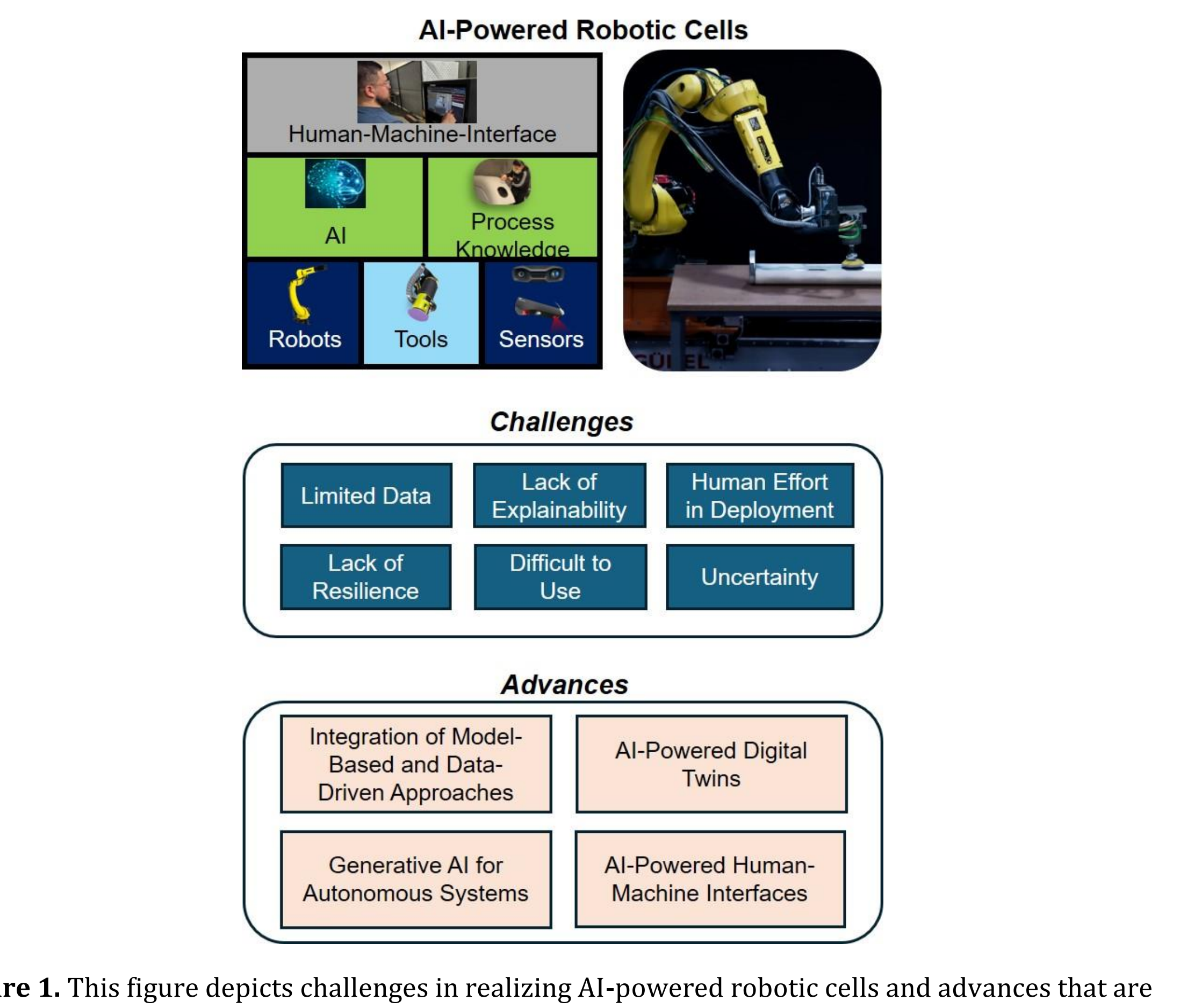

**Figure 1.** This figure depicts challenges in realizing AI-powered robotic cells and advances that are necessary to address these challenges.

### Advances in science and technology to meet challenges

AI is increasingly being used to augment capabilities of digital twin technology and create new capabilities to support the next generation of robotics. Here are a few examples:

- Simulations are necessary to generate optimal plans for finishing operations. Traditional simulations lack the speed required when dealing with part models with uncertainties. Machine learning is being used to create fast simulations based on neural networks, endowing digital twins with new planning and prediction capabilities.
- AI-based prognostics and health management can be used by digital twins to ensure that the onset of adverse events can be automatically detected, and corrective actions can be taken. For example, the digital twin can utilize the force and vision data to determine the cause of rapid tool wear in robotic finishing and take corrective measures to prevent it.

Recent efforts are showing early signs of success in using generative AI in robotics applications to make humans more productive [8,15,16]. The examples below highlight opportunities for using generative AI in the field of robotics.

- Robots often need to perform complex motions to successfully execute a task. Consider the example of sanding where the robot needs to move the sanding tool in a complex motion pattern to produce a scratch-free surface finish. Generative AI now offers the capability to generate code from the text description, which enables humans to communicate with robots in a more natural, time-efficient manner and automatically create robot motion.
- Many applications require robots to perform complex tasks [12]. This requires the top-level task to be decomposed into much simpler subtasks and to determine the sequence of tasks. With the latest advancements in Large Language Models (LLMs) [20] we can pose a query such as, "Provide step-by-step directions to obtain a tool from a locked shelf." and generate a sequence of various subtasks necessary to perform the overall task. Once atomic tasks have been identified, the robot can use a motion planner to generate the motion to execute the task.

AI is revolutionizing human-machine interfaces in the following manner:

- Recent advances in natural language processing and human speech understanding are enabling new modalities for humans to interact with robots [18].
- Sometimes humans might make mistakes and ask the machine to perform an unsafe operation [1]. By monitoring human behaviors and the task state, the machine can predict occurrences of future unsafe situations and alert humans. AI can be used to simulate possible futures and perform risk assessment by accounting for uncertainties.
- Most traditional interfaces were not designed with ease of training in mind. AI-powered interfaces can provide real-time feedback, guidance, and assistance to users during the training phase, helping them navigate complex tasks or troubleshoot problems effectively. Moreover, virtual assistants equipped with AI can offer interactive support and tutorials, improving user productivity and learning outcomes during the training phase.

## Concluding remarks

Physical AI needed in robotics applications cannot be realized as a monolithic system running on the cloud. Physical AI in the context of robotics should be viewed as a complex system that involves interactions among multiple AI components. The system should use the right functional decomposition to ensure that it is able to achieve the desired trade-off in performance and modularity. Many different AI approaches exist. It is unlikely that a single approach will suffice to deliver the desired performance. Therefore, each functional block should use the right AI approach by carefully considering pros and cons. Therefore, having the right system architecture in the physical AI system is the key to success in industrial applications. Generating a large amount of data is not possible in industrial applications from a time and cost perspective. Physical AI should be designed such that it can be trained with limited data generated by physical experiments. An approach that combines model-based and data-driven method is needed to successfully deploy physical AI in industrial applications. Deploying robotic cells in complex applications currently requires significant human effort. The availability of human resources needed to get this accomplished often emerges as a bottleneck and can cause delays in deployment. Generative AI is offering new tools to reduce the human expertise needed to deploy robots in industrial applications. AI-powered digital twins are ushering a new era of smart systems by lowering costs, reducing errors, improving quality, increasing performance, and reducing the environmental footprint. Humans are important parts of industrial operations and therefore human-robot interaction issues need to be proactively addressed during the system design. AI can be used to revolutionize human-robot interfaces by promoting more intuitive interactions for workers.

### Acknowledgements

This work was supported by the Center for Advanced Manufacturing at University of Southern California. I would like to thank my current and former students who contributed to this work. The authors have confirmed that any identifiable participants in this study have given their consent for publication.

# AI-enabled Sustainable Manufacturing

**Byung Gun Joung, Albin John, and John W. Sutherland**

School of Sustainability Engineering and Environmental Engineering, Purdue University, West Lafayette, USA

E-mail: bjoung@purdue.edu

## Status

Artificial Intelligence (AI) is positively transforming manufacturing, and it is envisioned that one key dimension where the application represents a tremendous opportunity is AI for Sustainable Manufacturing, i.e., AI for improved environmental performance. As global concerns over climate change, resource depletion, and environmental impact intensify, manufacturers are beginning to leverage AI technologies to optimize resource efficiency, reduce wastage, and lower carbon emissions. The application of AI to manufacturing can be a key enabler in advancing international sustainability goals such as approaching net-zero emissions and meeting the targets outlined in the UN Sustainable Development Goals (SDGs), while complementing other approaches, e.g., alternative energy adoption [1], energy efficiency improvements [2], and sustainable product design [3]. AI-enabled manufacturing is perhaps the next radical step after digital manufacturing, which seeks to computerize manufacturing. Existing manufacturing technologies, though presently limited in addressing environmental impacts and production variability, can be hyper-optimized through AI to embed environmental intelligence, enhance flexibility and scale to address the demands of a changing world.

With the advancement of IoT technology and computational capabilities, Artificial Intelligence (AI) and Machine Learning (ML) are increasingly being adopted in manufacturing. In addition to their other potential performance benefits, we believe that AI/ML can accelerate the pursuit to "greener" manufacturing, e.g., decarbonization [1]. As an example, AI is being used to improve facility-wide energy efficiency by embedding real-time environmental intelligence, predictive adaptability, and scalable optimization to reduce peak demand and carbon intensity.

AI plays an increasingly important role in reshaping how industries manage resources, reduce waste, and minimize environmental impact. Research on AI for sustainability not only enables data-driven analysis and learning but also calls for the development of new metrics and indicators to effectively evaluate sustainability performance [4]. Currently, AI applications in sustainable manufacturing are concentrated in a few key areas: i) Process Optimization [5]: optimizing (in real-time) resource utilization and process efficiency (e.g., highly variable demand); ii) Process Control and Quality Assurance [6]: vision systems powered by deep learning model are used to detect defects, monitor emissions, and ensure process precision—reducing rework and material waste; iii) Supply Chain Optimization [7], [8]: AI forecasts demand, manages inventories, and optimizes transportation routes, indirectly reducing emissions and resource use.

Despite these successes, widespread adoption of AI in manufacturing is still limited. Many manufacturers—particularly small to medium-sized enterprises (SMEs)—face implementation barriers, e.g., lack of employee expertise, high upfront costs for infrastructure and training, concerns about data privacy, and return on investment. Moreover, many AI implementations are still focused on economic performance. The alignment of AI outcomes with environmental KPIs (e.g., carbon footprint, water use, material efficiency) is still emerging. In parallel, digital twins are becoming a crucial component to manage scalability and adaptability to handle complexity and variability in process design and optimization. These virtual models can simulate various operating conditions, material flows, equipment configurations, and uncertainties associated with real-world deployment—such as fluctuating resource availability, equipment degradation, and process variability—enabling engineers to identify low-carbon and low-waste pathways before physical implementation. For instance, they are used to assess different production scenarios to minimize poor quality products, energy use, and chemical waste. AI can significantly improve digital twins by enhancing real-time data analysis, predictive modeling, and decision-making through advanced machine learning algorithms.

AI may also be used to accelerate life cycle assessment (LCA) workflows by replacing manual inventory analysis with automated estimation based on historical data [9], product specifications [10], and production logs [11]. AI-powered LCA tools can now predict cradle-to-grave environmental impacts for new, complex designs using surrogate models trained on previously assessed products, which can easily be implemented within the design and development process to provide environmental footprint information. In materials engineering, generative models such as variational autoencoders and reinforcement learning are being applied to discover sustainable alternatives—such as bio-based polymers or recyclable alloys—that meet performance constraints while minimizing environmental burdens. These tools significantly reduce the time for R&D and cost to develop materials with less environmental impact. However, most current life cycle indicators rely heavily on predefined emission factors with various uncertainties [12], which aggregate environmental impact per unit of activity (e.g., kg $CO_2$ per kWh). While useful, these factors often lack spatial, temporal, and contextual granularity necessary to forecast real-world behaviors in the realm of sustainable manufacturing. As a result, they overlook site-specific environmental and health hazards associated with certain raw materials—such as toxicity, particulate emissions, heavy metal exposure, endocrine-disrupting properties, and water contamination risks—that may not be reflected in traditional GHG-focused metrics.

Currently, AI in sustainable manufacturing shows great promise, but real-world implementations are isolated. Early adopters are leading the way, but a broader, systemic shift is needed to utilize the full potential of AI for sustainable manufacturing. This chapter explores current and future challenges that may hinder/delay the widespread adoption of AI in sustainability-driven manufacturing, while also identifying the gaps that must be addressed for long-term impact. It then highlights scientific and technological advances that can bridge these gaps, paving the way for transparent, adaptive, and environmentally responsible AI-enabled green manufacturing systems.

### Current and future challenges

Despite their significant potential, AI and ML face structural, technological, and cultural barriers that limit their full-scale implementation in sustainable manufacturing. A primary concern for AI applications related to sustainable manufacturing is securing meaningful, relevant, and accurate data. Clean, labeled, and accessible datasets are critical for effective model training, yet many facilities operate with siloed, inconsistent, or incomplete data. Legacy systems often lack interoperability, making data integration costly and time-consuming. Additionally, concerns over intellectual property and cybersecurity create resistance to open data sharing across supply chains.

The transparency, interpretability, and trustworthiness of AI models are also key issues in ensuring their effective and responsible deployment in various manufacturing applications [13]. Many state-of-the-art AI models (e.g., deep neural networks) operate as "black boxes," making it difficult for engineers and decision-makers to understand or trust their outputs completely. This limits the adoption of cross-domain and multimodal AI for tasks where accountability and traceability are crucial, such as compliance with environmental regulations or safety standards. Also, data heterogeneity and computational costs and infrastructure limitations need to be addressed to fully leverage the potential of AI in sustainable manufacturing.

Another challenge with respect to adopting AI is workforce readiness, as employees need the skills to effectively use the technology [14]. The successful implementation of AI requires not only data scientists and engineers but also skilled operators who can understand how to interpret model outputs and act upon them. Upskilling the workforce for AI-integrated environments has only occurred in a few instances, perhaps due to cost of the training/education . Recent advancement in large language models (LLMs) can support on-the-job training. These models are best utilized in general contexts but will likely struggle with more detailed/highly specialized contexts.

Cross-domain and multimodal AI represents a promising frontier in advancing sustainability within manufacturing. By integrating diverse data types—such as sensor readings, textual documentation, visual inspection of images, and environmental indicators, an AI system can develop a more holistic understanding of complex manufacturing ecosystems. Also, centralized data platforms can play a critical role in coordinating domain-specific knowledge throughout the different phases of sustainability efforts. For instance, combining machine sensor data with maintenance logs and supply chain records can improve fault diagnosis, reduce material waste, and optimize energy usage across the product life cycle. Table 1 shows current application areas along with associated technologies/infrastructure needs that are essential for realizing how AI can be utilized for sustainable manufacturing.

**Table 1.** Manufacturing Application Areas and Associated Infrastructure /Technologies Needs

| Maturity | Application Area | Infrastructure Needs | Technology Needs |
|---|---|---|---|
| Emerging | Circular Economy-Optimization<br>Green Material Discovery | Sustainability Data Hubs (for scalability, interoperability, and ) | Cross-domain and Multimodal AI<br>Agent-based Autonomous AI |
| Low | Quality Assurance<br>Product Design<br>Life Cycle Assessment | Digital Twin<br>Sensing and Actuation Systems | Real-time LCA and TEA<br>Explainability and Trustworthiness of AI<br>AI-enabled Adaptive Manufacturing |
| Medium | Process Control<br>Process Optimization<br>Energy Optimization<br>Supply Chain Optimization<br>Predictive Maintenance | Standardized LCA databases | Broader dissemination of existing technologies across the workforce |

### Advances in science and technology to meet challenges

Recent scientific and technological advances are crucial to overcoming challenges to AI adoption in sustainable manufacturing and such areas as energy, materials, and processes. These developments enhance efficiency, optimize resource use, and enable better monitoring and reduction of environmental impacts across the product life cycle.

One major area of progress is in AI-assisted energy optimization. Machine learning models can now analyze large volumes of sensor and operational data to dynamically control energy consumption in manufacturing systems [15]. Advanced algorithms enable real-time decision-making to reduce energy waste, schedule machinery for off-peak hours, and integrate renewable energy sources into production lines. Additionally, predictive models enhance demand forecasting and energy storage management, making industrial energy use more sustainable and resilient.

In parallel, advances in AI based digital twins and simulation technologies have revolutionized the way manufacturers design, build, operate, and evaluate systems with sustainability in mind [16]. Digital twins, i.e., virtual representations of physical assets, allow engineers to simulate various scenarios to minimize emissions, water use, and material waste before implementation. When combined with AI, these models can adapt to changing conditions and continuously optimize performance throughout a product's life cycle.

Another key development lies in sustainable materials discovery using AI. Machine learning algorithms are accelerating the identification of low-carbon materials [17], recyclable polymers [18], and eco-friendly composites [19] by predicting material properties and behaviors from large experimental datasets. This significantly reduces the dependence on trial-and-error methods traditionally associated with material innovation and speeds up the transition to greener alternatives.

Additionally, progress in AI interpretability and domain-specific modelling is bridging the gap between data science and industrial practice. New methods in explainable AI and physics-informed machine learning enable practitioners to better understand how AI models make decisions and ensure their alignment with engineering principles and sustainability goals [20]. These developments are critical for gaining trust, improving transparency, efficiency and supporting responsible adoption of AI aligned with human interaction in complex manufacturing environment.

Finally, the integration of real-time AI with breakthroughs in energy systems, materials research, process simulation, and interpretability are enabling transformative improvements in green manufacturing. These scientific and technological advances are essential to overcome current challenges and ensure AI becomes a core driver of sustainable industrial development.

**Concluding remarks**

Artificial Intelligence (AI) has emerged as a transformative enabler in the pursuit of sustainability and green manufacturing. Its ability to analyze complex datasets, optimize resource use, and support intelligent decision-making, offers significant opportunities for reducing environmental impact across manufacturing systems. From predictive maintenance and energy-efficient scheduling to sustainable product design and supply chain transparency, AI technologies are driving operational improvements that align with long-term sustainability goals.

However, realizing the full potential of AI in this context requires more than technological readiness. It demands a multidisciplinary approach that combines data science, domain expertise, and sustainability science – in addition to, of course, manufacturing science and engineering. The successful integration of AI into manufacturing must consider not only technical performance but also explainability, data governance, and ethical implications. Additionally, it is essential to ensure that AI solutions are accessible and scalable, especially for small- and medium-sized enterprises (SMEs) that often lack the resources to adopt advanced technologies.

As industries accelerate their transition toward net-zero emissions, AI will play a growing role in enabling adaptive, transparent, and resilient manufacturing systems. The design of highly connected systems across multiple levels and layers in manufacturing can accelerate large-scale integration of AI and unleash its maximum potential. Future research should focus on advancing interpretable and centralized AI systems, integrating real-time LCA with sustainability metrics into decision-making processes, and collaborating across sectors to share knowledge and best practices. With continued innovation and responsible implementation, AI can significantly contribute to reshaping manufacturing systems into engines of sustainable development.

# Section 3: Non-Traditional Machine Learning Techniques for Smart Manufacturing

# Machine Learning and Deep Learning for Manufacturing

**Sang Won Lee**[1]

[1] School of Mechanical Engineering, Sungkyunkwan University, Suwon-si, Republic of Korea

E-mail: sangwonl@skku.edu

### Status

In the field of Smart Manufacturing, the application of Deep Learning (DL) and Machine Learning (ML) technologies has expanded across various areas such as quality control, machine and device maintenance, and process optimization. Recent advances have primarily focused on predictive maintenance, fault diagnosis, and product inspection.

Predictive maintenance has evolved into sophisticated frameworks which integrate LSTM-based DL for time-series data analysis, ML models such as Random Forest and XGBoost, and hybrid framework leveraging digital twins. For instance, in predictive maintenance research for industrial robots, a model combining temporal features extracted by LSTM and KNN classifier has demonstrated improved equipment uptime and extended failure intervals by incorporating knowledge graph-based maintenance strategies. Additionally, in machining centers, IoT-based data collection paired with Random Forest prediction models has shown high accuracy and real-time performance, effectively supporting operator decision-making [1].

In fault diagnosis, domain adaptation techniques have emerged as a key solution to data scarcity, while semi-supervised learning and federated learning address the need to utilize small amounts of labeled data. A representative study combining Active Learning with Semi-Supervised Learning demonstrated significant cost reduction in data labeling while maintaining fault classification accuracy across various industrial datasets [2].

In the domain of product inspection, real-time YOLO-based detection models, weakly supervised object localization networks, and semantic segmentation approaches have been explored. For instance, the CADN framework proposed a model capable of detecting defects using only image-level labels, while a comparative study between YOLOv4 and RCNN demonstrated their effectiveness in detecting defects during automotive spot-welding processes [3].

Attention-based lightweight networks have also reported for defect detection task [4]. By integrating custom-designed architectures on top of large-scale pretrained backbones, these models achieve superior performance in terms of both accuracy (mAP) and inference speed (FPS) on public datasets such as NEU-DET and GC10-DET.

**Current and future challenges**

The most demanding challenge in ML/DL-based smart manufacturing is the lack of high-quality labeled data. Due to security, cost, and operational complexity in industrial environments, collecting sufficient labeled datasets—especially for rare failure scenarios—is extremely difficult. While some studies have attempted to use GANs for synthetic fault data generation and LSTM architectures for time-series modeling [5], these approaches are not always applicable in the manufacturing sites where real failure data cannot be collected.

Secondly, limited generalizability and explainability remain critical issues. For instance, predictive models optimized for specific tools often struggle to generalize across varying machines or operational conditions. To address this, explainable AI (XAI) techniques are being increasingly incorporated to improve the interpretability of AI models [6].

Thirdly, achieving real-time performance and lightweight model deployment is an ongoing technical bottleneck. Especially in Edge-AI environments, resource-constrained environments demand efficient yet accurate models. Recent trends include compressing large-scale models via knowledge distillation and fine-tuning using domain-specific data to balance performance and complexity [7].

Lastly, cultural and institutional barriers must not be overlooked. SMEs often face delays or failures in AI adoption due to limited infrastructure, technical capacity, or upfront investment. In documented failures of Industry 4.0 transitions, factors such as unrealistic managerial expectations, infrastructure deficits, and internal resistance have been key contributors [8].

**Advances in science and technology to meet challenges**

Several technological pathways have been proposed to overcome the challenges as shown in Figure 1:

First, data scarcity and imbalance are addressed using transfer learning, self-supervised learning, GAN-based synthetic data generation, and federated learning. Specifically, domain adaptation has been employed to generalize the performance under limited samples, while GANs have been used to synthesize fault data or exploit unlabeled datasets [9,10,11].

Second, explainable AI (XAI) technologies are increasingly utilized to reduce computational costs during quality inspection while providing interpretable AI predictions for operators. Visualization techniques such as CAM and Grad-CAM help identifying decision rationale, while virtual sensor reconstruction and noise correction methods are developed to enhance data quality and trustworthiness [12,13].

Third, significant efforts have been made toward developing lightweight and real-time ML/DL model architectures. Techniques such as deformable convolution, channel attention, and bidirectional feature fusion have been embedded into modern network designs to achieve inference speeds of 30–100 FPS while maintaining high accuracy, enabling edge deployment in production environments [14,15].

Fourth, the integration of digital twins with hybrid modeling has gained increasing importance. Hybrid frameworks that combine physics-based simulations with sensor-driven data models have shown notable improvements in predicting component lifetimes. These approaches are particularly effective in enhancing prediction accuracy and scalability in real-world operations [16,17].

Lastly, privacy-preserving learning approaches have received considerable attention to address cross-enterprise data sharing concerns. Federated Learning frameworks enable collaborative training of high-performance models without exposing sensitive manufacturing data, and their effectiveness has been demonstrated on real production datasets [18].

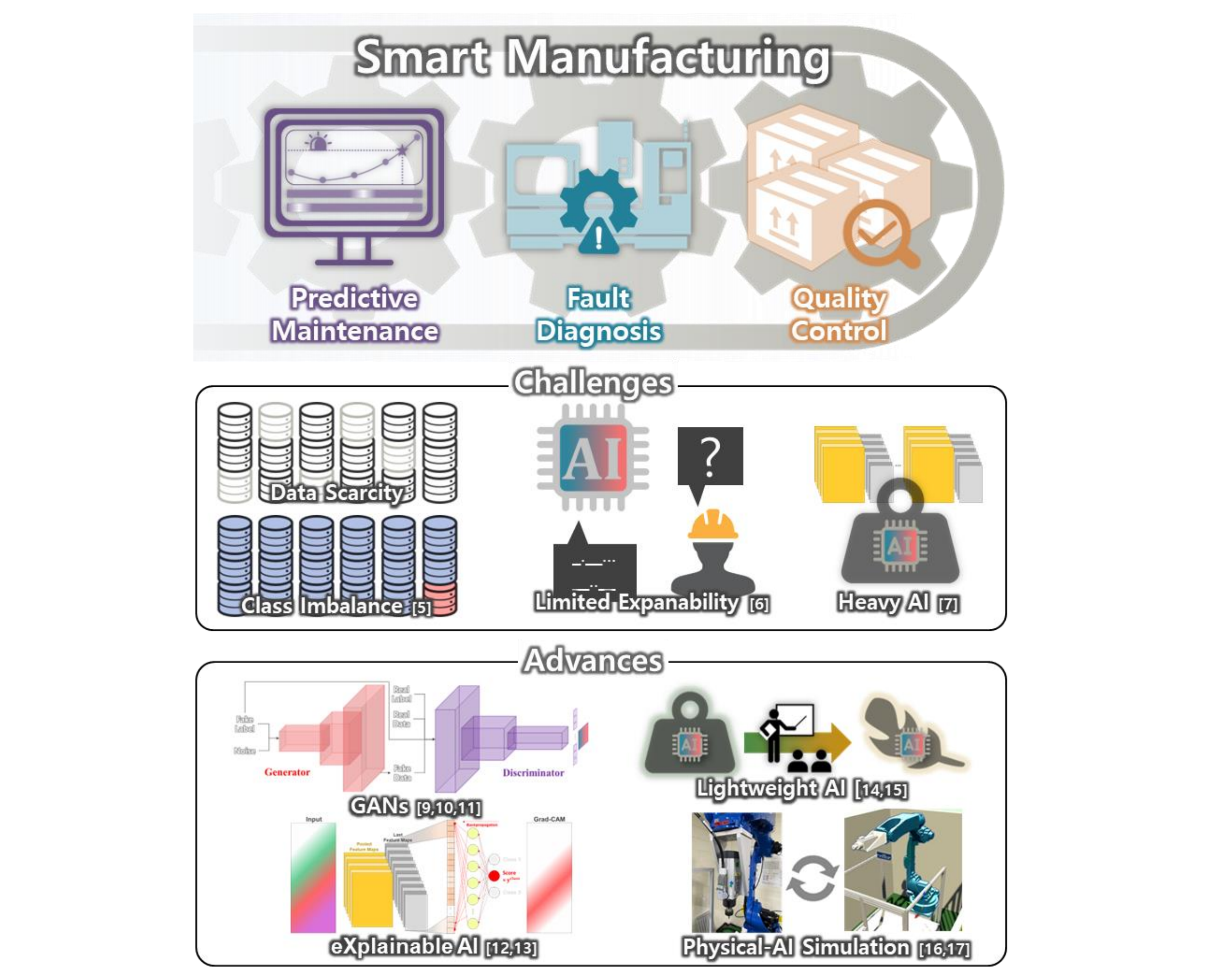

**Figure 1.** Recent challenges and technological advances in ML/DL for smart manufacturing.

**Concluding remarks**

In conclusion, although ML/DL-based smart manufacturing is rapidly advancing and offers significant potential, it continues to face several foundational challenges. These include difficulties in securing labeled high-quality industrial datasets, limitations in generalization across diverse operational scenarios, low interpretability of complex models, and substantial gaps in infrastructure and readiness—especially for SMEs.

To address the challenges, a comprehensive and layered strategy is required. First, enhanced data acquisition pipelines must incorporate not only sensor-based monitoring but also simulation (digital twin)-driven synthetic augmentation. Integration of heterogeneous structured and unstructured data across domains will facilitate the development of more comprehensive AI models.

Human-centered interpretability must be also incorporated from the early design stage. In addition to visualizations for transparency, techniques such as feature importance highlighting, attention heatmaps, and contextual explanations should empower operators to understand and validate AI-based decisions. This fosters trust and improves field-level acceptance.

Furthermore, continuous innovation in lightweight AI model architecture is essential. In edge computing environments, where real-time decision-making is required, new forms of DL/ML methodologies must be developed that balance the trade-off between accuracy and computational efficiency. In particular, there is growing interest in model designs that minimize energy consumption while maintaining high performance, as well as federated fine-tuning techniques that incrementally improve model accuracy using localized field data.

Ultimately, the successful deployment and scaling of AI-enabled manufacturing systems will require coordinated collaboration among academia, industry, and government. When technical innovations are validated through empirical testing and aligned with institutional and workforce strategies, smart manufacturing will evolve beyond automation to become transparent, resilient, and human-aligned intelligent production systems.

### Acknowledgements

This article was supported by the Industry Technology Alchemist Project funded by the Ministry of Trade, Industry & Energy (MOTIE, Korea) (No. 20025702), and National Research Foundation of Korea (NRF) grant funded by the Korea government (MSIT) (No. 2022R1A2C3012900).

# Physics Informed Machine Learning through Inductive Bias

**Olga Fink[1] and Vinay Sharma[1]**

[1] Intelligent Maintenance and Operations Systems, EPFL, Lausanne, Switzerland
E-mail: olga.fink@epfl.ch

### Status

In smart manufacturing, Physics-Informed Machine Learning (PIML) [1] is emerging as a critical enabler for building AI systems that are not only data-efficient but also physically consistent, interpretable, and trustworthy. Manufacturing environments are characterized by complex processes, interconnected systems, and often sparse or noisy sensor data [2] PIML addresses these challenges by embedding domain knowledge—such as governing equations, structural constraints, or physical symmetries—directly into the learning process.

PIML incorporates three types of bias to guide models toward generalizable and physically plausible solutions: inductive, observational, and learning biases. Inductive biases are prior assumptions embedded in the model architecture—such as spatial locality [3], conservation laws [4], or temporal continuity [5]—that constrain the solution space. Observational biases stem from how data is sampled or represented, while learning biases arise from the optimization strategy, such as regularization or specific loss functions.

In smart manufacturing, inductive bias plays a particularly important role due to the complexity of physical processes and the presence of diverse sensor configurations. Knowledge about sensor location (relational), temporal dynamics, and signal characteristics can be effectively incorporated as inductive biases to enhance model generalization, interpretability, and data efficiency [6]. It enables learning from limited data while ensuring model predictions respect known physical and operational constraints. For example, models that embed knowledge of thermal dynamics, material behavior, wear processes, or conservation laws are better suited for predictive maintenance, quality assurance, and control [7][8]. Inductive biases also enhance interpretability—critical for operator trust and deployment in safety-critical settings.

A key subclass of inductive bias includes structural and relational biases [3], which are increasingly important for modeling modern manufacturing systems, where sensor measurements often exhibit strong spatial and temporal correlations [9]. Structural bias encodes assumptions about the hierarchical or modular organization of systems, while relational bias captures interactions and interdependencies among components. These are naturally leveraged by Graph Neural Networks (GNNs), which show strong potential in smart manufacturing.

GNNs can model machines, sensors, and subsystems as nodes in a graph, with edges representing physical, functional, or spatial relationships. This makes them well-suited for tasks such as fault propagation analysis, root-cause diagnosis, and system-level health monitoring. Spatiotemporal GNNs [10] further incorporate time dynamics, enabling real-time analysis of evolving sensor data across distributed systems.

In smart manufacturing, GNNs are increasingly used as surrogate models to approximate the behavior of complex dynamical systems where direct simulations (e.g., finite element or multi-physics models) are computationally expensive [11][12][13]. By learning from system-level sensor data and known component interactions, they emulate physical processes with high fidelity, enabling fast, scalable diagnostics, control, and optimization.

As smart manufacturing evolves toward connected, cyber-physical environments, integrating structural and relational biases—particularly through GNN-based PIML models—will be central to building robust, scalable, and transparent AI-driven decision-making systems.

**Current and future challenges**

Surrogate modeling is a key enabler in smart manufacturing, offering efficient approximations of complex physical and cyber-physical systems. These models are critical for real-time applications such as system diagnostics, control, and predictive maintenance, where full-scale simulations (e.g., finite element or multi-physics models) are computationally prohibitive. Despite their promise, surrogate models face a number of open challenges specific to manufacturing environments.

A primary requirement is fast, online dynamics prediction under changing operating conditions. Models must deliver reliable, real-time outputs even as system loads, speeds, or thermal states vary. Additionally, they must support long trajectory roll-outs —that is, iteratively predicting the system's future state over extended time horizons by feeding model outputs back as inputs—while minimizing error accumulation to ensure predictive stability over time.

Interpretability and explainability of learned dynamics remain critical. Many high-performing models behave as black boxes, making it difficult to verify or explain their outputs—an unacceptable limitation in safety-critical manufacturing contexts. Surrogates must not only be accurate but also transparent and physically meaningful.

Industrial systems are inherently noisy and data-rich. Models must handle noisy, heterogeneous sensor inputs, often with varying sampling rates and resolutions. They must also be capable of learning directly from observed trajectories, even when system parameters or governing equations are unknown, requiring strong inductive and relational biases.

Beyond robustness and expressivity, scalability is essential. Manufacturing processes increasingly involve multi-physics interactions (e.g., thermal-mechanical coupling) and multi-fidelity data streams—from high-resolution simulations to low-quality real-time sensors. Surrogates must integrate such information coherently and scale across spatial and temporal resolutions.

One of the most demanding challenges is generalization and extrapolation. Surrogate models are not just expected to generalize within the domain of their training data, but also to extrapolate to entirely new system configurations and operating conditions without retraining. This is particularly important in flexible or modular manufacturing settings, where system layouts and use-cases evolve continuously. Many current models lack the adaptability to handle such deployment scenarios, especially when failure data is sparse or evolving [14].

Additionally, inverse parameter inference—such as identifying process conditions, material properties, or system-level parameters from observed data—and the explicit modeling of degradation, wear, or other long-term system evolutions remain significant challenges. These limitations restrict the broader applicability of surrogate models in smart manufacturing, where accurate, interpretable, and dynamic models are essential for process optimization, adaptive control, quality assurance, and lifecycle management. **Error! Reference source not found.** illustrates the desired capabilities of surrogate models.

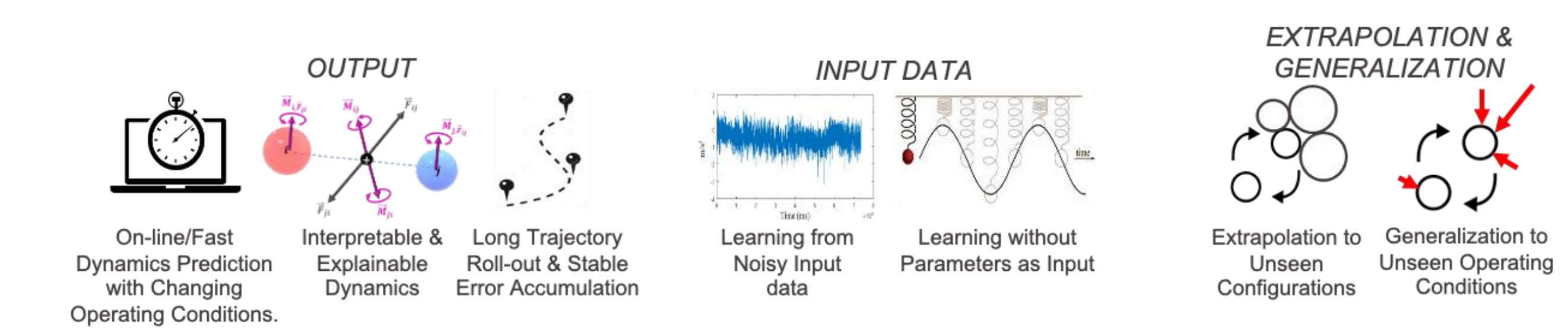

***Figure 1.*** Desired Capabilities of Surrogate Models for PIML-Enabled Smart Manufacturing: Models must deliver interpretable and stable outputs, handle noisy and incomplete inputs, and generalize to unseen configurations and operating conditions.

**Advances in science and technology to meet challenges**

Physics-Informed Graph Neural Networks (PI-GNNs) offer a promising approach for addressing the challenges outlined above. Recent developments integrating geometric and physical inductive biases—ranging from domain-specific physics priors, such as Kirchhoff's laws [15] or heat-flux continuity [16], to broader principles like symmetry [17], thermodynamic laws [18], and momentum balance [4]—extend traditional GNNs into PI-GNNs. These physical inductive biases augment GNN's inherent relational biases, enabling physically grounded and computationally efficient modeling of inter-component interactions, leading to following key capabilities:

- *Data efficiency and long rollout prediction:* Physics-aware inductive bias constrains the hypothesis space, enabling accurate learning from sparse, noisy data and preventing unphysical drift in long -term predictions.

- *Generalization and Extrapolation:* Modular interaction learning and embedded physics enhance transfer to unseen topologies, boundaries, and operating regimes.

- *Interpretability and Explainability:* Physical laws guide message passing, enabling pairwise interactions to represent meaningful internal variables (e.g., stress or heat flux), thus supporting transparency in safety-critical systems.

To fully realize the potential of PI-GNNs as production-grade tools, research must advance in the following three key areas:

1. *Cross-Domain application with universal physics-informed priors:* While some recent PI-GNNs—such as thermodynamics-consistent networks [18] and momentum-preserving equivariant graph nets [4]—show cross-domain potential, most remain constrained by domain-specific biases. To be applicable across coupled thermo-mechanical, electro-mechanical, and fluid-structure systems, future architectures must embed universal physical laws—conservation of energy, momentum, mass, and charge—directly into their message-passing mechanisms.

2. *Virtual sensing*: Estimating unobserved variables—such as residual stresses in metal additive manufacturing or internal shear forces in high-viscosity mixers—is essential for process monitoring. Since these quantities are not directly measurable, PI-GNNs with strong physical priors can infer these internal variables from heterogeneous, multi-fidelity data. Well suited to this task, their learned pairwise messages implicitly represent quantities like contact loads [5]. Guided by physical laws (e.g., momentum conservation), these messages can be decoded into interpretable variables from observable dynamics, supporting adaptive control and defect prevention.

3. *Community benchmarks:* Advancing PI-GNN research in smart manufacturing requires benchmark datasets that combine high-fidelity simulations with real-world sensor data across diverse operating conditions. These datasets should capture domain shifts, sensor noise, and process variability to enable rigorous evaluation of generalization and robustness. Internal state measurements—such as contact loads from specialized test rigs—are especially important for advancing virtual sensing.

When integrated, these key advances will lay the foundation for robust, interpretable, and scalable PI-GNN models that function as reliable digital surrogates—applicable across a wide range of system types— to support smarter and more sustainable system design, real-time monitoring, and adaptive control in next-generation smart manufacturing environments.

### Concluding remarks

Physics-Informed Machine Learning is emerging as a key enabler for building AI systems that are data-efficient, interpretable, and aligned with physical principles—an essential requirement in smart manufacturing environments characterized by complex dynamics, sparse sensing, and safety-critical constraints.

While Graph Neural Networks are not inherently physics-informed, they offer a natural framework for representing the structured, relational nature of manufacturing systems. When extended with additional general physics-based inductive bias—such as conservation laws, symmetry, or energy consistency—GNNs can serve as effective surrogate models and reasoning engines across multi-physics, multi-scale environments.

These physics-informed GNNs show strong potential for supporting tasks such as fault propagation analysis, virtual sensing, long-horizon control, and adaptive monitoring. They also offer differentiable, structured models that can be integrated into modern control architectures.

To fully realize this potential, further research is needed in areas such as generalization, extrapolation, dynamic graph adaptation, and sim-to-real transfer. Developing standardized benchmarks that combine simulation and sensor data will be critical.

As smart manufacturing advances toward autonomous, cyber-physical systems, physics-guided learning frameworks like PI-GNNs will play an increasingly important role in building robust, trustworthy AI solutions.

### Acknowledgements

This work has been supported by the Swiss National Science Foundation (SNSF) Grant 200021_200461.

# Generative AI for Design and Manufacturing

**Faez Ahmed[1], Wei “Wayne” Chen, and Mark Fuge[3]**

[1] Department of Mechanical Engineering, Massachusetts Institute of Technology, Cambridge, USA
[2] J. Mike Walker ’66 Department of Mechanical Engineering, Texas A&M University, College Station, USA
[3] Department of Mechanical and Process Engineering, ETH Zürich, Switzerland

E-mail: faez@mit.edu

### Status

**Why Can’t Machines Design Other Machines (Yet)?** Engineers have long imagined a world where machines could design other machines [1,2], and recent advances in generative AI systems, such as diffusion models and large language models (LLMs), have revived that dream. They can already draft code, summarize documents, and even propose initial engineering concepts. So why aren’t aerospace companies, for example, using them to design and certify entire UAVs from scratch?

This paper argues that the problem is not a lack of imagination, but a set of scientific and organizational barriers. To map that landscape, we divide the problem along two axes (Fig. 1). On one axis lies *design depth*: how well AI can perform specialized tasks such as modeling 3D geometry, surrogate modeling of simulations, design optimization, or uncertainty quantification. On the other axis lies *design breadth*: how well AI can integrate data and knowledge across domains and the entire product lifecycle.

Each axis is constrained in two ways. First, there are *scientific limits*—where the methods themselves are not yet capable. Second, there are *adoption barriers*—issues of trust, data, interoperability, and organizational inertia. Taken together, these four quadrants (Fig. 1) capture key reasons why AI remains more promise than practice in real-world engineering design. This paper first reviews some of those barriers, and then proposes a brief roadmap to overcome them.

## Current and future challenges

### 1.1 Scientific Barriers

#### 1.1.1 Scientific Barriers to Breadth: Why AI Struggles Across the Design Process

Although there are many barriers to learning design breadth, the five main issues are: (1) lack of interoperability among design representations, (2) representation heterogeneity, (3) limited model reasoning, (4) inefficient human-AI Collaboration, and (5) inability to explore high-dimensional design spaces.

The first obstacle to breadth is fragmented data. Modern engineering lifecycles generate enormous amounts of information, including requirements documents, CAD models, simulations, test logs, bills of materials, and manufacturing documents, but these are scattered across incompatible systems. Without consistent, interoperable histories and formats, AI systems cannot learn end-to-end mappings or models that span the entire design lifecycle.

The second problem is heterogeneity. Representations vary in different domains—meshes, splines, point clouds, bills of materials—and subsystems interact in strongly coupled, nonlinear ways. Design often requires multi-discipline, multi-fidelity models in a multi-code environment [3]. Today's domain-specific surrogates break down in these out-of-distribution (OOD) regions. Real-world design may require modeling of behaviors outside of the original training data, such as airplanes facing flutter, control instability, or electronic–thermal interactions.

The third obstacle is a gap in how deeply existing models can reason and plan. Language models can stitch together ideas from diverse fields [4], but their tendency to hallucinate [5] makes them unreliable as integrators of mission-critical systems. Currently, they are often limited to assisting in brainstorming or connecting high-level knowledge [6]. The risk of producing plausible but unsafe or underperforming designs limits their applicability.

The fourth obstacle is the underdeveloped science of human–AI collaboration. We do not yet understand how engineers actually interact with modern AI tools across the lifecycle, how trust is built, how tacit goals are communicated, how authority is shared, and what roles are best played by machines versus humans [7].

Lastly, the fifth obstacle is that many current AI models struggle to generate transformative, detailed designs. Models trained on historical data interpolate well, but can they genuinely invent? As design spaces scale combinatorially, brute-force data-driven or optimization approaches become untenable. We need reliable methods that compose knowledge and search efficiently and effectively. Yet methods and benchmarks today focus almost exclusively on narrow, single-stage tasks, not on cross-stage innovation.

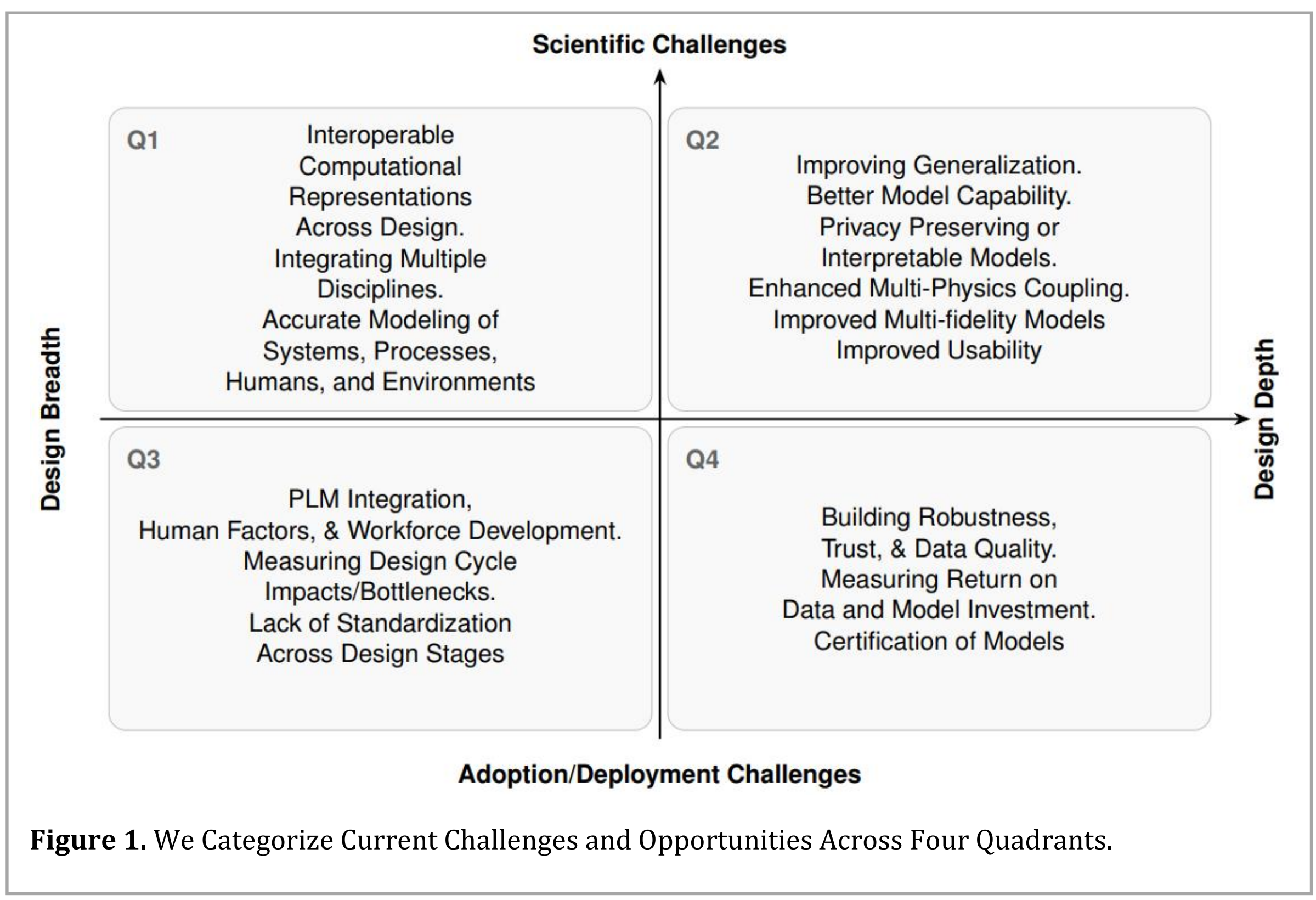

**Figure 1.** We Categorize Current Challenges and Opportunities Across Four Quadrants.

### 1.1.2 Scientific Barriers to Depth: Why Narrow Models Still Fail

Scientific obstacles to achieving depth fall into four main categories: (1) difficulties generalizing, (2) lack of model diagnostics, (3) fast and accurate verification of outputs, and (4) integrating AI with existing toolsets.

First, even within single tasks, AI struggles to generalize. Datasets in engineering are sparse, proprietary, and task-specific. Transfer learning has shown promise in domains like computer vision to reduce data demand [8], but it is less effective in engineering design due to the high heterogeneity of data, representations, and problems. For instance, a model trained on automotive aerodynamics is unlikely to transfer to UAV wings. Engineers need ways to know when a model is operating inside its training manifold and when it is not.

A second challenge is understanding when machine learning models fail and why. Simulation tools, such as CFD solvers, can be checked via methods like mesh convergence. Machine learning models, in contrast, are often difficult to debug, and their error bounds are hard to interpret, especially when the training data is hidden. Engineers require calibrated uncertainty estimates and transparent diagnostics for reliably using machine learning.

A third hurdle is verification. If an AI proposes a design, how do we know it meets our needs? In some cases, physics-based checks are possible, but in most of the applications, just physics-based verification may not be sufficient, and human verification is difficult.

The last hurdle is the need for integration. Too often, AI seeks to replace well-established tools rather than complement them. Hybrid tools could be helpful in such situations: AI accelerating optimizers, suggesting experiments, or guiding simulations—while integrating with the existing design process, tools, and human users.

### 1.1.3 Scientific Desired Future State: What Machines Must Learn

What would it take for machines to design machines? The answer lies in four broad AI capabilities.

First, composition: AI must recognize cross-domain couplings and emergent phenomena [3]. It must understand how local geometry affects global aeroelasticity, or how additive manufacturing paths influence microstructure and thus fatigue life.

Second, abstraction: AI should learn to build and select the right surrogate models, at the right fidelity, and know when those abstractions fail. Like the Wright brothers, it should be able to design experiments to correct its own theories.

Third, decision-making under uncertainty: AI must plan experiments and simulations intelligently, reuse knowledge across tasks, quantify transfer uncertainty, and recognize when "enough" is enough.

Fourth, collaboration with humans and society: AI must elicit intent, surface tradeoffs, justify its decisions, and pass ethical and regulatory scrutiny. It must know when to ask for help.

Milestones on the way include the ability to detect wrong theories, uncover emergent hazards, reframe ill-posed problems, reveal hidden tradeoffs, adapt under constraints, demonstrate genuine novelty, and make credible analogies across domains. Finally, organizations themselves must be ready to absorb the advances.

### 1.2 Adoption Barriers

#### 1.2.1 Adoption Barriers to Breadth: Organizational Gravity

Even when the methods are proficient, organizations face practical roadblocks. Interoperability is one: data integration across tools, vendors, and legacy systems is expensive and difficult to manage and maintain. Intellectual property and privacy are also sensitive topics. Companies need assurance that their proprietary models and datasets will not leak into public systems or to other companies.

Feedback loops are another. Many critical objectives in real-world design scenarios—ease of inspection, maintainability, evolving regulatory priorities—are not explicitly written into AI objectives. As the situation evolves, humans must be able to impose new constraints, and the AI tools must adapt.

Process fit is an important barrier. Engineering workflows rely on version control, iterative reviews, and certification milestones. AI tools that don't align with these are often abandoned. Cultural factors also play a role: engineers are skeptical of "creative" machines, and many enjoy doing the work themselves.

Meanwhile, the workforce also lacks training. Engineers in subjects like civil, mechanical, or aerospace are trained to interpret CFD, FEA, or physical experimentation, not AI methods. We lack both the tools and the training for AI in design and manufacturing.

#### 1.2.2 Adoption Barriers to Depth: Trust and Certification

For depth-focused tools, the key adoption issue is trust [9]. First, engineers want to know what went into the model: how much data, what quality, what diversity. Today, such "datasheets" and contextual information rarely exist. This problem is compounded by a lack of integration with existing tools. Second, engineers cannot rely on models that lack robustness, generalization, explainability, transparency, reproducibility, and accountability [10].

Certification adds another hurdle. Aviation regulators, for example, require auditable chains of evidence. If a design comes from a black-box AI, it is unclear who signs off. Incremental modifications "close to existing designs," such as those obtained by iterative optimization, may be easier to certify, but that disincentivizes radical AI-supported innovation.

Return on investment (ROI) is also difficult to assess due to a lack of standardized assessment. When does AI outperform classical methods, on what metrics, which problems, and by how much? Until clear comparative evidence exists and demos move to production, ROI for companies against trusted workflows might be hard to estimate.

#### 1.2.3 Adoption Barriers to Depth: Trust and Certification

For depth-focused tools, the key adoption issue is trust [9]. First, engineers want to know what went into the model: how much data, what quality, what diversity. Today, such "datasheets" and contextual information rarely exist. This problem is compounded by a lack of integration with existing tools. Second, engineers cannot rely on models that lack robustness, generalization, explainability, transparency, reproducibility, and accountability.

**Advances in science and technology to meet challenges**

We discuss the roadmap for both research and deployment based on four pillars: data and infrastructure, model capability and generalization, workforce and organizations, and trust and compliance. (Four Pillars, Three Horizons, and Clear Metrics)

### 2.1 A Roadmap for Research

*Data and Infrastructure.* We need open, design-relevant datasets and benchmarks [11] that span modalities, disciplines, and lifecycle stages and measure metrics, such as manufacturability and ROI [12]. Shared experimental platforms should allow large-scale, controlled comparisons.

*Model Capability and Generalization.* Multimodal foundation models and agent-based architectures can interpret diverse engineering artifacts and manage interdependencies. But they must move beyond interpolation to robust OOD generalization: physics priors [13], causal rules [14], active learning [15], and meta-learning [16] could be key. Sample efficiency and navigating multi-modality are also needed in engineering.

*Workforce and Organizations.* We need ethnographic and human-computer interaction research on how engineers actually work, plus new interoperability standards to ease toolchain integration. To reduce the gap between academic studies and industry practice, AI integration in real engineering teams should be studied [17]. Education must also prepare "AI-fluent" engineers for hybrid workflows and real-world challenges.

*Trust and Compliance.* Models must produce explainable, auditable design traces that align with certification requirements [18]. Standardized evaluation frameworks should match regulatory expectations.

### 2.2 A Roadmap for Deployment

*Data and Infrastructure.* Near-term priorities are setting data governance standards, unifying platforms, and curating open datasets for benchmarking. Cross-industry investment producing large-scale synthetic and real-world datasets that span across domains and multimodal representations will expand GenAI's scope. Ultimately, allowing AI to access solvers in a self-supervised fashion would enable its own scalable dataset augmentation.

*Model Capability and Generalization.* Early efforts should focus on model accuracy, efficiency, and producing constraint feasibility. Once achieved, the focus can then shift to robust generalization, handling multiple objectives, and reliable OOD performance. Ultimately, AI should move toward handling system-level design tasks while minimizing human guidance.

*Workforce and Organizations.* In the short term, organizations should launch AI literacy programs, update curricula, and run pilots to build skills and adapt workflows. Progress requires formal collaboration frameworks, hybrid teams, and shared best practices. Long term, the aim is an AI-fluent culture where engineers routinely co-design with GenAI, guided by workforce studies and adoption metrics.

*Trust and Compliance.* Short-term efforts must map AI-generated designs to existing certification standards and establish ethics/IP guidelines. Standardized validation, open benchmarks, and accountability frameworks will also be critical. Long term, formal certification pathways for AI-designed products will emerge, with regulators, industry, and academia ensuring reliability, safety, and compliance.

### 2.3 Measuring Progress: From Benchmarks to Real Outcomes

Metrics must track both high-level system outcomes and technical details. On the system side: reductions in design-cycle time, higher first-pass certification rates, broader cross-discipline integration, and workforce adoption and satisfaction. On the technical side: effectiveness, sample efficiency, calibrated uncertainty, OOD robustness, manufacturability, and ROI are key.

Importantly, metrics must be disentangled across the four quadrants—breadth vs. depth, science vs. adoption—so stakeholders can see exactly where progress is happening and where it is not. Significant research and investments are still needed to establish a large number of diverse and realistic benchmarks for progress within Design and Manufacturing (e.g., [19]). By comparison, LLM progress has been accelerated due to a variety of diverse benchmarks [20] could the same benefits be brought to design?

## Concluding remarks

**Why Machines Don't Design Machines?** The reason machines don't yet design machines is not a single missing breakthrough. It is a combination of scientific and institutional deficits. Scientifically, our models lack reliable generalization, uncertainty quantification, and system-level reasoning across heterogeneous, tightly coupled domains. Institutionally, we lack interoperable data infrastructure, certification frameworks, cultural acceptance, and workforce integration.

The path forward is clear: build domain-relevant benchmarks and datasets; harden models with physics, uncertainty, and compositional reasoning; redesign workflows for human–AI collaboration; and codify trust and certification pipelines. Only by tackling science and adoption together, across both breadth and depth, will we turn AI design from clever demos into certified engineering practice.

## Acknowledgements

PI Ahmed acknowledges support from the National Science Foundation under CAREER Award No. 2443429.

# Semantic Framework Enabling Machine Learning in Manufacturing

**Arild Waaler[1], Martin G. Skjæveland[1] and Dimitris Kyritsis[1]**

[1] Department of Informatics, University of Oslo, Norway

E-mail: arild@uio.no

## Status

Engineering practice today is fundamentally dependent on technical information, information that is document-based, fragmented across tools and domains, and tied to siloed organizational structures. While individual disciplines to varying degrees employ formal models like ontologies, there is no shared foundation for structuring system-level knowledge across the engineering lifecycle. As a result, the so-called *digital thread* — the traceable connection from requirements through design, implementation, and operation — is frequently broken or opaque.

At the same time, artificial intelligence, particularly in the form of machine learning and generative models, is rapidly being integrated into engineering workflows. However, these approaches struggle to deliver trustworthy results in the absence of clearly defined objects and relations. Where structure is implicit or missing, AI models become unreliable, and engineers are left without the means to validate or interpret the outputs of AI models. This creates a gap between the promise of AI and the reality of high-stakes engineering practice, where precision, safety, trust, and traceability are imperative.

To address this, there is an urgent need for structured representations of engineering knowledge that are both verifiable and understandable. This requires enabling expert users — across engineering disciplines — to validate, reuse, and refine the information and data models that underpin AI and automation. Crucially, such validation should be grounded in established systems engineering principles: abstraction for information hiding, hierarchical decomposition for modularity, topology for managing flows, interfaces for encapsulation, and classification for reuse.

Formal verification demands even more: a logical foundation that supports tractable reasoning. This includes the ability to define and check class axioms, detect inconsistencies, and infer consequences within decidable subsets of logic. Without such foundations, technical information models cannot be reliably queried, reused, or integrated at scale.

The lack of structure in technical information also undermines efforts to build scalable industrial knowledge graphs. While knowledge graph and ontology standards such as RDF [8] and OWL [9] are widely used, they often lack connection to engineering practice and do not capture the structural logic of systems. A principled foundation is needed, one that connects engineering semantics with semantics-based representations and can serve both as a modeling framework and as machine-readable input to AI pipelines [1].

### Current and future challenges

To enable digital transformation in engineering, we need languages that allow engineers to describe systems in a way that is both human-readable and machine-actionable. Such languages must support abstraction, modularity, encapsulation of interfaces, and reuse — all core principles of systems engineering. They must also allow engineers to express partial, evolving structures, reflecting the reality that system models are rarely complete at any one time.

At the same time, to support automation, AI, and formal reasoning, these languages must have a well-defined logical foundation. Ontology languages like OWL offer precise semantics based on Description Logic [10], supporting classification, consistency checking, and inference. However, they are often ill-suited to capture the structural and contextual richness of engineering systems: they lack native support for system-level constructs such as breakdown hierarchies, connectivity relations, modalities like intended versus actual configurations, and lifecycle-specific views.

This reveals a fundamental conceptual gap. Engineering requires modelling languages that express *intensional structure* — definitions of system elements in terms of their roles, constraints, and relationships — while ontology-based approaches typically focus on *extensional* classification and static taxonomies. Current tools and languages seldom support the coherent expression of intensions in a way that can be incrementally developed, reused, and verified.

This gap also manifests in knowledge graph construction: current ontology languages provide semantic rigor, but not the structural expressiveness engineers need to model real systems. Scalable industrial knowledge graphs thus remain difficult to construct and maintain. Moreover, many advanced ML algorithms rely on structured, semantically rich graph inputs — a need yet to be fully met in engineering domains [1].

Bridging these gaps requires rethinking how we represent engineering knowledge: starting from the needs of engineering practice, while grounding models in logic-based semantics and enabling automation. Several key challenges emerge:

- How can we define structural specifications that are both readable and logically precise?

- How can we preserve traceability across partial models and evolving designs?
- How can we align domain-specific engineering practices with shared ontological foundations?
- How can we combine human-driven design with machine-generated model structures?

To meet these challenges, a new framework must treat intensions as first-class citizens, enabling models that are contextual, compositional, semantically transparent, and AI-ready.

### Advances in science and technology to meet challenges

Addressing the gap between engineering modelling needs and formal ontology capabilities requires a new kind of framework, one that combines the expressive power of systems engineering with the precision of formal logic. The *Information Modelling Framework (IMF)* [7, 2] is designed to meet this need. It provides a simple yet comprehensive and extensible core language that allows engineers to express intensional definitions of system elements in a way that is formally interpretable, incrementally buildable, and amenable to verification.

The foundation of IMF is a clear separation between *intension* and *extension*. An IMF element expresses a structural specification, a parameterized definition of what a system element is intended to be, rather than what currently exists. Each such specification is annotated with an *aspect*: a structured context capturing the *information domain* (e.g., *function*, *implementation*), the *modality* (e.g., *intended*, *actual*), and the *lifecycle perspective* (e.g., *product*, *project*), see Figure 1. This framing enables coherent modeling of alternative designs, requirements vs. implementations, and different stakeholder views — all within a unified formalism.

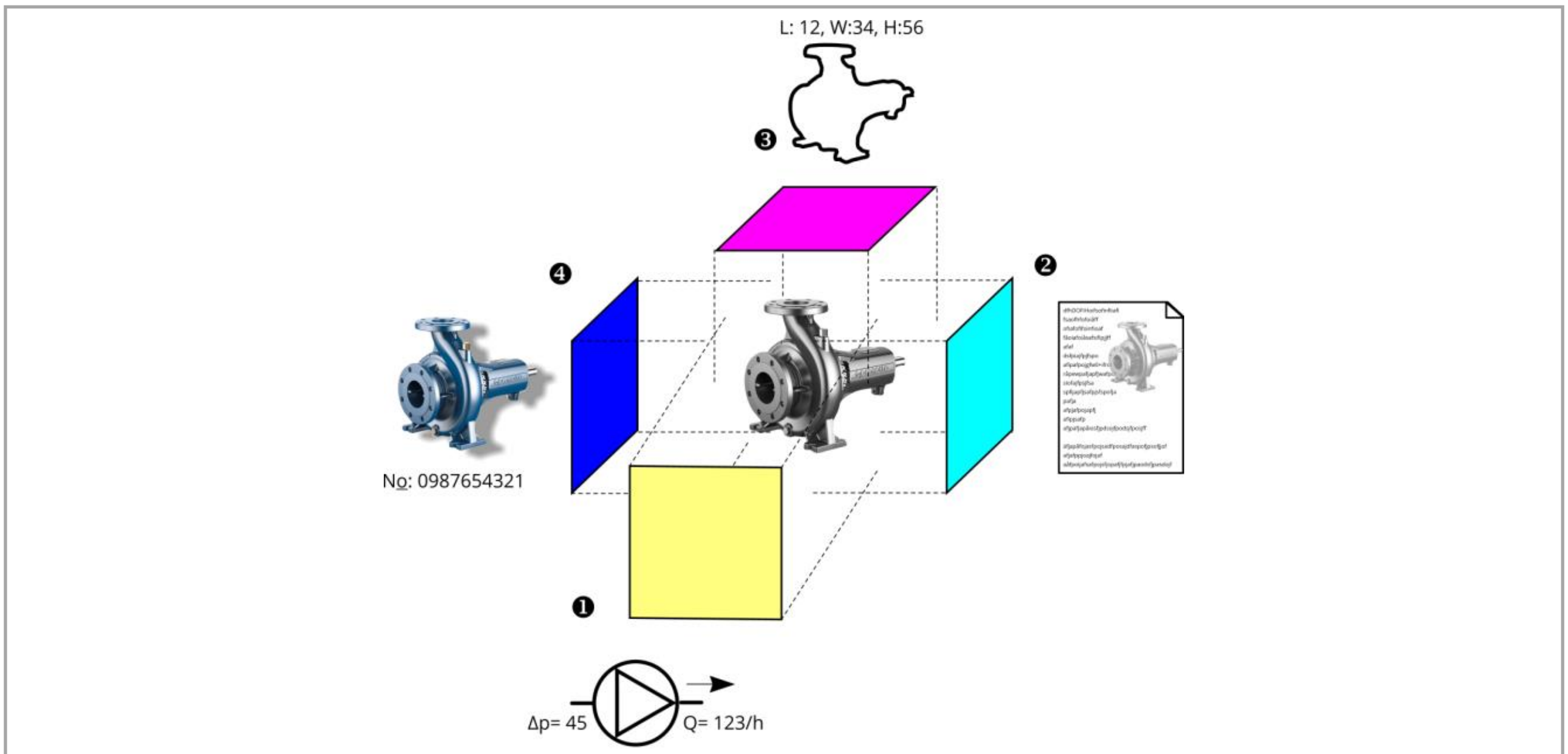

**Figure 1.** Information in the Information Modelling Framework (IMF) is represented in distinct and interrelated aspects. The figure illustrates four complementary specifications of different aspects that together provide a comprehensive description of a pump: *intended functional requirements* (1, yellow), *intended implementation (product) specification* (2, cyan), *intended spatial requirements* (3, magenta), and *actual installed descriptions* (4, blue).

Formally, IMF uses constructs from *typed lambda calculus* to define structural specifications. Relations such as hasPart, connectedTo, and hasTerminal are interpreted as function applications, with semantics grounded in intensional logic. The resulting models can be translated into extensional axiom sets (e.g., in Description Logic), enabling formal reasoning tools to verify properties, detect inconsistencies, and support model completion.

Because IMF models are grounded in formal semantics and can be serialized as knowledge graphs using RDF, they serve as a natural foundation for *industrial knowledge graphs*. These graphs are semantically rich, structured according to systems engineering logic, and readable by machines. Moreover, the RDF representation of IMF models can be directly consumed by advanced ML pipelines, providing AI models with well-formed, semantically validated engineering structures.

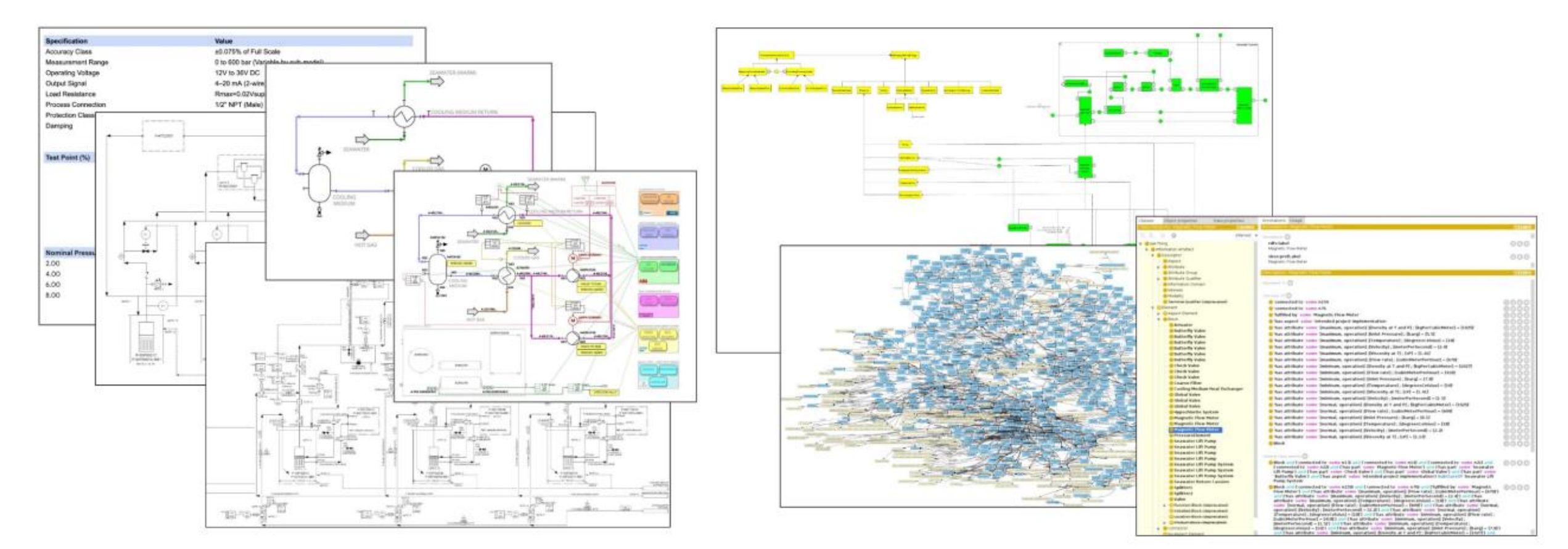

**Figure 2.** In the EU-funded project Tec4MaaSEs [6], the Information Modelling Framework (IMF) is used to capture and represent both the requirements for complex engineering artefacts and the specifications of the equipment intended to fulfil them. Technical information from various perspectives and in different formats (depicted on the left) is consolidated within a structured IMF information model (top right). This model serves as the foundation for generating an RDF knowledge graph and an OWL ontology (bottom right), enabling automated verification, reasoning, and analysis.

IMF is defined in an openly accessible specification [7] and supported by a Recommended Practice for Asset Information Modelling [2]. Furthermore, the IMF program publishes its semantic technology resources, including its OWL ontology and SHACL shape patterns [3]. IMF is not a fixed standard, but a platform for structured, formal, and open-ended model development. It is already being explored in the EU-funded projects RE4DY [4], SM4RTENANCE [5], and Tec4MaaSEs [6] — see Figure 2 for an example.

## Concluding remarks

The Information Modelling Framework (IMF) offers a principled foundation for structuring engineering knowledge in ways that support human understanding, AI-driven automation, and formal verification. By bridging the gap between system modelling practices and logic-based ontologies, IMF enables a new generation of engineering tools that are modular, semantically precise, and adaptable across contexts.

Crucially, IMF provides a foundation for scalable industrial knowledge graphs. Its RDF serialisation format allows both semantic integration and direct input to machine learning workflows — a key enabler for engineering AI.

More than a static language, IMF is a research and innovation program. It invites collaboration from engineers, logicians, data scientists, and tool developers to extend its capabilities and apply it to real-world challenges. By combining systems principles with formal semantics, IMF supports the development of trustworthy, explainable, and scalable digital engineering infrastructure.

### Acknowledgements

This research has been partially supported by the European Commission in the Horizon Europe projects RE4DY (grant 101058384), SM4RTENANCE (grant 101123490) and Tec4MaaSEs (grant 101138517).

# Physics-Based Predictive Control and Real-Time Decisions for Digital Twin–Enabled Autonomous Manufacturing

**Wei Chen[1], Vispi Nevile Karkaria[1], Yi-Ping Chen[1] and Ying-Kuan Tsai,[1]**

[1] Department of Mechanical Engineering, Northwestern University, Evanston, IL, USA

E-mail: weichen@northwestern.edu

### Status

Digital twins are beginning to transform manufacturing beyond early "digital shadows," which only mirror system states for monitoring, by enabling bidirectional feedback for real-time predictive control and decision-making [1,2]. According to the National Academies report [3], a digital twin is "a set of virtual information constructs that mimics the structure, context, and behavior of a physical system or system-of-systems, is dynamically updated with data from its physical counterpart, has predictive capability, and informs decisions that realize value." Figure 1 illustrates our digital twin framework, where physics-based modeling and simulation build a machine learning surrogate for model predictive control (MPC), enabling bidirectional feedback with the physical system; this approach has been demonstrated in directed energy deposition (DED) using a Time Series Dense Encoder (TiDE) to predict melt pool features and optimize laser power for improved quality and reduced defects [4].

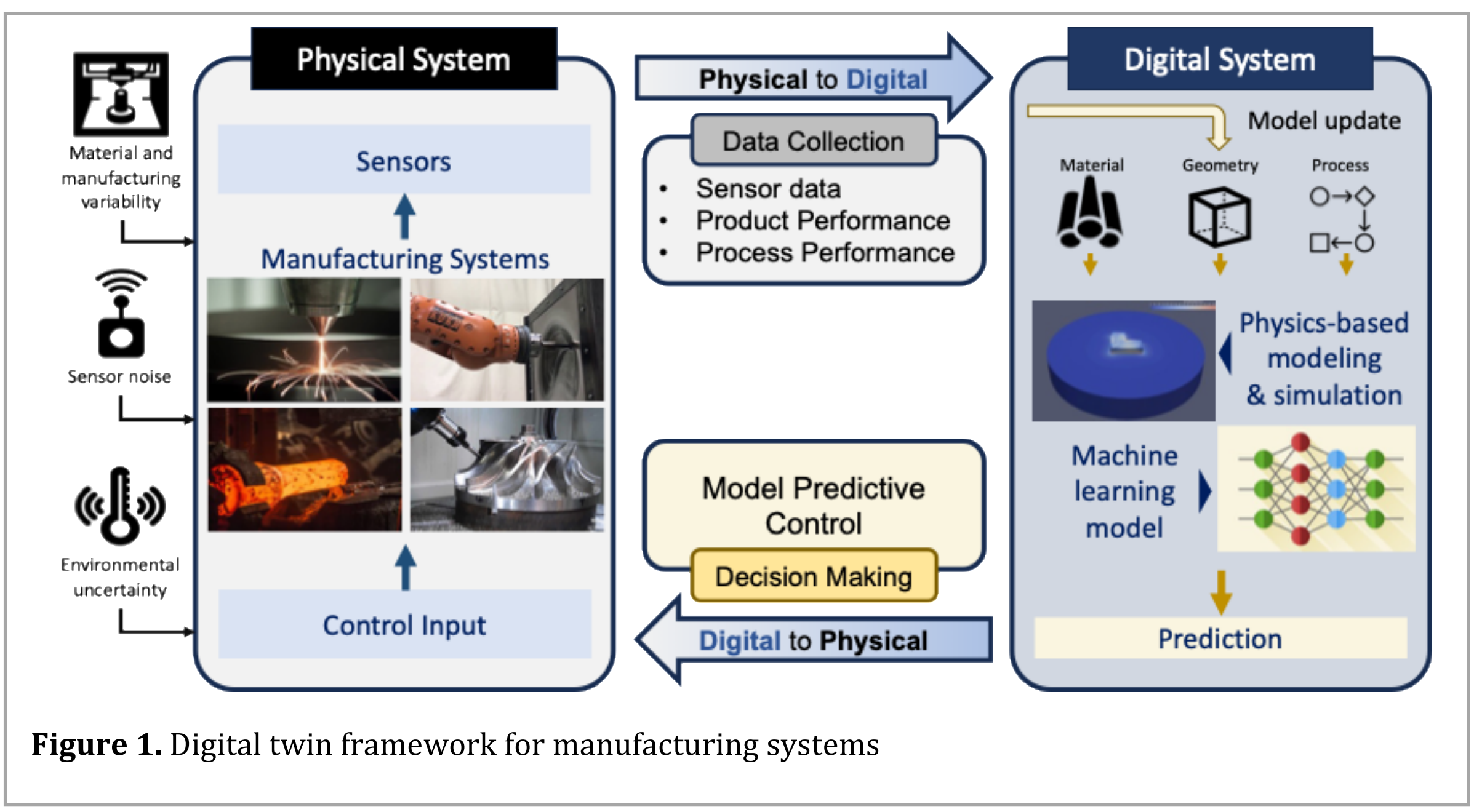

**Figure 1.** Digital twin framework for manufacturing systems

Despite these recent advances, most deployed digital twins in manufacturing still fall short of being fully autonomous, and need human experts in the loop to validate model updates and resolve edge-case decisions, which are rare or unusual scenarios outside normal operating conditions [5]. Looking ahead, the status quo in manufacturing digital twins must evolve from digital shadows to self-optimizing agents that not only forecast failures and optimize process but also learn evolving process dynamics, adapt to new products, and collaborate across multiple manufacturing processes, laying the foundation for the next generation of digital twins that drive lean, resilient, and sustainable manufacturing ecosystems [6].

### Current and future challenges

Modern manufacturing processes such as additive manufacturing and precision welding involve complex spatio-temporal physics that are difficult to model in real time. While high-fidelity methods like finite element or computational fluid dynamics can capture these dynamics, they are too computationally expensive for closed-loop control, making fast machine learning surrogates essential [7]. By extracting patterns from sensor and simulation data, these surrogates can deliver millisecond-scale predictions on edge devices, enabling ultra-low latency inference. However, most digital twins still model only temporal or spatial dynamics in isolation, and many treat processes as black boxes rather than learning the underlying physical laws, creating challenges in both physics discovery from sparse data and generalization to new conditions [8]. These difficulties are amplified in high-cost processes where data are scarce, especially for rare defects, limiting calibration and adaptation. While offline retraining can restore accuracy, continuous online updating without disrupting production remains challenging, with methods such as Koopman operators and reduced-order models offering potential but requiring further research [9].

Transitioning from surrogate modeling to real-time decision making in the digital twin paradigm requires uncertainty-aware machine learning and decisions throughout the system's lifecycle, which often involves a tradeoff between model fidelity and computational efficiency. For example, when implementing MPC with discrete event simulation [10], the choice of model and optimization method, such as quadratic programming with simplified linear approximation or numerical optimization with a detailed physics-based model, strongly affects accuracy, decision frequency, and the overall quality of control. Moreover, as system behavior may change under different conditions or at various stages of life cycles, effective and data-efficient methods for continuous model adaptation remain an unsolved challenge.

The challenges pointed out so far highlight the importance of continuous verification, validation, and uncertainty quantification (VVUQ) which is still a hurdle in the deployment of digital twins for manufacturing [3,6]. While standards such as ISO 23247 (digital twin framework for manufacturing) offer structural guidance, they do not address credibility assessment or VVUQ. Digital twins are inherently complex and face epistemic, aleatoric, and even unknown uncertainties, requiring robust VVUQ to ensure reliability. The dynamic nature of digital twins requires ongoing validation to remain accurate as conditions evolve, raising questions about update frequency and computational feasibility.

Microstructure plays a critical role in manufacturing, as its evolution depends on geometry and process conditions and ultimately governs the mechanical, thermal, and chemical properties that determine component performance. Understanding the process–microstructure–property–performance relationships is essential for optimization [11] , yet predicting part-scale evolution remains challenging due to the high cost of simulations, stochastic physical phenomena, and limited in-situ monitoring in high-speed or high-temperature processes. In additive manufacturing, for example, rapid thermal cycling produces nonlinear, path-dependent grain structures that are difficult to model [12]. These complexities also impact the co-design of geometry, materials, and processes, where microstructural heterogeneity complicates optimization. Addressing this challenge requires frameworks that incorporate microstructure-related objectives into early design decisions and refine them through adaptive process optimization.

**Advances in science and technology to meet challenges**

To model dynamic manufacturing processes that evolve across time and space, machine learning methods such as spatio-temporal neural operators, graph neural networks, and transformer encoders have been used to extract patterns from sensor data, simulations, and historical logs. Heterogeneous data fusion has been addressed with approaches like latent variable Gaussian processes (LVGP) and Temporal Fusion Transformers [13], while adaptive sampling, multi-fidelity modeling, transfer learning, and synthetic data pre-training help mitigate data scarcity [14]. Uncertainty is handled through models such as Gaussian processes, Bayesian neural networks, or deep ensembles, which provide calibrated confidence intervals for risk-aware decision making. To meet performance demands, lightweight architectures and optimized edge deployment enable fast, cost-efficient inference. Building on these advances, pre-trained neural networks now act as surrogates for rapid evaluation, support efficient policy learning to map states to actions, and generate probabilistic forecasts that guide uncertainty-aware, real-time decision-making [15].

To ensure that surrogate models remain reliable throughout the system lifecycle, recent research has focused on continuous model validation [16] and adaptation. This involves not only detecting when a model no longer aligns with the physical systems but also updating the model efficiently with least compromised performance. Emerging approaches for concept drift detection have improved the robustness of real-time monitoring, especially for neural networks and multivariate predictive frameworks [17]. Concurrently, lightweight adaptation methods such as low-rank fine-tuning enable rapid updates with minimal data, preserving fidelity while adapting to evolving manufacturing conditions. These developments are critical for enabling trustworthy, self-improving digital twin systems that support resilient and adaptive decision-making across dynamic operational environments.

Recent advances in AI and ML are transforming how microstructure is predicted and optimized in manufacturing to address the impracticality of extensive experimentation and the high computational cost of physics-based modeling [18]. Multi-scale modeling approaches now couple macro-level processing conditions with micro- and meso-scale material behaviors, allowing for more accurate prediction of microstructural evolution and its effect on material performance. These models are increasingly being enhanced by data-driven techniques, such as Generative Adversarial Networks (GANs) that generate realistic microstructures for virtual design [19], and convolutional neural networks (CNNs) that predict stress–strain curves and material properties from microstructural images; and recurrent neural networks (RNNs) capture history-dependent plasticity by learning deformation-path-sensitive responses [12]. These capabilities are embedded in integrated optimization frameworks that use Bayesian optimization, evolutionary algorithms, and transfer learning to discover process conditions yielding optimal microstructures and properties.

By extending these methods to co-design of geometry, materials, and processes, deep learning and inverse optimization can account for heterogeneous microstructural evolution [20]. Although co-design is high-dimensional, differentiable physics-based modeling such as JAX-FEM, enables gradient-based optimization that directly links design variables to microstructural outcomes, accelerating the search for optimal solutions while preserving physical fidelity.

### Concluding remarks

In summary, advancing manufacturing digital twins requires addressing key needs such as spatio-temporal modeling, surrogate-based real-time control, uncertainty quantification, and microstructure-aware co-design. State-of-the-art physics-informed machine learning approaches, including spatio-temporal neural operators, graph neural networks, and transformer encoders, provide unified process modeling capabilities. Reliable deployment also depends on continuous model updating through fast surrogates, multimodal data fusion, adaptive sampling, transfer learning, concept drift detection, and multi-scale predictive frameworks. Looking ahead, the next generation of autonomous digital twins will incorporate cyber-secure federated learning, scalable data infrastructure, trustworthy AI, and edge-to-cloud integration, enabling self-learning and adaptive decision-making. These advances will support resilient, sustainable, and agile manufacturing with concurrent optimization of processes, materials, and designs.

### Acknowledgements

We are grateful for the grant support from the National Science Foundation's Engineering Research Center for Hybrid Autonomous Manufacturing: Moving from Evolution to Revolution (ERC-HAMMER), under the Award Number EEC-2133630. Grant support from the Remade institute research program (DE-EE0007897) is greatly acknowledged. Yi-Ping Chen also acknowledges the Taiwan-Northwestern Doctoral Scholarship.

# Trustworthy AI for Manufacturing

**Joseph Cohen[1] and Xun Huan[2]**

[1] Department of Mechanical and Aerospace Engineering, Rutgers University, Piscataway, NJ, USA
[2] Department of Mechanical Engineering, University of Michigan, Ann Arbor, MI, USA

E-mail: joseph.cohen3@rutgers.edu, xhuan@umich.edu

## Status

History has shown that trust in new technologies is rarely immediate. Industrial automation, aviation autopilots, and self-driving cars all faced initial skepticism, earning acceptance only after demonstrating reliability, safety, and integration with human oversight and regulations. Artificial intelligence (AI) in manufacturing stands at a similar crossroads: to be trusted, AI systems must not only perform well, but also be explainable, accountable, and situated within human and institutional contexts.

AI is rapidly transforming manufacturing, advancing prognostics and health management [1], predictive maintenance [2], quality control [3], process optimization [4], and industrial digital twins [5]. These technologies improve throughput and yield, reduce downtime, and enhance quality in complex, multistage manufacturing environments. However, the opaque, black-box nature of AI raises critical concerns about transparency, reliability, and safety, especially in high-stakes domains such as aerospace and medical device manufacturing. These concerns can lead to resistance or overreliance, ultimately undermining the transformative potential of AI.

Trust in AI extends beyond technical performance, encompassing also cultural and organizational elements. While organizations with strong innovation cultures may be more willing to adopt AI, others require stronger assurances of reliability and compliance. Even highly accurate models paired with explainable AI can fail to earn trust if their explanations do not meaningfully represent system behavior. After all, an AI “explanation” that humans cannot understand or act upon is not truly an explanation.

The institutional dimension of trust is equally critical. Since AI systems cannot be held legally accountable, responsibility ultimately rests with people and organizations. This makes trustworthiness essential not only for adoption, but for assigning responsibility, mitigating liability risks, and enabling regulatory oversight. In high-consequence manufacturing domains, trust also depends on insurability and certifiability: systems must not only be reliable in terms of performance, but also legally and operationally dependable.

## Current and future challenges

The path toward trustworthy AI for manufacturing faces interconnected challenges that we organize into two categories: **computational and modeling** and **human and societal** (see Figure 1).

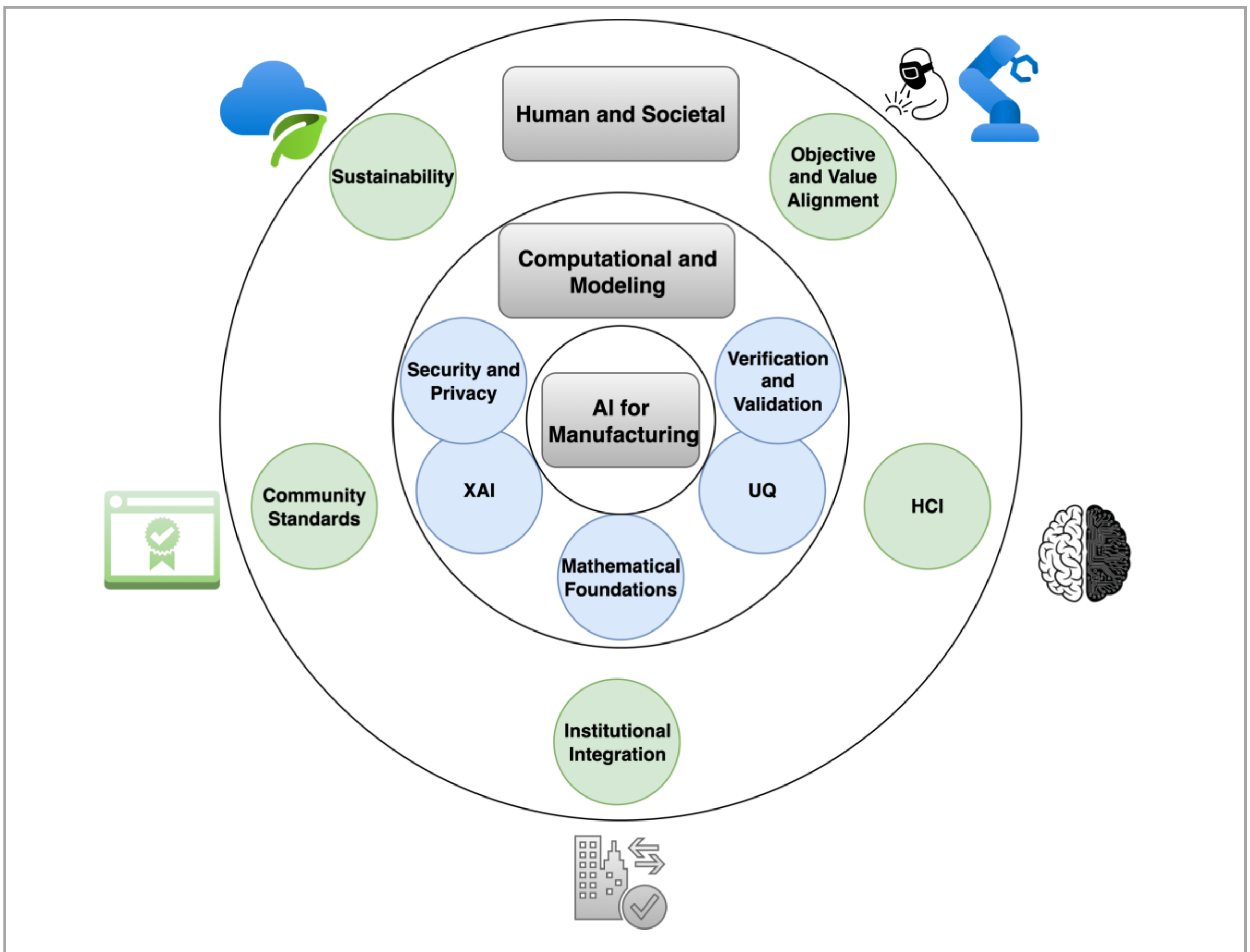

**Figure 1.** Trustworthy AI framework for manufacturing. Technical foundations in computational modeling (inner ring) enable human and societal factors (outer ring) that are essential for successful AI adoption in manufacturing environments.

1. Computational and modeling challenges

**Interpretability and explainability:** Modern AI systems often function as black boxes with deep architectures creating abstract latent representations that are not readily interpretable, making it difficult for users to understand prediction processes. Most systems also lack mechanisms to articulate the reasoning behind their outputs, constraining their explainability.

**Mathematical foundations:** Unlike established computational frameworks such as the finite element method and numerical integration, many modern AI models—especially those with complex architectures and foundation models—have relatively limited theoretical analysis [6,7] (e.g., on consistency, stability, and convergence). This theoretical gap means systems are often deployed based on empirical performance alone, making it difficult to generalize into unseen and sparse regimes, understand unexpected behavior, or make systematic improvements.

**Verification, validation, and uncertainty quantification (VVUQ):** Despite operating with noisy, sparse data and potentially misspecified assumptions, AI systems typically provide predictions without meaningful confidence measures. This absence of principled uncertainty quantification (UQ) can lead to overconfident predictions—resulting in unexpected risks and failures—or overly cautious predictions, which may cause unnecessary safety margins and wasted resources. Robust deployment also requires formal verification (establishing mathematical correctness and numerical accuracy) and validation (ensuring accurate representation of real-world systems), under a comprehensive VVUQ framework [8].

**Security and privacy:** As AI systems become increasingly dependent on digital infrastructure, they also become increasingly vulnerable to adversarial attacks from malicious actors. Centralized data repositories, cloud services, and connected devices face threats including data poisoning, model inversion, and data leaks [9] that can degrade performance, compromise sensitive information, and erode trust.

**Generative AI:** The emergence of large language models (LLMs) and agentic systems introduce novel concerns including hallucinated outputs, unpredictable behavior, and untraceable reasoning pathways. As such systems are deployed in sensitive manufacturing stages—process planning, design, maintenance, and operations—it becomes increasingly important to understand not only what they can do, but when, where, and why they fail.

2. Human and societal challenges

**Human-computer interaction (HCI):** Trust requires effective communication of predictions, uncertainties, and explanations tailored to users' roles, backgrounds, and decision contexts [10]. For example, a floor operator needs intuitive visualizations, while a manager requires aggregated summaries tied to strategic performance indicators. Trust also evolves over time, requiring systems to support iterative interaction, feedback, and adaptation as both users and the manufacturing environment change.

**Organizational integration:** AI must facilitate consistent communication across diverse teams, including operations, engineering, and business strategy. Explanations must be both role-specific and translatable across functions, often needing specialized multi-agent systems rather than isolated models. AI adoption must also navigate organizational culture while upholding fairness, equity, and non-discrimination principles, particularly amid labor-management tensions and perceived job displacement. Additionally, successful integration of AI with legacy systems and domain expertise is essential to preserve valuable human knowledge and intuition.

**Objective and value misalignment:** AI models typically optimize narrow mathematical loss functions that may fail to reflect real-world priorities, ethical standards, or user intentions. In manufacturing, this can create new failure modes or neglect long-term tradeoffs important to stakeholders. The saying "all models are wrong" takes new significance in AI, where this wrongness can manifest silently and catastrophically when misaligned with emergent domain realities.

**Sustainability:** The growing energy demands of training and deploying large models strain power infrastructure. High computational costs also impede industrial adoption, particularly for smaller manufacturers. Without innovation in energy-efficient algorithms and infrastructure, the environmental footprint of AI may become a barrier inhibiting responsible long-term adoption.

### Advances in science and technology to meet challenges

We organize the recent advances in trustworthy AI for manufacturing to mirror the earlier challenge categories: **computational and modeling** and **human and societal**.

1. Computational and modeling advances

**Explainable AI (XAI):** Post-hoc explanation methods such as Shapley values [11], counterfactual explanations [12], and saliency-based attribution [13] continue progress to provide interpretable model outputs. Furthermore, developments in causal modeling [14] and counterfactual reasoning offer robust insights into cause-and-effect relationships and support "what-if" scenario analysis critical for manufacturing decision-making.

**Mathematical foundations:** Researchers increasingly investigate AI models' analytical properties including complex training dynamics such as phase transitions, double descent, and grokking [6,15]. Efforts to bridge AI with classical numerical analysis and scientific computing are essential for integrating AI into the manufacturing computational infrastructure. Physics-informed AI and hybrid modeling approaches, such as neural operators [16] and physics-informed neural networks (PINNs) [17], combine data-driven learning with domain-specific physical laws, offering enhanced trust through constraint-based learning and improved extrapolation beyond training regimes.

**Advanced UQ:** UQ provides frameworks to characterize and communicate both epistemic and aleatoric uncertainty [18]. Bayesian approaches provide principled frameworks for representing epistemic uncertainty (what models do not know due to limited knowledge), particularly valuable for updating uncertainty estimates by assimilating sparse, noisy, and indirectly observed data. Ensemble-based and non-Bayesian methods effectively capture aleatoric uncertainty (inherent variability or randomness), crucial in manufacturing where processes are often stochastic. These approaches together support decision-making by providing not only AI model predictions but also their confidence levels.

**Monitoring for AI lifecycle:** Techniques to detect distributional shift, concept drift, and anomalies are enabling real-time awareness of model degradation [19]. These capabilities are essential to managing the full lifecycle of AI models, where they can be continuously validated, updated, or decommissioned in response to evolving real-world conditions.

2. Human and societal advances

**Human-centered design:** Research increasingly emphasizes user-adaptive, interactive explanations over static output. Modern explanation interfaces support dialogue-based refinement where users can ask follow-up questions and steer explanations toward what they find meaningful [10]. This approach supports trust calibration, helping users develop appropriate mental models of when to rely on AI and when to challenge them.

**Human-robot collaboration:** Advances in human-robot teaming [20] focus on settings where AI assists rather than replaces human decision-makers, such as factory floors with autonomous systems and human operators. These directions will enable smoother, safer coordination between human expertise and AI capabilities.

**Standards and community infrastructure:** Growing momentum toward shared standards, regulations, and community infrastructure include benchmark datasets with labeled outcomes, metadata, and uncertainty; standardized metrics for evaluating performance as well as explainability and robustness; third-party certification and validation frameworks; and open-source platforms and documentation standards to promote transparency and reproducibility.

**Sustainable AI development:** Energy-efficient AI advances including model compression, hardware-aware training, and carbon-conscious deployment aimed at reducing the resource demands of both development and operation. These methods become increasingly important as the scale and number of deployed models continue to grow, ensuring that AI adoption remains environmentally responsible.

### Concluding remarks

Trustworthy AI is a fundamental requirement for next-generation manufacturing. As AI systems become more capable and ubiquitous, the risks of opacity, misalignment, and failure scale alongside their potential benefits. Meeting these challenges requires coordinated progress across computational and modeling as well as human and societal foundations.

The maturation of AI mirrors traditional engineering disciplines' evolution through rigorous theory, verifiable practices, and earned community trust. Achieving this vision requires bringing together mathematicians, computer scientists, engineers, social scientists, domain experts, and end users. Success will bring AI systems to manufacturing that are not only powerful, but also understandable, reliable, and worthy of our trust.

# Enabling Dependability in Smart Manufacturing: RAMS and AI/ML Integration

**Jing (Janet) Lin[1], Liangwei Zhang[2]**

[1] Department of Civil, Environmental and Natural Resources Engineering, Luleå University of Technology, Luleå, Sweden
[2] Department of Industrial Engineering, Dongguan University of Technology, Dongguan, China

E-mail: janet.lin@ltu.se

### Status

Reliability, Availability, Maintainability, and Safety (RAMS) have long provided the foundation for asset design, maintenance, and operational optimization in manufacturing. Traditional RAMS approaches—often supported by Reliability-Centered Maintenance (RCM), Condition-Based Maintenance (CBM), and Prognostics and Health Management (PHM)—have primarily addressed physical degradation and failure patterns [1]. These methods focus on maximizing uptime and minimizing risk through structured maintenance planning and statistical analysis.

However, the nature of manufacturing systems is changing. The emergence of Cyber-Physical Systems (CPS), the Industrial Internet of Things (IIoT), and embedded Artificial Intelligence (AI) is transforming industrial environments into intelligent, interconnected ecosystems [2]. In this context, assets are no longer standalone mechanical components; they are dynamic, software-integrated entities that operate in real time, interact with users and other machines, and continuously adapt to contextual and environmental changes.

To meet the demands of this new landscape, we introduce Dependability-Centered Asset Management (DCAM)—a forward-looking framework that extends and evolves the RAMS paradigm [3]. DCAM integrates traditional reliability engineering with system-level dependability science and AI assurance methods to address modern challenges such as digital traceability, cyber-physical resilience, and lifecycle adaptability.

DCAM promotes a holistic, lifecycle-oriented approach to dependability. It embeds reliability thinking from the earliest stages of design through to operation, evolution, and renewal. It draws upon digital technologies—including machine learning, digital twins, and edge/cloud architectures—to support predictive diagnostics, adaptive maintenance, and context-aware decision-making.

By shifting focus from static reliability metrics such as Mean Time Between Failures (MTBF) to dynamic indicators of system resilience and AI model trustworthiness, DCAM aligns technical performance with broader goals of transparency, sustainability, and operational integrity. It enables manufacturers to respond not only to mechanical failure, but also to the risks and uncertainties introduced by AI-driven automation and distributed intelligence.

As manufacturing systems become increasingly complex and autonomous, RAMS must evolve accordingly. DCAM offers a bridge between legacy reliability principles and future-ready dependability strategies—uniting physical, digital, and organizational dimensions into a unified framework for smart manufacturing.

While RAMS principles provide the foundation, Table 1 highlights how asset management paradigms have evolved—from reliability-centered and software-driven approaches toward the integrated, AI-enabled perspective embodied in DCAM, which addresses the complexity of cyber-physical systems and lifecycle sustainability in smart manufacturing.

**Table 1.** Evolution of Asset Management Paradigms Toward Smart Manufacturing Dependability

| Aspect | Traditional Asset Management (Reliability-Centered) | Traditional Dependability Management | Dependability-Centered Asset Management (DCAM) |
|---|---|---|---|
| Primary Focus | Preventing physical failures and planning maintenance | Ensuring system behavior under faults and threats | Lifecycle-wide trust, resilience, and value creation |
| Origin Discipline | Reliability engineering, maintenance | Computer science, systems engineering | Interdisciplinary: engineering, computing, sustainability |
| Typical Domains | Manufacturing, transportation, utilities | Embedded systems, software, cyber-physical systems | Smart factories, autonomous assets, socio-technical systems |
| Core Concepts | RAMS, RCM, PHM | Dependability, fault tolerance, robustness | Context-aware modeling, AI diagnostics, digital twins, sustainability |
| Key Metrics | MTBF, failure rate, lifecycle cost (LCC), RUL | Availability, safety, integrity, security | Composite dependability index, adaptability, RUL, environmental KPIs |
| Tools & Methods | FMEA, LCC analysis, CBM, RCM platforms | Fault trees, formal methods, verification tools | AI/ML models, digital twins, twin-based optimization, sustainability analytics |
| Limitations | Hardware-focused; limited handling of software and context | Strong on software; weak in physical lifecycle integration | Designed for CPS; integrates physical, digital, and sustainability dimensions |

**Current and future challenges**

The integration of Artificial Intelligence (AI) and Machine Learning (ML) into manufacturing systems has catalyzed a shift from reactive maintenance toward predictive diagnostics and autonomous decision-making [4]. While this transition brings substantial benefits, it also introduces new complexities that challenge traditional approaches to Reliability, Availability, Maintainability, and Safety (RAMS) [5].

A key challenge lies in the fragmentation between traditional reliability engineering and broader system-level dependability. Established RAMS tools such as Failure Modes and Effects Analysis (FMEA) and Reliability-Centered Maintenance (RCM) remain largely hardware-focused, often overlooking the cyber, digital, and contextual elements that now define modern industrial systems. In contrast, dependability frameworks from the software domain emphasize attributes such as fault tolerance, robustness, and integrity—but frequently lack integration with physical degradation models or lifecycle asset management [6]. This disconnect hampers the development of unified strategies for managing the hybrid nature of AI-enabled manufacturing infrastructures.

A second challenge stems from the lack of lifecycle assurance and traceability in AI/ML-driven diagnostics and decision systems. As AI models are increasingly embedded into operational processes, their outputs must be explainable, auditable, and resilient to real-world uncertainty. Yet, current practice often lacks mechanisms to assess AI reliability across different deployment scenarios, data distributions, or operational contexts [7]. This creates a twofold concern: AI must contribute to system dependability, while its own behavior must also be dependable. Addressing this requires continuous validation pipelines, runtime monitoring, fallback mechanisms, and assurance frameworks tailored for AI models operating in safety-critical environments.

Third, the growing interdependence of physical and digital components introduces new vulnerabilities. Failures in edge computing devices, corrupted sensor data, or loss of network connectivity can cascade through systems and undermine availability and safety at scale [8]. These cyber-physical risks call for diagnostic models and maintenance strategies that are context-aware, adaptive, and capable of responding dynamically—capabilities that traditional RAMS methods are not well equipped to provide.

Organizational and human factors also present significant barriers. Adopting AI-driven RAMS requires not just technology but transformation—including workforce upskilling, changes in operational culture, and trust in data-centric decision-making [9]. Resistance to automation or lack of interdisciplinary collaboration can delay or derail the transition, especially in sectors where legacy systems and practices remain dominant.

Finally, sustainability goals introduce a transformative pressure on RAMS thinking. Today's manufacturing systems must be evaluated not only on technical and economic performance but also on environmental impact, resource efficiency, and long-term societal value. This expands the role of RAMS from failure avoidance to lifecycle stewardship—requiring methods that can integrate environmental and circularity metrics alongside traditional reliability indicators [10].

Together, these challenges signal the need for a new generation of dependability frameworks—ones that unify physical, digital, and organizational dimensions. The Dependability-Centered Asset Management (DCAM) approach responds to this need, offering a holistic foundation that embeds adaptability, explainability, and sustainability at the heart of next-generation manufacturing dependability.

**Advances in science and technology to meet challenges**

Overcoming the limitations of traditional RAMS frameworks in the era of smart manufacturing requires new scientific and technological approaches—ones capable of addressing the complexity of cyber-physical systems, autonomous operations, and data-driven decision environments [11]. The Dependability-Centered Asset Management (DCAM) framework offers a pathway forward by unifying lifecycle thinking, artificial intelligence (AI), and systems-level resilience into a cohesive strategy for modern manufacturing.

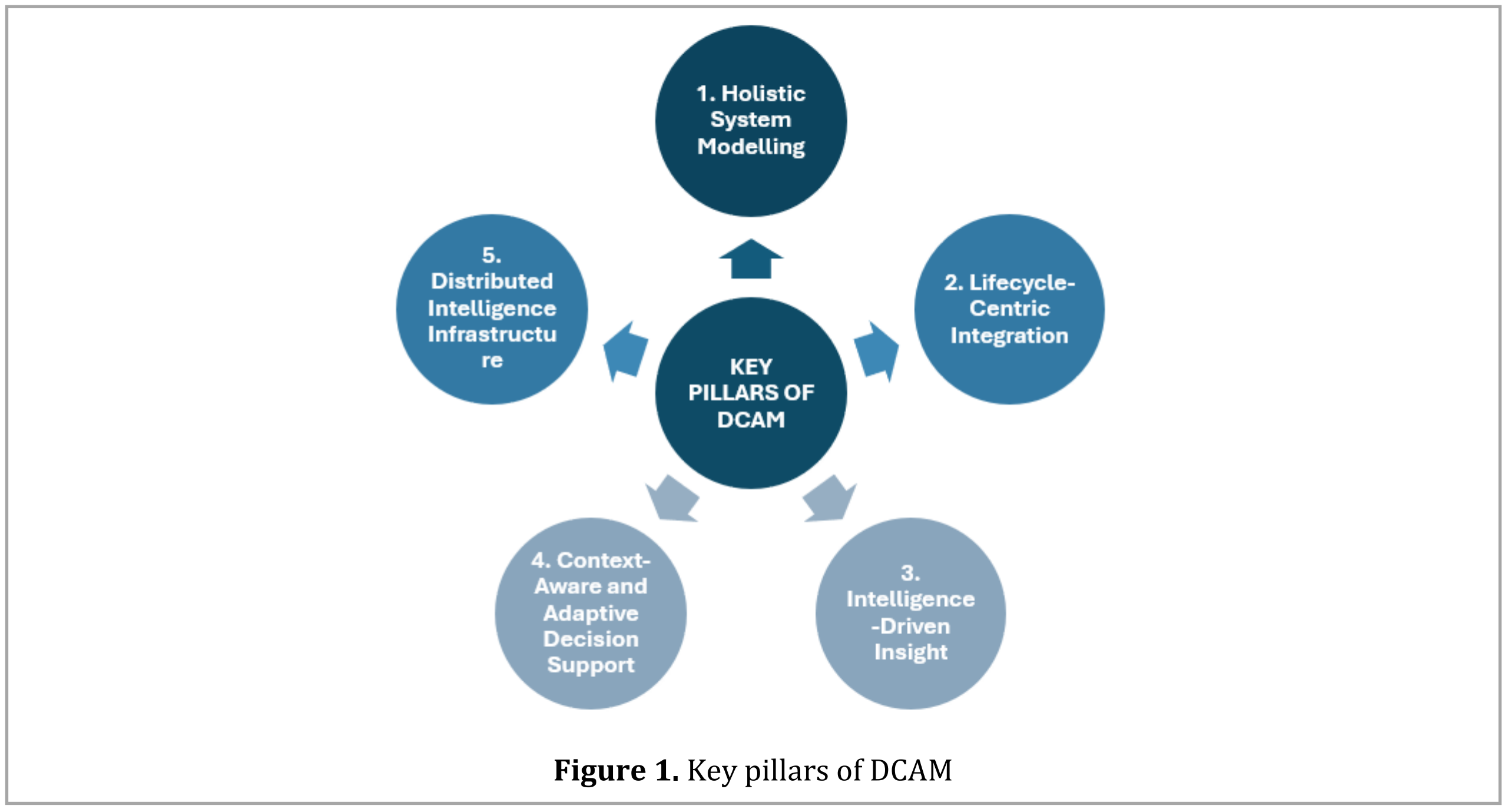

**Figure 1.** Key pillars of DCAM

Figure 1 illustrates the five foundational pillars of the DCAM framework, which collectively support AI-driven dependability across the asset lifecycle.

A foundational advancement lies in the application of non-traditional AI and machine learning (ML) methods for predictive diagnostics, anomaly detection, and remaining useful life (RUL) estimation. These technologies enable continuous asset monitoring, early detection of performance degradation, and data-driven optimization of maintenance schedules. Advanced learning techniques—such as reinforcement learning and federated learning—facilitate localized adaptation and decentralized intelligence across distributed factory systems, particularly where real-time responsiveness and data privacy are critical [12].

A second technological pillar is the increasing use of digital twins—virtual representations of physical assets and processes. Digital twins enable simulation of failure modes, evaluation of control strategies, and proactive assessment of system resilience under various operating conditions [13]. When synchronized with real-time sensor data, they support predictive analytics, scenario-based intervention planning, and long-term lifecycle optimization.

Another key advancement is the convergence of RAMS, Prognostics and Health Management (PHM), and Aging Management. Each of these disciplines offers unique contributions across different stages of the asset lifecycle. RAMS provides statistical indicators of population-level reliability (e.g., MTBF), PHM focuses on real-time monitoring and individualized RUL estimation, and aging management ensures sustainable operation through end-of-life decision-making. AI/ML technologies play a central role in integrating these domains by enabling dynamic risk modelling, context-aware scheduling, and asset-specific decision support [14].

This convergence is further supported by emerging AI paradigms such as hybrid AI (e.g., physics-informed ML) and uncertainty-aware models, which enhance both interpretability and predictive robustness [15]. These approaches are closely aligned with advances in trustworthy AI and digital twin ecosystems, reinforcing the growing need for coordinated RAMS and AI assurance strategies at the system level.

To enable safe, transparent, and explainable AI in manufacturing contexts, progress is being made in AI assurance. Techniques such as model verification, runtime monitoring, and explainability frameworks are gaining traction, particularly in safety-critical applications [16]. Additional tools—including fault injection, adversarial robustness testing, and uncertainty quantification—are increasingly used to validate ML models under operational stress and shifting data distributions.

The advancement of edge and cloud computing infrastructures has also laid the foundation for scalable, distributed intelligence. Edge devices provide low-latency monitoring and localized decision-making, while cloud platforms enable fleet-level analysis, benchmarking, and optimization [17]. Together, they establish a flexible architecture for responsive and comprehensive RAMS decision support.

Emerging capabilities in context-aware and adaptive decision-making offer another leap forward. These approaches allow maintenance and safety strategies to be tailored based on real-time contextual variables—such as asset criticality, usage intensity, environmental conditions, and cybersecurity posture. The result is more flexible, risk-informed asset management across varying operational scenarios.

Lastly, the integration of AI with sustainability analytics introduces a multi-dimensional perspective on dependability. Future systems will increasingly depend on composite performance metrics that incorporate environmental and societal factors—such as energy consumption, carbon footprint, and material circularity—alongside traditional reliability indicators [18].

Together, these advances signal a transformative shift in how dependability is conceived and managed in smart manufacturing. When embedded within the DCAM framework, they enable a move away from static, failure-avoidance paradigms toward adaptive, lifecycle-oriented, and resilience-driven strategies.

Yet, while these technologies greatly enhance the capabilities of modern RAMS, they also introduce a parallel imperative: ensuring the dependability of AI itself. As AI becomes embedded in decision-making systems, its behavior must remain trustworthy, robust, and explainable under real-world conditions. AI models must be continuously validated, monitored for data drift and degradation, and supported by fallback strategies to ensure operational reliability. In this dual role—as both enabler and subject of reliability—AI demands rigorous lifecycle assurance. Its trustworthiness is essential not only for technical performance but also for safety, compliance, and user confidence in future manufacturing systems.

### Concluding remarks

Smart manufacturing is entering a new phase—one defined by the convergence of physical, digital, and cognitive systems. While RAMS principles continue to serve as the foundation for reliable operations, they must evolve to address the complexity of modern industrial environments shaped by AI-driven automation, cyber-physical interconnectivity, and sustainability imperatives. Traditional reliability tools alone are no longer sufficient to ensure trust, adaptability, and long-term value in these intelligent, dynamic ecosystems.

The Dependability-Centered Asset Management (DCAM) framework represents a timely and necessary evolution of RAMS thinking. By embedding lifecycle awareness, system-level dependability, and AI assurance into asset management strategies, DCAM provides a holistic and future-oriented approach for managing manufacturing systems that are increasingly autonomous, data-driven, and software-defined.

This chapter has outlined the limitations of existing RAMS practices in the face of emerging technological and organizational challenges. It has also highlighted scientific and technological advances—ranging from hybrid AI models and digital twins to edge-cloud intelligence and context-aware maintenance—that enable a new generation of dependability solutions. At the center of these advances lies AI, which plays a dual role: both as a powerful enabler of predictive and adaptive capabilities, and as a source of new reliability and safety concerns. Managing this duality requires robust mechanisms for AI traceability, contextual awareness, and human-centric integration.

As manufacturing systems continue to scale in complexity and autonomy, DCAM offers a strategic framework to ensure that smart factories are not only productive and efficient, but also dependable, transparent, and sustainable across their entire lifecycle. It bridges the gap between traditional reliability engineering and the evolving needs of AI-integrated industrial systems—enabling manufacturers to design for resilience, operate with confidence, and innovate with responsibility.

Ultimately, DCAM provides a practical and adaptable roadmap for the future of manufacturing dependability—grounded in engineering rigor, enriched by AI, and guided by the principles of lifecycle stewardship and system trustworthiness.

### Acknowledgements

This work was supported in part by the National Natural Science Foundation of China (NSFC) under Grant 72471060

# Data-Centric Metrology in Future Manufacturing

**Gregory W. Vogl[1], Aaron W. Cornelius[1] and Xiaodong Jia[2]**

[1] Engineering Laboratory, National Institute of Standards and Technology, Gaithersburg, USA
[2] Department of Mechanical and Materials Engineering, University of Cincinnati, Cincinnati, USA

### Status

In modern smart factories, process data is collected and logged throughout the entire manufacturing process [1]. This creates huge datasets which include machine settings and process parameters, equipment sensor data, metrology results, and maintenance logs. Data-centric metrology (DCM) aims to leverage this collected data to provide more accurate estimates for part quality, reduce the overall measurement cost, and provide information necessary to optimize manufacturing [2]. While external information has long been used to improve measurement quality at a basic level (e.g., compensating measurements for temperature changes), DCM will actively monitor, optimize, and adapt measurements as required. With advancements in robotics and automation, DCM is poised to become a key enabler of future intelligent metrology technologies with (semi-) automated decision-making abilities to enhance precision and efficiency.

DCM combines three main pillars which have seen independent research:

(1) Integrated metrology is the incorporation and exploitation of metrologically traceable data within manufacturing systems [3]. These on-machine measurements include not just part features but also machine performance, providing real-time feedback on part quality and process health [4].

(2) Virtual metrology is used to estimate part quality for features or process parameters which cannot be directly measured *in situ*. This is done using digital twins and models which incorporate what process and metrology data is available to estimate missing parameters [5].
(3) A data management system collects all available metrology information to track part quality, provide real-time uncertainty estimates for measurements and system behaviour, and suggest actions to improve measurement performance, e.g., scheduling additional measurement cycles when virtual metrology uncertainty is too high or flagging unreliable sensors for maintenance [6].

These three areas are unified using models based on artificial intelligence (AI) to help parse and act upon the vast volumes of collected data. The semiconductor industry offers perhaps the best look into the future potential for DCM. Process data is used to generate real-time defect estimates and select wafers for further inspection [6]. The process data is then collected and analysed by machine learning (ML) tools to help operators understand and improve processes. However, the transition to smart manufacturing is not uniform: many industries lag behind and are not well-positioned to implement DCM [7]. There are significant technical gaps that must be filled to make widespread deployment of DCM practical and trusted.

**Current and future challenges**

Metrology is critical for production as the key to process and quality control, and new developments must be thoroughly validated to raise manufacturer confidence in data-driven metrology and drive adoption. The following challenges currently restrict the viability of DCM:

(1) Integrated metrology

Integrated metrology and *in situ* measurements are challenging to perform. In process measurements cannot interfere with the manufacturing process, but at the same time the measurements may be affected by the process since they occur in the same workzone [1] and may encounter various uncontrollable variations [8]. The measurements must also keep pace with the manufacturing process, further restricting what measurements are feasible to perform *in situ*. As a result, in one survey only 38% of companies performed in-situ measurements [1]. New developments are necessary to create sensors which can deliver low-uncertainty results, at an acceptable pace, and in a variety of environmental conditions.

(2) Virtual metrology

It is critical for human operators to provide their expertise and maintain visibility of the system health, which is difficult as the number of sensor data streams and automated decisions increases. Hence, new methods must be developed to help users rapidly digest, evaluate, and act upon large amounts of process data [9]. One likely path is the use of AI for automation, virtual metrology, and dynamic sampling of metrological data. Challenges for practical virtual metrology include the creation of an effective initial model using historical data and self-learning updating of models using online data [10].

(3) Data management with uncertainty quantification

The future of manufacturing depends upon secure, searchable, scalable, and standardized data architectures in which digitized information from all levels of production will enable real-time adjustements, e.g., with language-neutral identifiers and standardized machine-readable SI formats [11]. Also, data systems should be secure against cyberattacks since increasing connectivity has contributed to dramatic increases in the number of cyberattacks [12]. Challenges towards applying AI-driven insights across the product lifecycle include the curation of big data, interpretation and trust of AI-driven results [13], automatic updating of AI-based models, privacy-preserving methods, and robustness to both class imbalances [14] and variable data quality [15]. AI technologies are often difficult to generalize for deployment, since most AI/ML methods require significant training data and still may not work as intended in a different setting [13]. To gain user confidence, it is therefore imperitive to provide quantifiable uncertainty estimates for AI-based models [13].

**Advances in science and technology to meet challenges**

Figure 1 shows a roadmap to achieve data-centric metrology based on the three main pillars of DCM; (1) integrated metrology, (2) virtual metrology, and (3) data management with uncertainty quantification:

(1) Integrated metrology

Since traceability is difficult for integrated sensors that cannot be easily removed, new methods should be developed for *in situ* verification and calibration with traceability to international standards. To facilitate "hot-swapping" of poorly performing sensors, instruments can communicate real-time performance estimates to a centralized measurement management system to trigger verification cycles and flag sensors for repair or replacement. Smart sensors, which are sensors with custom ML inferences, may also be integrated into chips for real-time measurements of chip health [16]. Methods of traceability and calibration may be incorporated via calibration artifacts, self-calibration methods, and standardized processing of metrology-specific data [17], e.g., for robot-assisted metrology with fully automatic data handing.

(2) Virtual metrology

Data sampling rates within digital twins should be based on AI-driven intelligence to measure the "right amount" of data and minimize the cost of data collection and storage while maintaining product quality. For example, whenever a real-time, AI-estimated uncertainty exceeds a threshold, a measurement may be triggered to gain a data point and minimize the uncertainty at that moment while adding additional data for updating the model. DCM leverages the pattern-learning nature of AI with the trustworthiness of metrology to create trusted, yet machine-unique, models for process control [18]. Periodic comparisons of real-time traceable measurements and model estimations will help quantify the uncertainty of AI-based models. Also, the challenge of an initial model may be aided by transfer learning [10] with an initially heavy dependence on integrated metrology that lessens as the machine-specific model is learned over time.

(3) Data management with uncertainty quantification

A ubiquitous standardized data architecture is needed for all manufacturing data which validates data quality and provenance, e.g., based on OPC-UA and the digitalization of calibration reports via digital calibration certificates [19]. Methods for quantifying the total output uncertainty of AI-based algorithms, including the inherent uncertainties of the learned model and the input data uncertainties, should be developed and internationally standardized, similar to the GUM [20]. Uncertainties should be estimated to enable dynamic sampling [6] and the propogation of uncertainties, such as with a Shapley Additive exPlanations (SHAP)-based human-readable explainable AI framework [15]. Fully automated data stream handling with low computational latency presents another major challenge for DCM, requiring the innovations in Internet of Things (IoT) hardware and hardware-software optimization.

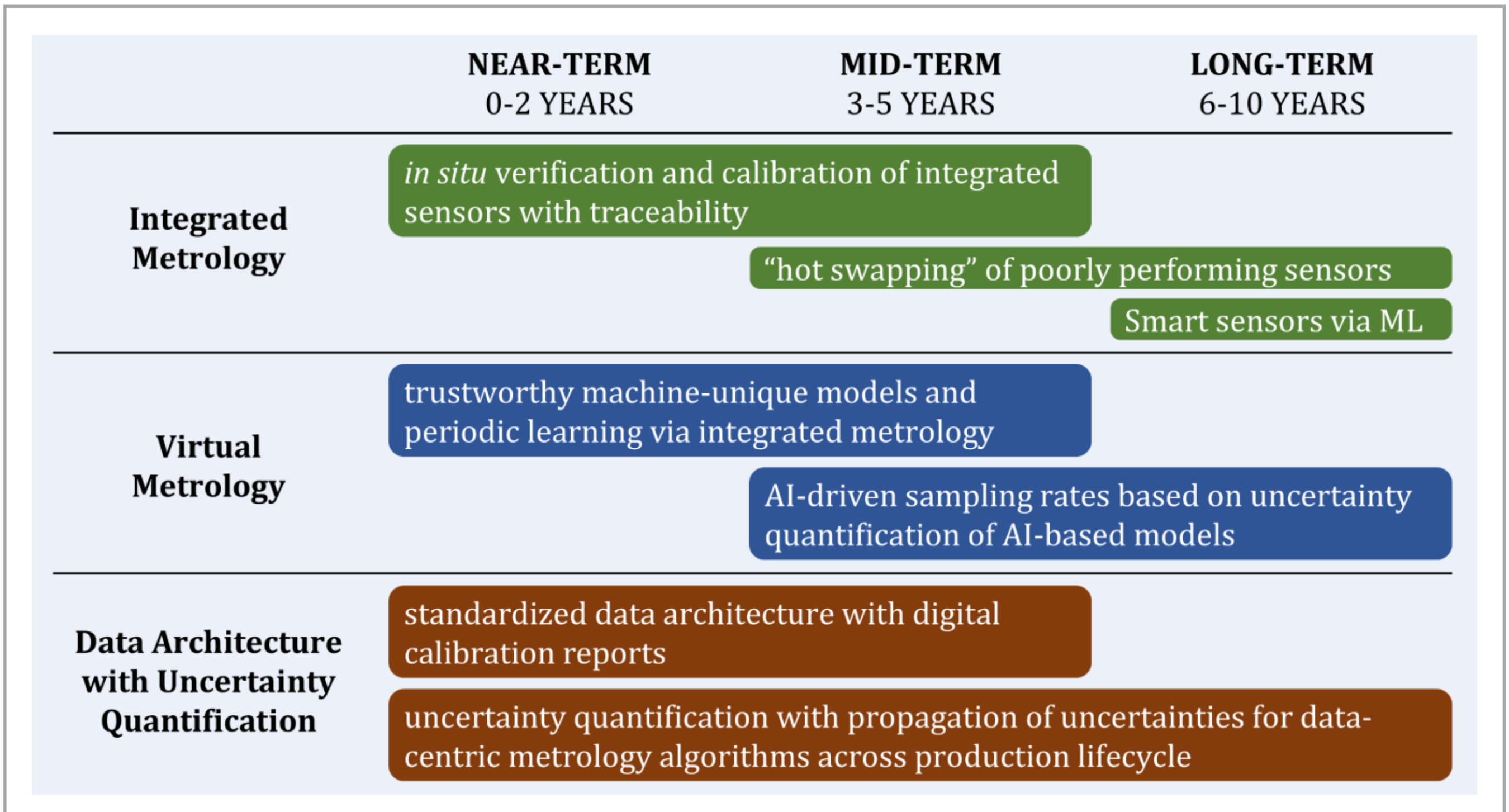

**Figure 1.** Roadmap for data-centric metrology (DCM) in future manufacturing based on the three main pillars of DCM: (1) integrated metrology, (2) virtual metrology, and (3) data management with uncertainty quantification.

## Concluding remarks

Data-centric metrology can improve manufacturing via reduced measurement costs and increased information for process optimization. Integrated metrology is used to take measurements on-machine during the manufacturing process, virtual metrology uses AI to estimate part quality based on in-process information that cannot be measured, and a data management platform uses all logged data and information to track part quality and provide uncertainty estimates. The future of DCM in manufacturing will address all challenges, e.g., via a standardized data architecture and sampling rates based on uncertainties. Ultimately, data-centric metrology will enable AI to become a trusted extension of human intelligence in manufacturing.

# Non-Traditional Machine Learning for Highly Connected and Complex Manufacturing Systems

**Dai-Yan Ji[1], Takanobu Minami[1], Ruoxin Wang[1] and Jay Lee[1]**

[1]Center for Industrial Artificial Intelligence, Department of Mechanical Engineering, University of Maryland, College Park, MD 20742, USA

E-mail: leejay@umd.edu

### Status

The development of prognostics and health management (PHM) and its integration into industrial AI has progressed from component-level monitoring toward system-level intelligence [1], [2]. In its early stages, PHM relied on reliability engineering and model-based analysis, later evolving into hybrid AI frameworks that incorporate data-driven methods with expert knowledge [3], [4]. These advances have produced valuable results for predictive maintenance and monitoring, yet the increasing scale and interconnectivity of highly connected and complex manufacturing systems (HC-CMS) continue to challenge traditional machine learning (ML) methods. Today's industrial systems are no longer limited to single components or individual units; they have become fleet-based, distributed, and deeply networked across factories, supply chains, and operational domains [5], [6]. Such systems generate vast sensor data and involve heterogeneous assets that must be managed collectively. As illustrated in Figure 1, HC-CMS connect multiple domains of industrial AI, including manufacturing AI, new energy AI, transportation AI, and healthcare AI, all of which share common requirements for resilience, adaptability, and intelligent decision-making. This system-level complexity underscores why PHM remains vital. In aerospace, energy, marine, and mobility, the ability to ensure uptime, optimize maintenance, and extend life cycles remains a cornerstone of reliability [7], [8]. With systems expanding into fleets and multi-plant networks, the importance of scalable PHM continues to grow. Further advances promise significant benefits: improved safety, reduced downtime, enhanced efficiency, and more trustworthy decision support [9]. Non-traditional machine learning is increasingly central to achieving these gains. Rather than focusing narrowly on algorithmic accuracy, the field now emphasizes resilience, interpretability, and enterprise-scale integration, and is supported by methodology platforms such as the continuous machine learning: stream-of-quality (SoQ), 5C-level cyber-physical system, and digital twin framework. Representative non-traditional ML approaches—topological data analysis (TDA) for structural insight, domain adaptation and transfer learning for fleet-wide generalization, similarity-based models for interpretable reasoning, surrogate models for efficient optimization, and industrial large knowledge models (ILKM) for knowledge integration—illustrate how PHM in HC-CMS is being reshaped and extended. These paradigms demonstrate why PHM is not only still important but also positioned to deliver even greater impact as Industrial AI continues to evolve [10].

### Current and future challenges

Despite the progress of Industrial AI, scaling PHM in HC-CMS continues to face formidable obstacles. Cross-domain variability remains a critical barrier: models trained on one production line often degrade when deployed across different plants or fleets due to domain shifts in operating conditions. This problem is amplified in industries such as wind energy, mobility batteries, and marine engines, where operational environments change rapidly and sensor distributions are inconsistent [11]. Label scarcity and data imbalance also persist, as failure events are rare, costly to capture, and frequently undocumented, limiting the applicability of supervised deep learning approaches [12]. Equally important is the challenge of system interconnectivity. Reconfigurable and sustainable production paradigms introduce nonlinear couplings, shifting bottlenecks, and complex scheduling dynamics that cannot be reduced to isolated equipment analysis. Interpretability further complicates adoption: practitioners often reject black-box models without transparent reasoning. Studies in aircraft engine prognostics have emphasized the role of aggregated feature importance and interpretable dimensionality reduction to build trust in PHM predictions [8]. Hybrid models face their own limitations. Data-driven frameworks excel when rich signals are available but struggle under distributional drift. Physics-based models are transparent yet require detailed failure physics, which are not always accessible. Hybrids attempt to combine these strengths, but parameter calibration and model updating remain complex [13]. From an organizational standpoint, knowledge fragmentation is a systemic issue. PHM efforts often rely on expertise contained in manuals, reports, and personal experience, making consistent integration into PHM difficult. Smart factory environments add further constraints, including cybersecurity, CPS/IoT integration, and the governance of heterogeneous big data. Recent reviews stress that a persistent gap between algorithmic advances and practical implementation continues to hinder PHM adoption. Based on the above description, these issues underscore the urgency of non-traditional ML techniques that embed robustness, adaptability, and explainability. Without addressing data scarcity, cross-domain adaptation, and interpretability, PHM systems will remain fragile and limited in delivering enterprise-wide value.

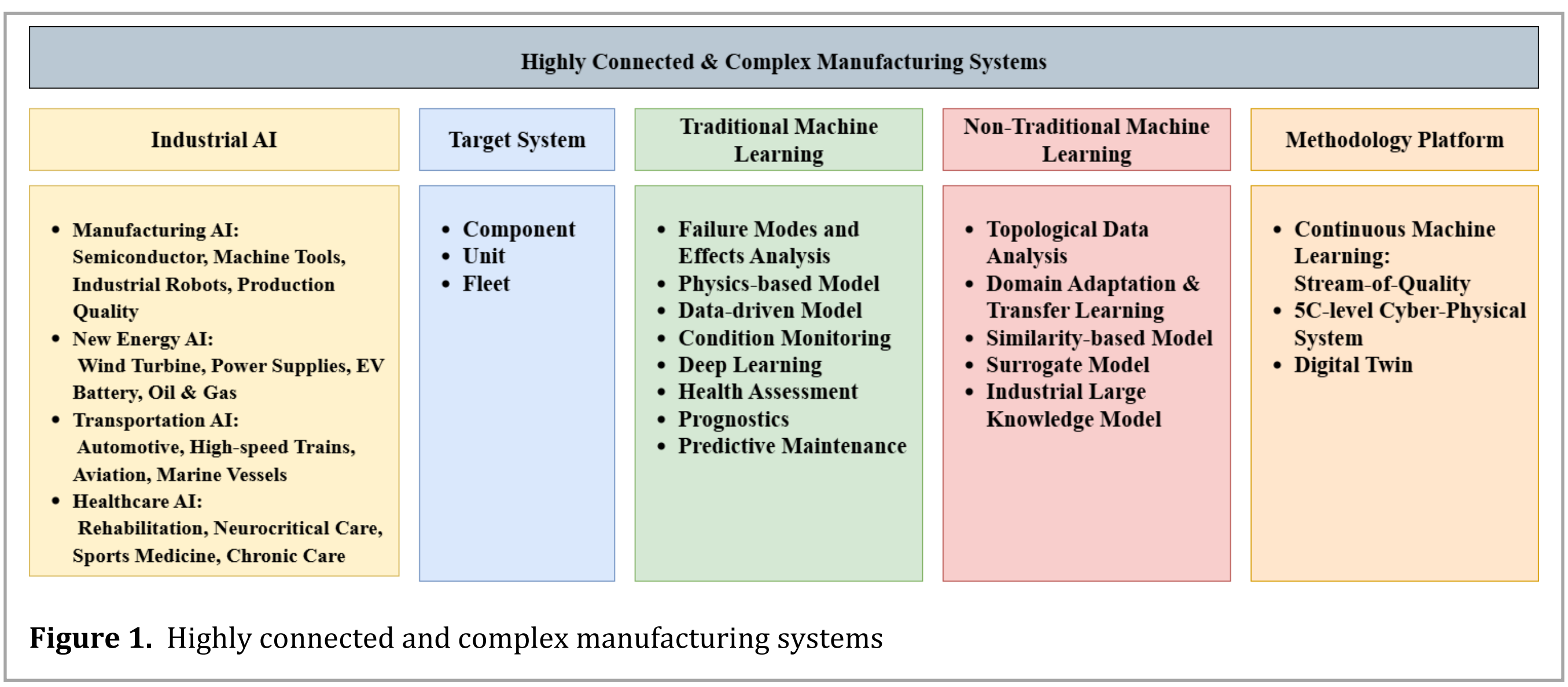

**Figure 1.** Highly connected and complex manufacturing systems

### Advances in science and technology to meet challenges

Addressing these challenges requires a series of focused scientific and technological advances. Building on this foundation, TDA has gained prominence as a streaming-compatible tool. Persistence-based descriptors allow for the detection of subtle distributional changes in production data, providing robust health indicators even under noisy and high-dimensional conditions [14]. TDA is particularly useful for long-term monitoring in HC-CMS since it not only makes anomaly detection possible but also clusters operational states and makes it easier to identify system transitions. In parallel, domain adaptation and transfer learning have been extended with ensemble, meta-learning, and continual strategies that enable models to adapt to evolving conditions in fleets of assets, such as wind farms and marine systems [15]. These methods not only reduce the cost of retraining but also support rapid deployment when labeled target data are scarce, ensuring greater generalization across diverse industrial scenarios. Extending from these developments, Similarity-based models have emerged as a crucial non-traditional method. By leveraging case libraries and distance metrics, similarity-based approaches provide transparent reasoning, enabling interpretable RUL estimation and diagnostics that practitioners can validate against historical precedents [16]. When enriched with retrieval-augmented embeddings, these models enhance both efficiency and explainability, supporting decision-making in complex operational contexts. Moreover, similarity-based reasoning fosters knowledge reuse across assets, allowing engineers to justify decisions with concrete historical references. The use of surrogate models has grown quickly. From energy systems to compact lens assemblies, neural operators, Bayesian surrogates, and differentiable simulators are being used more and more in industrial design and optimization to provide real-time decision support with quantified uncertainty [17]. Surrogate models not only speed up computationally costly simulations but also offer a way to combine physics-informed constraints with machine learning models to produce hybrid solutions that maintain a balance between interpretability and accuracy. The most transformative development is the ILKM framework. By constructing structured knowledge libraries, aligning them with industrial workflows, and coupling them with instruction-tuned large models, ILKMs enable retrieval-augmented, auditable decision-making across smart factories [18]. This approach allows PHM systems to incorporate human expertise, domain knowledge, and large-scale analytics within a unified platform, directly addressing the issue of fragmented expertise. ILKMs also create opportunities for cross-domain reasoning, linking maintenance records, quality data, and operational best practices into an integrated knowledge ecosystem. Recent reviews across Industrial AI applications consistently emphasize the importance of explainable AI, uncertainty quantification, and enterprise-level integration [19], highlighting the crucial role of non-traditional ML. Recent progress in foundation models further extends these non-traditional approaches toward system-level industrial intelligence. Under the SoQ paradigm (Figure 2), foundation model-based SoQ can be structured into four research thrusts: representation, efficient adaptation, dynamic learning, and cognitive reasoning. This emerging framework enables continuous modeling of quality propagation across stages, scalable deployment in distributed environments, real-time adaptation to evolving processes, and lifecycle-aware knowledge-driven reasoning. Such developments mark a transition from data-centric predictive maintenance toward integrated industrial cognition.

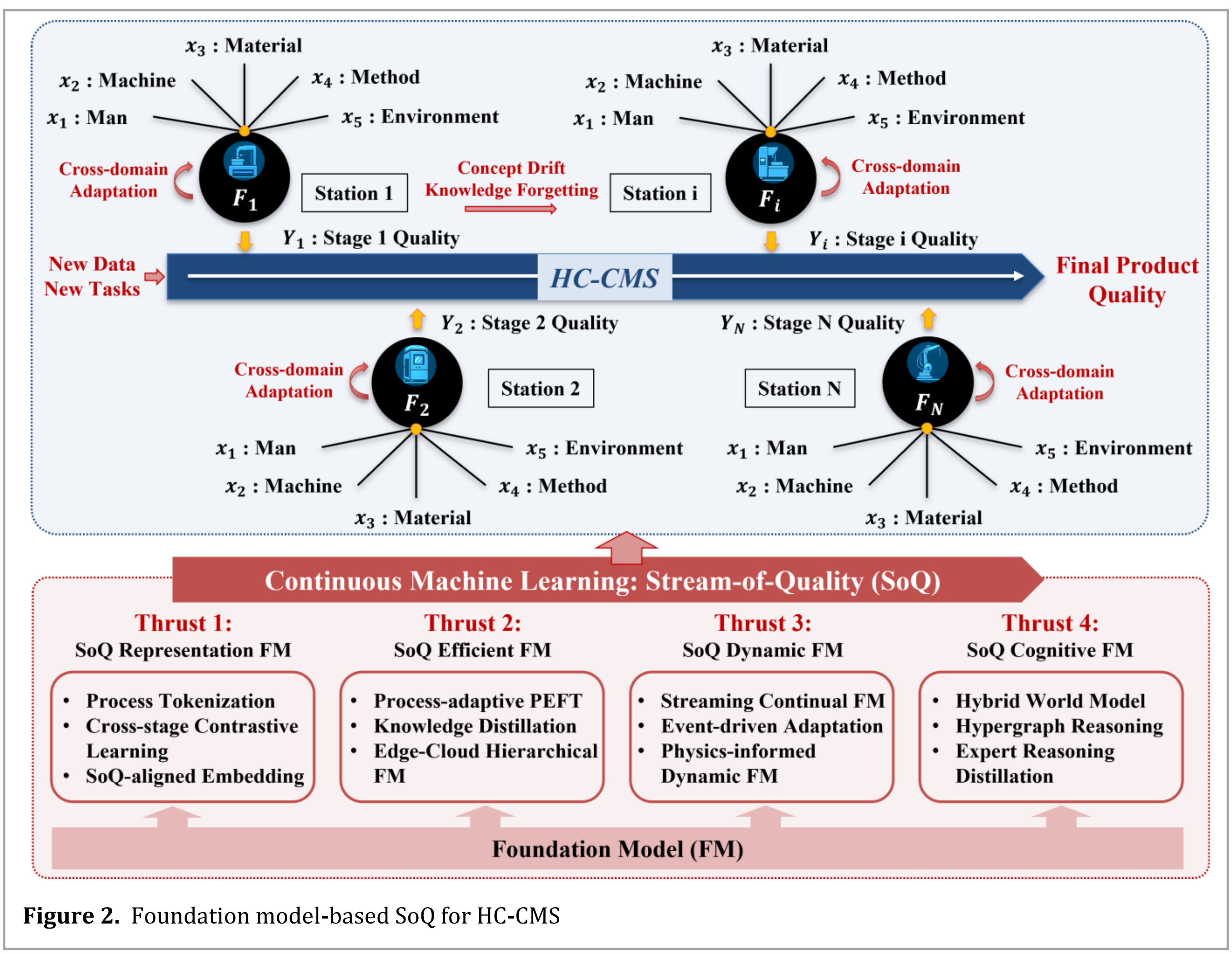

**Figure 2.** Foundation model-based SoQ for HC-CMS

### Concluding remarks

The direction of PHM research shows a clear shift from algorithm-focused investigations toward a broader emphasis on system-level intelligence in HC-CMS. Traditional ML methods, while successful in controlled settings, struggle with distributional shift, data scarcity, and limited interpretability. Non-traditional ML techniques—TDA for robust structure discovery, domain adaptation for generalization, similarity-based reasoning for transparency, surrogates for cost-efficient optimization, and ILKM for enterprise-wide integration—together provide a coherent roadmap not only for the next phase of PHM but also for the advancement of Industrial AI applications. Future PHM within Industrial AI must place emphasis on adaptability, interpretability, and scalability. In order to convert predictive accuracy into actionable intelligence, these strategies—such as ILKM frameworks that integrate disparate areas of expertise and non-traditional ML methods grounded in domain knowledge—will be crucial. Higher uptime, safer operations, lower maintenance costs, and more robust production ecosystems are expected advantages of these advances. The main challenge [20] for the community is not the pursuit of algorithmic novelty, but rather the successful integration of these techniques into enterprise-scale, auditable, and reliable industrial systems that can deliver lasting impact.

## Acknowledgements

This work was supported by the U.S. Department of Education through the Fund for the Improvement of Postsecondary Education (FIPSE) under Grant No. P116S230014.